%% file: emnlp2023.tex
\pdfoutput=1
\documentclass[11pt]{article}

\usepackage[]{acl}
\usepackage{times}
\usepackage{latexsym}
\usepackage[T1]{fontenc}
\usepackage[utf8]{inputenc}
\usepackage{microtype}
\usepackage{inconsolata}
\usepackage{hyperref}
\usepackage{caption}
\usepackage{subcaption}
\usepackage{enumitem}

\usepackage{graphicx}
\usepackage{subfloat}
\usepackage{booktabs} % To thicken table lines
\usepackage{tabularx}
\usepackage{multirow}
\usepackage{amsmath,amssymb}
\usepackage{cuted}
\usepackage{flushend}
\usepackage{tablefootnote}
\usepackage{wrapfig}
\usepackage{xcolor}

\usepackage{ragged2e}
\usepackage{colortbl}
\usepackage{mymacros}
\usepackage[framemethod=TikZ]{mdframed}
\usepackage{subcaption}
\usepackage{float}
\extrafloats{100}
\newif\ifcomments
\commentstrue

\newmdenv[%
    backgroundcolor=gray!10,
    linecolor=black,
    outerlinewidth=0.5pt,
    roundcorner=1mm,
    skipabove=\topsep,
    skipbelow=\topsep,
    font=\ttfamily\tiny,
]{promptbox}

\newcommand{\systemname}{\textsc{Megaverse}}
\newcommand{\prevsystemname}{\textsc{Mega}}

\title{\systemname\ : Benchmarking Large Language Models Across \\ Languages, Modalities, Models and Tasks}

\author{\parbox{0.9\linewidth}{\centering{Sanchit Ahuja \quad Divyanshu Aggarwal \quad Varun Gumma \quad Ishaan Watts\\ \quad Ashutosh Sathe \quad Millicent Ochieng \quad Rishav Hada \quad Prachi Jain \\ \quad Mohamed Ahmed \quad Kalika Bali \quad Sunayana Sitaram \\
 {\rm  Microsoft Corporation } \\
 {\tt \{t-sahuja,sunayana.sitaram\}@microsoft.com}}}}

\begin{document}
\maketitle

\input{content/abstract}
\input{content/intro}
\input{content/related}
\input{content/expts}

\input{content/results}

\input{content/contamination}
\input{content/discussion}

\input{content/limitations}

\bibliography{anthology,custom}

\newpage
\appendix
\input{appendix/tasks_and_datasets}
\input{appendix/prompts}
\input{appendix/fertility}
\input{appendix/vision_visualization}
\input{appendix/results_figures}
\input{appendix/results_tables}
\input{appendix/apex_cont}
\end{document}

%% file: content/abstract.tex
\begin{abstract}
There has been a surge in LLM evaluation research to understand LLM capabilities and limitations. However, much of this research has been confined to English, leaving LLM building and evaluation for non-English languages relatively unexplored. Several new LLMs have been introduced recently, necessitating their evaluation on non-English languages. This study aims to perform a thorough evaluation of the non-English capabilities of SoTA LLMs (GPT-3.5-Turbo, GPT-4, PaLM2, Gemini-Pro, Mistral, Llama2, and Gemma) by comparing them on the same set of multilingual datasets. Our benchmark comprises 22 datasets covering 83 languages, including low-resource African languages. We also include two multimodal datasets in the benchmark and compare the performance of LLaVA models, GPT-4-Vision and Gemini-Pro-Vision. Our experiments show that larger models such as GPT-4, Gemini-Pro and PaLM2 outperform smaller models on various tasks, notably on low-resource languages, with GPT-4 outperforming PaLM2 and Gemini-Pro on more datasets. We also perform a study on data contamination and find that several models are likely to be contaminated with multilingual evaluation benchmarks, necessitating approaches to detect and handle contamination while assessing the multilingual performance of LLMs.
\end{abstract}

%% file: content/intro.tex
\section{Introduction}
\label{sec:intro}

Large Language Models (LLMs) have surpassed the performance of previous generation of language models on several tasks and benchmarks, sometimes even approaching or exceeding human performance \cite{Hubert_Awa_Zabelina_2024}. However, the root cause of the observed capabilities in these models is not always apparent, whether stemming from augmented model capabilities or other factors like contamination in test datasets and the absence of datasets that genuinely measure the capabilities of these models \cite{balloccu-etal-2024-leak}. Thus, evaluation of Large Language Models has become an important field of study. 

Most of the work on evaluating LLMs via benchmarking \cite{liang2022holistic}, qualitative tests for specific capabilities \cite{bubeck2023sparks} or human evaluation have focused solely on English. However, studies have shown that there is a large gap between the capabilities of LLMs in English and other languages \cite{choudhury2023ask}. Evaluation of LLMs in languages other than English is challenging due to a variety of factors, including the lack of benchmarks covering a large number of languages from diverse language families and the lack of multilingual benchmarks covering tasks such as reasoning, chat, and dialogue. Therefore, it is crucial to prioritize multilingual evaluation to enhance the development of more effective multilingual models. Neglecting this critical aspect may result in a significant population being left behind and may widen the digital divide \cite{joshi2021state}.

Our prior work on evaluating multilingual capabilities of LLMs, \prevsystemname\ \cite{ahuja-etal-2023-mega}, yielded the following observations: 
GPT-4 \cite{gpt4techreport} comes close to the performance of SOTA fine-tuned language models such as TULRv6 \cite{patra-etal-2023-beyond}. GPT models perform worse on languages that are written in non-Latin scripts, and on low-resource languages. Other LLMs such as BLOOMZ \cite{muennighoff-etal-2023-crosslingual} usually perform worse than GPT-4. However, several newer models are comparable to GPT-4 in performance on English, and it is essential to study their multilingual performance as well. Moreover, there is a rising interest in Large Multimodal Models (LMMs), and the convergence of multimodal and multilingual LLMs remains an understudied area \cite{hu2024large}. Our contributions are as follows:

\begin{itemize}
    \item We build on top of the \prevsystemname\ benchmark and add 6 new datasets, thus extending coverage to 22 datasets and 83 languages including many low-resource African languages.
    \vspace{-2mm}
    \item We benchmark nine new SOTA text LLMs - PaLM2 \cite{palm2techreport}, Llama2 (3 variants) \cite{touvron2023llama}, Mistral-v1.0 (2 variants),  \cite{jiang2023mistral}, Gemma (2 variants) \cite{team2024gemma}, Gemini 1.0 pro \cite{team2023gemini} in addition to GPT-4 and GPT-3.5-Turbo.
    \vspace{-2mm}
    \item We benchmark the multimodal LLaVA family models \cite{liu2023improved}, GPT-4-Vision \cite{openai_gptv} and Gemini-Pro-Vision \cite{team2023gemini} on two  multilingual multimodal datasets.
    \vspace{-2mm}
    \item We present a thorough contamination study of both commercial and open-source set of LLMs on a subset of our datasets.
    % \vspace{-2mm}
    \item We study the overall trends in our experiments by studying the deviation of performance across language families and tasks, and provide directions for future research.
\end{itemize}

%% file: content/related.tex
\section{Related work}
\label{sec:related}

\paragraph{Evaluation of LLMs} Recently, there has been an increasing interest in evaluating LLMs on a wide range of capabilities, given the surge in their popularity and effectiveness. BIG-Bench \cite{srivastava2023imitation} consists of 204 tasks to evaluate LLMs. 

While BIG-Bench includes tasks in non-English languages as well, they are largely related to translation. \citet{liang2022holistic} proposed HELM, defining a taxonomy of scenarios and metrics that define the space of LLM evaluation, and evaluating 30 language models on 42 scenarios and 7 metrics. However, all the scenarios are focused on datasets in standard English or dialects, and they highlight coverage of languages as an important area for improvement. \citet{bubeck2023sparks}, has pointed out the limitations of using standard NLP benchmarks to evaluate generative models, due to the pace at which these benchmarks become saturated. There are also concerns about benchmark contamination in LLM evaluation. \newcite{zhou2023don} show that test dataset contamination in training and fine-tuning data leads to a significant impact on LLM performance. 

\paragraph{Multilingual Benchmarks and Evaluation} \newcite{bang2023multitask} evaluates the multilingual capabilities of ChatGPT and shows that it fails to generalize to low-resource languages with non-Latin scripts. However, multilingual evaluation is performed only on a few tasks, and a subset of 50-100 examples are used for testing the model. \citet{hendy2023good} evaluate the translation abilities of GPT-3.5 models and find that these models perform well in translating high-resource languages, but their capabilities for low-resource languages are limited. BUFFET \cite{asai2023buffet} covering 54 languages across 15 datasets and \citet{lai2023chatgpt} covering 37 languages across 7 datasets also perform multilingual benchmarking of LLMs such as ChatGPT and BLOOMZ. \newcite{yang2023dawn} does a comprehensive study of GPT4-Vision's capabilities that include analyzing its performance on multilingual image description, scene text recognition, and translation. Our work builds on the \prevsystemname\ benchmarking effort \cite{ahuja-etal-2023-mega}, which evaluates GPT models across 16 datasets. We extend the \prevsystemname\ benchmark to more tasks including multimodal tasks, evaluate several SoTA LLMs, and perform a more comprehensive analysis of contamination. 

\paragraph{Contamination} Several techniques have been proposed to study the contamination of publicly available evaluation datasets. \citet{ahuja-etal-2023-mega} study contamination by prompting the models to fill dataset cards. Other methodologies encompass \citet{golchin2023time}, which does not provide quantification of contamination, and \citet{oren2023proving}, which requires access to log probabilities, thereby limiting their studies to open-sourced LLMs. 

% Further, they qualitatively analyze these cards and assign the datasets one of 3 categories for contamination: full, partial, or none.

% Our benchmarking suite \prevsystemname\ is extended in \prevsystemname\VERSE to encompass 6 new datasets, among which are two multimodal datasets. Furthermore, we conduct benchmarking on state-of-the-art language models such as PaLM2, Llama2, Mistral and vision models such as LlaVa and GPT4-Vision and compare the latest LLMs in terms of their multilingual performance.

%% file: content/expts.tex
\section{Experimental Setup}
\label{sec:expts}

\begin{figure*}
\centering
\begin{subfigure}[h]{\textwidth}
    \includegraphics[width=\textwidth]{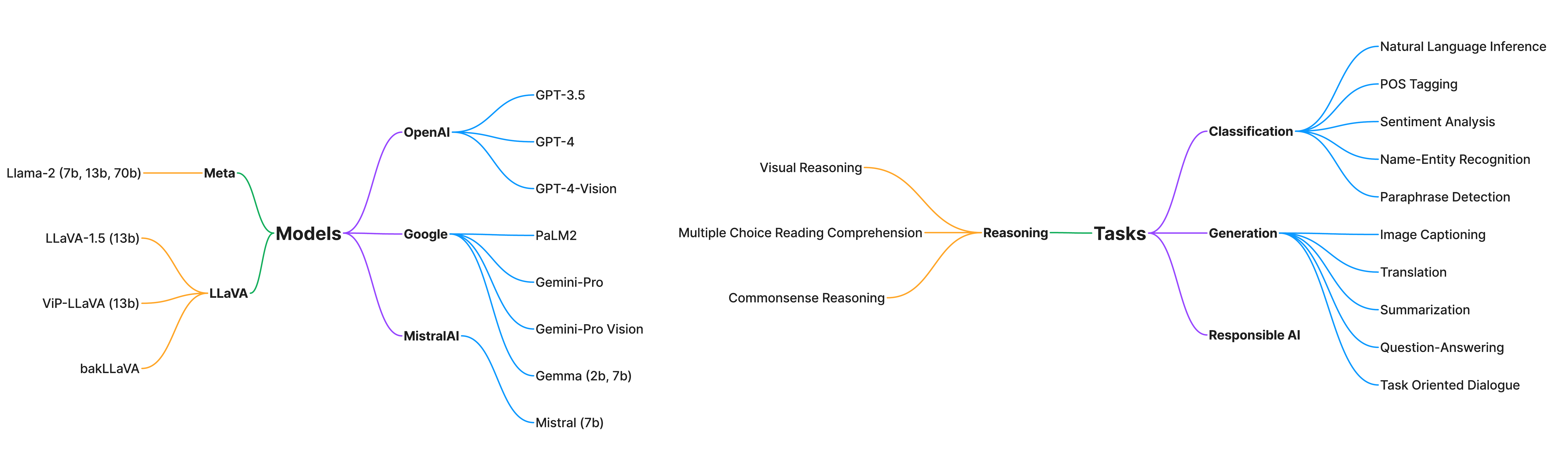}
\end{subfigure}
\caption{Hierarchy of Models and Tasks spread across \systemname}
\label{fig:task_dataset_lang_family}
\end{figure*}

\subsection{Datasets}
We perform experiments on the 16 datasets that are part of the \prevsystemname\ suite - XNLI \cite{Conneau2018xnli}, IndicXNLI \cite{aggarwal-etal-2022-indicxnli}, GLUECoS NLI \cite{khanuja2020new}, PAWS-X \cite{Yang2019paws-x}, XCOPA \cite{ponti2020xcopa}, XStoryCloze \cite{lin-etal-2022-shot}, GLUECoS Sentiment Analysis (En-Es-CS) \cite{vilares2016cs}, TyDiQA-GoldP \cite{clark2020tydi}, MLQA \cite{lewis2020mlqa}, XQUAD \cite{artetxe2020cross}, IndicQA \cite{doddapaneni-etal-2023-towards}, PAN-X \cite{pan2017cross}, UDPOS \cite{nivre2018universal}, Jigsaw \cite{jigsaw-multilingual-toxic-comment-classification}, WinoMT \cite{stanovsky2019evaluating} and XLSum \cite{hasan2021xl}. These datasets include a mix of classification, Question Answering, Sequence Labeling, and Natural Language Generation datasets, along with two datasets covering the Responsible AI tasks of toxicity detection and gender bias. The datasets we include also contain a mix of translated datasets verified by native speakers, as well as datasets created independently for each language. Figure \ref{fig:task_dataset_lang_family} shows a hierarchy of models and tasks spread across \systemname.\ For a more detailed description of the datasets included in the original \prevsystemname\ benchmark, we refer the readers to \citet{ahuja-etal-2023-mega}. We describe the six datasets added to our study below.

\subsubsection{AfriQA}

AfriQA \cite{ogundepo2023afriqa} is a QA dataset that does not have a context passage. It covers 10 African languages - Bemba, Fon, Hausa, Igbo, Kinyarwanda, Swahili, Twi, Wolof, and Yorùbá. We use the few-shot size of $k$ = 4 and the monolingual prompting strategy to perform experiments only on the GPT and Llama models, as the PaLM2 model only supports Swahili. 

\subsubsection{Belebele}

Belebele \cite{bandarkar2023belebele} is a multiple choice machine reading comprehension (MRC) dataset parallel across 122 languages. Each question is linked to a short passage from the FLORES-200 dataset \cite{nllbteam2022language}. The human annotation procedure was carefully curated to create questions that discriminate between different levels of language comprehension. We evaluated Arabic, Czech, Danish, German, English, Spanish, Finnish, French, Hebrew, Hungarian, Italian, Japanese, Korean, Dutch, Norwegian, Polish, Portuguese, Russian, Swedish, Thai, Turkish, Chinese Simplified and Chinese Traditional. Results for Llama2 and GPT-3.5-Turbo are reported from the dataset paper. We perform zero-shot monolingual prompting for our experiments, as this dataset does not have a dev set.

\subsubsection{IN22}
IN22 \cite{gala2023indictrans} is a translation benchmark for all 22 scheduled Indic languages. IN22-Gen is a general-purpose multi-domain evaluation subset of IN22 which has been curated from two sources: Wikipedia and Web Sources offering diverse content spanning news, entertainment, culture, legal, and India-centric topics. IN22-Conv is the conversation domain subset of IN22. Due to resource constraints, we evaluate 14 languages: Assamese, Bengali, English, Gujarati, Hindi, Kannada, Kashmiri, Malayalam, Marathi, Nepali, Odia, Punjabi, Tamil, Telugu, and Urdu. 

\subsubsection{MaRVL}
MaRVL (Multicultural Reasoning over Vision and Language) \cite{liu2021visually} is a dataset of images and associated captions. The concepts and images collected were entirely driven by native speakers and are representative of various cultures across the globe and span 5 languages, i.e., Indonesian, Chinese, Swahili, Tamil, and Turkish. Each instance in the dataset consists of a pair of images (left image and right image) and a statement, and the task is to determine whether the statement is consistent for the given pair of images.

\subsubsection{XM-3600}
CrossModal-3600 \cite{thapliyal2022crossmodal} is a multilingual image captioning dataset consisting of 3600 geographically diverse images directly captioned in 36 different languages, avoiding any inconsistencies due to translations. We experimented on 20 out of 36 languages due to resource constraints: Arabic, Chinese, Czech, Danish, Dutch, English, Finnish, French, German, Italian, Japanese, Korean, Norwegian, Polish, Portuguese, Russian, Spanish, Swedish, Thai, and Turkish. 

\subsubsection{XRiSAWOZ}
XRiSAWOZ \citep{moradshahi-etal-2023-x} is a task-oriented dialogue modeling dataset. The dataset is a multilingual (English, Hindi, French, Korean) translation of the Chinese-only RiSAWOZ dataset \citep{quan-etal-2020-RiSAWOZ}. XRiSAWOZ also includes an \textbf{English-Hindi code mixed} setting. For each conversation, the agent must make use of structured knowledge from the databases to answer user queries. The task consists of  4 subtasks: ``Dialogue State Tracking'' (DST), ``API Call Detection'' (API), ``Dialogue Act Generation'' (DA) and ``Response Generation'' (RG). The metrics used for evaluation include BLEU, Slot Error Rate (SER) (factual correctness of generated response) \citep{wen-etal-2015-semantically}, (averaged/task) success rate \citep{lin2021bitod}, API call accuracy, dialogue act accuracy and joint goal accuracy \citep{budzianowski-etal-2018-multiwoz}. We refer the reader to  \citet{moradshahi-etal-2023-x} for detailed descriptions of subtasks and metrics. We perform experiments on 10\% of the data i.e. about 400 dialogue turns across 3 domains due to limited compute.

\subsection{Models\protect\footnote{We set the temperature parameter equal to 0 (or close to 0) for all our models to ensure deterministic output and reproducibility}}
Below is a list of all the models we evaluate:
\begin{itemize}[itemsep=1pt,parsep=1pt]
    \item \textbf{GPT-3.5-Turbo} \cite{ouyang2022training} 
    \item \textbf{GPT-4} \cite{gpt4techreport}
    \item \textbf{GPT-4-Vision} \cite{openai_gptv}
    \item \textbf{Llama2 (7B, 13B, 70B)} \cite{touvron2023llama}
    \item \textbf{PaLM2} \cite{anil2023palm}
    \item \textbf{Gemini-Pro} \cite{team2023gemini} 
    \item \textbf{Gemini-Pro-Vision} \cite{team2023gemini}
    \item \textbf{Gemma (2B, 7B)} \cite{team2024gemma}
    \item \textbf{Mistral} \cite{jiang2023mistral}
    \item \textbf{BakLLaVA-v1} \cite{liu2023improved}
    \item \textbf{ViP-LLaVA (13B)} \cite{cai2023making}
    \item \textbf{LLaVA-1.5 (13B)} \cite{liu2023improved}
\end{itemize}

\subsection{Prompting strategies} \label{sec:prompting_strats}

\citet{ahuja-etal-2023-mega} explore three prompting variations based on the language of the few-shot and test examples, and find that monolingual prompting, featuring few-shot examples in the target language, outperforms zero-shot cross-lingual prompting in English for most datasets. Translate-test excels over monolingual for certain low-resource languages but with minimal gaps for models like GPT-4. Therefore, we default to monolingual prompting unless otherwise specified. Zero-shot cross-lingual prompting (zs-cl) is used when dev datasets are unavailable in the target language. English instructions are maintained for prompts, proven to outperform instructions in the target language \cite{ahuja-etal-2023-mega}. Prompt templates for our new datasets are in the Appendix \ref{sec:prompt_templates}.

\subsubsection{XRiSAWOZ} \label{sec:xrisawoz_prompting}

\citet{moradshahi-etal-2023-x} presents results in both end-to-end and turn-by-turn evaluation settings. We perform end-to-end evaluation with regex based careful filtering of the generated responses for DST/API/DA tasks after every turn. This is required to ensure correctness of the syntax in the state descriptions for these tasks. No such postprocessing is done for the RG task. 
For inferring a subtask on a dialogue turn, we provide in-context examples corresponding to the same turn from other domains. If for a particular turn, sufficient in-context examples are not available, we look for the \textit{latest previous turn} for which sufficient in-context examples are available. E.g. Assume the following turn to count distribution and $k$ = 4 (number of in-context examples). Turns 1--4: more than 10 examples, Turn 5: 3 examples, and Turn 6 has 1 example.

At turns 5 and 6, we do not have sufficient examples from turn 5 or 6. Therefore, we sample in-context examples from turn 4 for both of them. 
Our prompts for each subtasks can be seen in Fig. \ref{fig:xRiSAWOZ_prompt}, \ref{fig:DST}, \ref{fig:API}, \ref{fig:DA}, \ref{fig:RG}.

%% file: content/results.tex
\section{Results}

\label{sec:results}
\subsection{XNLI}

All models perform best on English, with slightly lower performance on Greek and German, and lower performance on languages like Hindi, Thai, Urdu, and Swahili. Overall PaLM2 performs best, closely followed by GPT-4. GPT-3.5-Turbo is worse on all languages, however, we find that all three Llama models perform substantially worse, with Mistral performing the worst. Since XNLI is a popular dataset, dataset contamination cannot be ruled out. (Figure \ref{fig:xnli}, Table \ref{tab:results_summary_xnli}).

\subsection{IndicXNLI}
We performed experiments on IndicXNLI on the GPT models, Mistral as well as Llama models, however, the Llama models gave scores of 0 for all languages, which is why we do not plot them. The Mistral model also performs poorly. We find that GPT-4 outperforms GPT-3.5-Turbo on all languages with the highest scores on Hindi, Punjabi, and Bengali. However, the overall accuracy is not very high on any language compared to the XNLI results seen earlier, and fine-tuned baselines such as MuRIL perform best. (Figure \ref{fig:indicxnli}, Table \ref{tab:results_summary_indicxnli}).

\subsection{GLUECoS NLI}
All models do well on this NLI task, with GPT-4 performing best. (Figure \ref{fig:gluecos}, Table \ref{tab:cm_results}).

\subsection{PAWS-X}
PaLM2 outperforms the GPT models on all languages and all models perform well, which could be because this dataset contains high-resource languages. However, dataset contamination cannot be ruled out, as shown in \citet{ahuja-etal-2023-mega}. The performance on English performs is the best, followed closely by Latin script languages, and a drop in performance for languages in other scripts. The Llama and Mistral models perform worse than the GPT models and PaLM2, although the difference in performance is not as large as in some of the other datasets. (Figure \ref{fig:pawsx}, Table \ref{tab:results_summary_pawsx}).

\subsection{XCOPA}
The performance of GPT-4, Gemma, Gemini and PaLM2 are comparable, with GPT-4 having the best perforamnce. Notably, they are all better than GPT-3.5-Turbo, which performs substantially better than the Llama2 and Mistral models except in Quechua, for which no model performs well. However, the results on all other languages for GPT-4 and PaLM2 are extremely high, which may be due to dataset contamination. (Figure \ref{fig:xcopa}, Table \ref{tab:results_summary_xcopa}).

\subsection{XStoryCloze}

Since the Llama models gave scores of 0 for all languages, we omit it from our analysis. We find that the gap between the GPT models and PaLM2 is very high, with both GPT models performing extremely well. For all languages except Telugu, Basque and Burmese Gemini-pro performs well.  The contamination
study from \citet{ahuja-etal-2023-mega} show a low 
chance of dataset contamination for GPT-4, which indicates that the GPT models can perform this task well. (Figure \ref{fig:xstorycloze}, Table \ref{tab:results_summary_xstorycloze}).

\subsection{Sentiment Analysis (En-Es-CS)}

Surprisingly, GPT-3.5-Turbo outperforms both GPT-4 and PaLM2 on this task, with the mBERT baseline performing the best, while Gemini-pro performs the worst by a large margin. (Figure \ref{fig:gluecos}, Table \ref{tab:cm_results}).

\subsection{TyDiQA GoldP}

The TuLR model performs best, followed by GPT-4, PaLM2, Gemini-Pro, and BLOOMZ, while Llama models perform poorly, with Mistral being slightly better. Smaller models, in particular, demonstrate a significant performance gap between English and all other languages. However, dataset contamination cannot be ruled out, as shown in \citet{ahuja-etal-2023-mega}. (Figure \ref{fig:tydiqa}, Table \ref{tab:results_summary_tydiqa}).

\subsection{MLQA}

TULR and GPT-4 outperform all other models for this dataset except for German. English exhibits superior performance, with Spanish (es), German (de), and Vietnamese (vi) following closely. The most significant gaps are noted between English and Arabic (ar), Hindi (hi), and Chinese (zh) The Llama2-13B model performs well for some languages, such as Arabic, German, and Spanish but performs poorly on Chinese Hindi, and Vietnamese, but is still better than Mistral and Gemma. This is one of the datasets where PaLM2 struggles, particularly for Arabic and Chinese. Dataset contamination in GPT-4 cannot be ruled out, as shown in \citet{ahuja-etal-2023-mega}. Smaller versions of the Llama model outperform the Llama 70B model across all languages. (Figure \ref{fig:mlqa}, Table \ref{tab:results_summary_mlqa}).

\subsection{XQUAD}

\begin{figure*}
\centering
    \includegraphics[width=\textwidth]{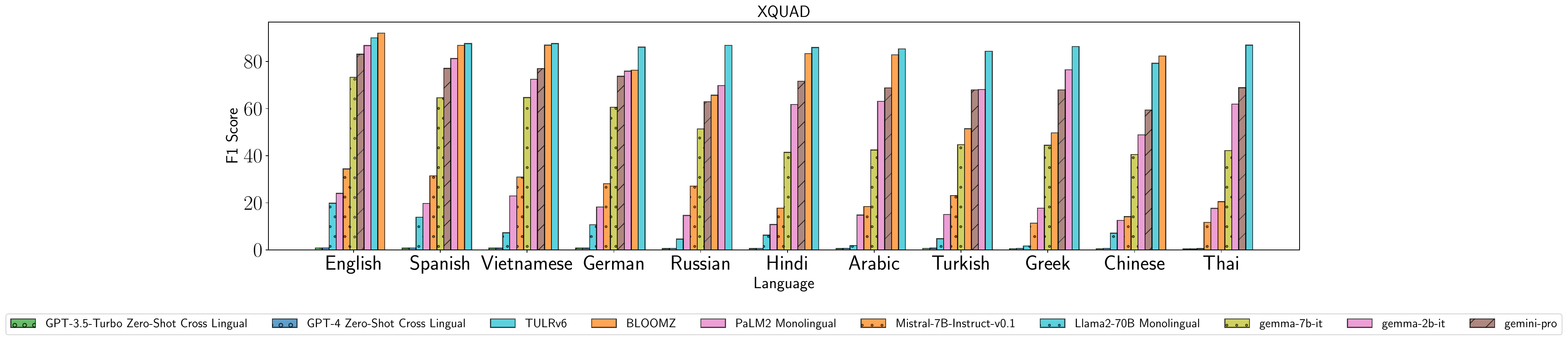}
    \caption{Results for XQUAD across all languages and models for zero-shot cross-lingual prompting}
    \label{fig:xquad}
\end{figure*}

TuLRv6 performs best across almost all languages in the XQuAD dataset, followed by GPT-4, PaLM 2, Gemini-Pro, and BLOOMZ. BLOOMZ's performance declines significantly in Greek and Thai as shown in Figure \ref{fig:xquad}. PaLM2 and Gemini-Pro exhibit competitive performance, closely trailing GPT-4-32K and TuLRv6 – XXL across languages from high to mid-resource tiers. All three Llama models perform poorly on this dataset. Gemma and Mistral perform slightly better than Llama on all languages but lags behind the larger models and finetuned models. Dataset contamination in GPT-4 cannot be ruled out, as shown in \citet{ahuja-etal-2023-mega}. (Figure \ref{fig:xquad}, Table \ref{tab:results_summary_xquad}).

\subsection{IndicQA}

Since the Llama models gave scores of 0 for all languages, we omit it from our analysis. We use the zero-shot cross-lingual prompting strategy due to the absence of a dev set. GPT-4 performs better than GPT-3.5-Turbo, with the best performance seen for Hindi, Marathi, and Bengali, while the smaller models like Gemma perform poorly. (Figure \ref{fig:indicqa}, Table \ref{tab:results_summary_indicqa}).

\subsection{PAN-X}
GPT-4 and GPT-3.5-Turbo outperform PaLM2 and gemini-pro for most languages. However, all models perform poorly on Thai, Japanese, and Chinese on this sequence labeling task. Since this is an older dataset, GPT-4 data contamination cannot be ruled out as shown in \citet{ahuja-etal-2023-mega}. (Figure \ref{fig:panx}, Table \ref{tab:results_summary_panx}).

\subsection{UDPOS}

PaLM2 performs the best followed by GPT-4, GPT-3.5-Turbo and Gemini-pro being the worst on average. All models show similar high performance across languages, except for Arabic, Greek, Hebrew, Hindi, and Vietnamese, where PaLM2 performs best. GPT-4 data contamination cannot be ruled out as shown in \citet{ahuja-etal-2023-mega}. (Figure \ref{fig:udpos}, Table \ref{tab:results_summary_udpos}).

\subsection{Jigsaw}
We perform experiments on the Jigsaw dataset for GPT-3.5-Turbo and PaLM2 using the monolingual prompting strategy and find that both models perform very well on all languages. Since the dataset cannot be accessed without download, models are less likely to be contaminated with this dataset. (Figure \ref{fig:jigsaw}, Table \ref{tab:jigsaw_results_summary}).

\subsection{WinoMT}

We perform experiments on the WinoMT dataset only for GPT-3.5-Turbo using the monolingual prompting strategy and report the results for completeness. We find that the model does not perform well on any of the languages. (Figure \ref{fig:winomt}, Table \ref{tab:wino-mt}).

\subsection{XLSum}

GPT-4 outperforms all other models, with some exceptions. GPT-3.5-Turbo performs best for African languages like Swahili, Somali, and Yoruba, while the Llama models perform best for Arabic, Kyrgyz, Vietnamese, and Welsh. According to the contamination analysis in \citet{ahuja-etal-2023-mega}, it is possible, though less likely that GPT-4 is contaminated with this dataset. (Figure \ref{fig:xlsum}, Table \ref{tab:results_summary_xlsum}).

\subsection{Belebele}

Gemini-Pro has the best performance amongst all the models for most languages, while for smaller models only Llama models come close. GPT-4 and PaLM2 outperform GPT-3.5-Turbo, Llama2, and Mistral, which performs worst. Most models do well due to the multiple-choice question-answering nature of the task, which makes parsing outputs and evaluation simpler and increases the probability of success even for weaker models. (Figure \ref{fig:belebele}, Table \ref{tab:results_summary_belebele}).

\subsection{AfriQA}

GPT-4 has best performance, while the Llama2 and Mistral models perform very poorly on all languages. (Figure \ref{fig:afriqa}, Table \ref{tab:results_summary_afriqa}).

\subsection{IN22} 

We report our results on the IN22-Gen and IN22-Conv subsets (Figure \ref{fig:in22}) where we randomly select $k$ = 8 translation pairs from the development set of FLORES-200 \cite{nllbteam2022language} as in-context examples. We also report GPT-3.5-Turbo 0-shot and IndicTrans2 scores from \citet{gala2023indictrans} for comparison. For consistency, we use the \texttt{indic\_nlp\_library}\footnote{\url{https://github.com/anoopkunchukuttan/indic_nlp_library}} and the evaluation scripts\footnote{\url{https://github.com/AI4Bharat/IndicTrans2}} from \citet{gala2023indictrans} to tokenize the predictions and references before computing chrF++ \cite{popovic-2017-chrf} for Indic languages. We do not evaluate PaLM2 on this dataset, as most languages in this dataset are not supported by it.

Llama2 and Mistral perform poorly on all Indic languages in the En-Indic direction, whereas the performance is better on the Indic-En direction. Gemma-7B performs significantly better than both Llama2 and Mistral in both directions and on all languages. GPT-4 performs the best among all LLM models considered. All LLMs perform better in the Indic-En direction and Conversational dataset since they are finetuned with chat or conversational style data. We compare results to IndicTrans2 \citet{gala2023indictrans} and find that it fares significantly better than LLMs. (Figure \ref{fig:in22}, Tables \ref{tab:in22-conv-en-indic} - \ref{tab:in22-gen-indic-en}).

\subsection{XRiSAWOZ}

We compare DA accuracy of various models in Figure \ref{fig:xrisawoz_plot}. Table \ref{tab:results_xrisawoz} shows the comparison with fine-tuned models as well. We find that GPT-4's performance on DA accuracy is the closest and comparable to fine-tuned baselines for the task. Poorer scores on other models seem to correlate with the model's hallucination %or overgeneration
tendencies.% as described in Sec. 

We compare results on all 6 metrics in Table \ref{tab:detailed_xrisawoz} to better understand model behavior. We find that PaLM2,GPT-4 and Gemini-pro generate very concise responses leading to consistently higher BLEU scores as compared to other models. On all other metrics, GPT family of models significantly outperforms both PaLM/Gemini and open-source models. Notably, all the proprietary models achieve less than 10\% SER on Chinese hinting contamination of RiSAWOZ (the original Chinese-only dataset). Open source models often hallucinated non-existent entities in their responses while proprietary models did not show this tendency.

In the code-mixed English-Hindi setting, the performance is worse than both English and Hindi on average across most metrics for all models. (Figure \ref{fig:xrisawoz_plot}, Tables \ref{tab:results_xrisawoz}, \ref{tab:detailed_xrisawoz}). This could indicate challenges in understanding as well as generating effective code mixed text for all models. 
\subsection{MaRVL}

We evaluate LLaVA models, GPT-4-Vision\footnote{Given the API costs and constraints, we evaluate a random sample of 300 data instances per language.}, and Gemini-Pro-Vision on the multimodal datasets with monolingual and translate-test prompting (Figure \ref{fig:marvl}). The Azure BING translate module was utilized for translating the sentences into English. We find that accuracy scores border on random classification LLaVA models, with the lowest score on Tamil and Chinese. The translate-test strategy is comparable to monolingual. However, the performance is still the same as a random classification. GPT-4-Vision is significantly better than LLaVA, and the gains due to translate-test are only visible on Turkish. Gemini-Pro-Vision performs slightly better than random, and the translate-test is preferable except in the case of Chinese. (Figure \ref{fig:marvl}, Table \ref{tab:MARVL_results}).

\subsection{XM-3600}

We test the LLaVA models, GPT-4-Vision\footnote{Due to API costs and constraints, we evaluate a random sample of 488 data instances per language.}, and Gemini-Pro-Vision models on the XM-3600 image captioning dataset and use the chrF metric \cite{popovic-2015-chrf} to report the performance, unlike the original paper \cite{thapliyal2022crossmodal} that uses CIDEr. We see that the LLaVA models are poor for most languages that are not written in Latin script, especially Japanese, Korean, Russian, Thai, and Chinese. bakLLaVA-v1 performs much worse compared to LLaVA-v1.5-13B and ViP-LLaVA-13B (except English), and the latter two are comparable on all languages. Most Latin script high-resource languages such as French, German, Dutch, Spanish, and Italian outperform or come close to English performance, with lower-resource languages such as Danish, Czech Polish, and Norwegian performing worse. GPT-4-Vision significantly outperforms LLaVA models on all languages, however, the scores on Chinese, Japanese, and Thai are still very poor. French has the highest score followed by Italian, Spanish, and then English, which again shows that GPT-4-Vision is good at Latin script and European languages. Gemini-Pro-Vision is the second-best model on all languages, and the results follow the same trend as GPT-4-Vision. (Figure \ref{fig:xm3600}, Table \ref{tab:xm3600_results}).

\subsection{The deviation of performance across language families and tasks}

Given the experiments conducted, we look at how performance for a given Language Family or Task varies from the average performance (across the models covered in \systemname). In doing so we are interested in ranking how well models support different Language Families or Tasks.

The deviation for a given experiment $i$ in the Language Family or Task ($j$) is defined as:
$$
\Delta_{(i,j)}  = p\_score_{(i,j)} - \frac{1}{N} \sum_{i}^{N} {p\_score_{(i,j)}}
$$
Where $p\_score_{(i,j)}$ is the penalized score for the experiment $i$, and a high positive value indicates that a given subject (Language Family or Task) performs better than average where as a low negative value indicates that the subject performs lower than the average (across all models). $p\_score_{(i,j)}$ is calculated as: 
$$
p\_score_{(i,j)} = (\frac{|X_j|} {\sum_{i} |X_{j}|)}) * score_i
$$
Where $score_i$ is the normalized score for the experiment, penalized by the ratio of the instances in a given language family/task ($j$) to the total number of instances in all the language families/tasks.

Because of the sparsity in (Language, Dataset, Model) combinations (see Table \ref{tab:datasets}), we apply the size penalization to limit the bias of outliers and combinations with little support. For example, there are total of $320$ IE: Iranian Language family experiments in our data, with an average score of $0.31$, and a penalized score of $0.05$, compared to Basque which has $10$ experiments with an average score of $0.54$, but a penalized score of $0.003$.

\begin{figure*}
\centering
\begin{subfigure}[h]{\textwidth}
    \includegraphics[width=\textwidth]{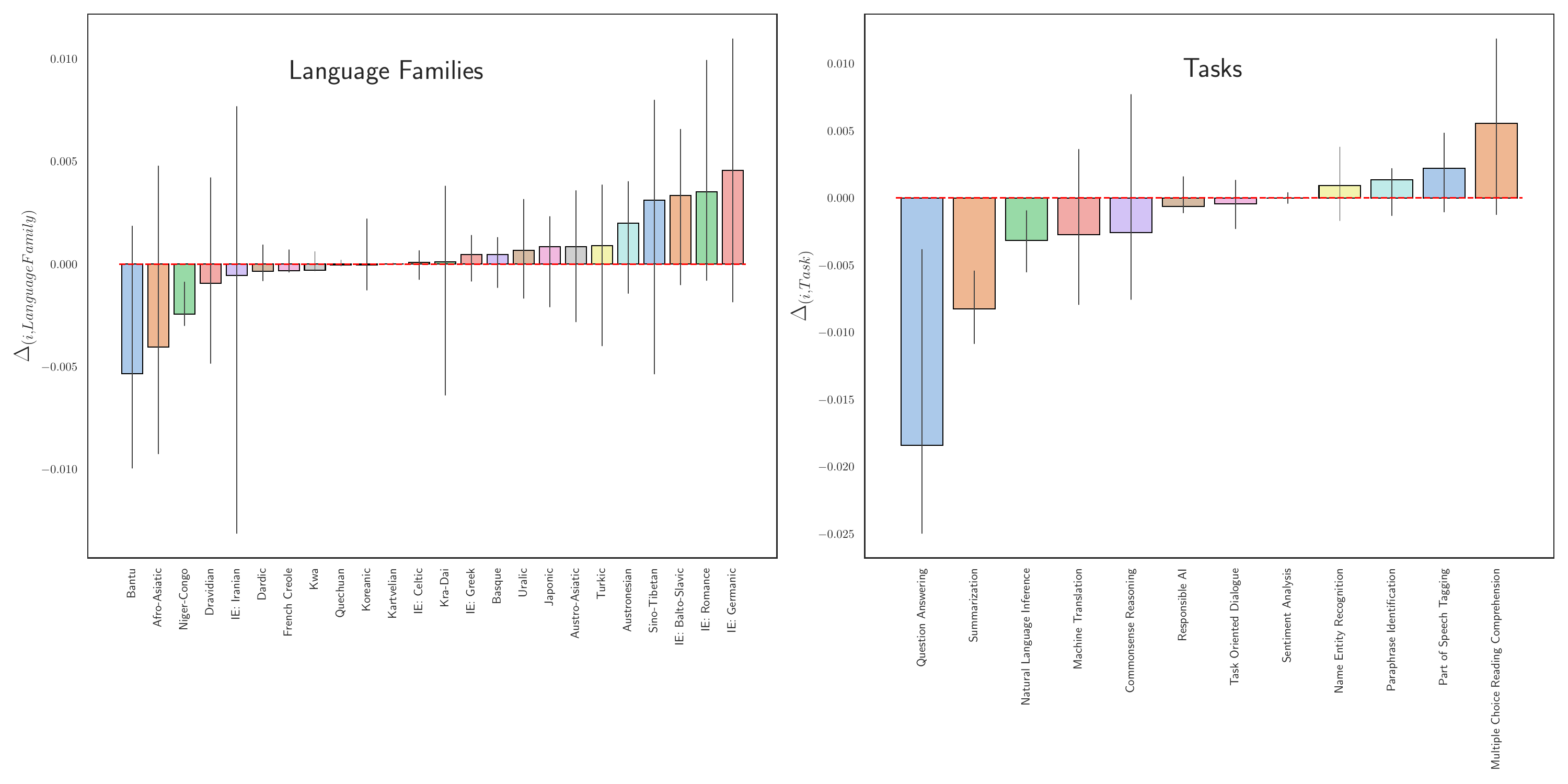}
\end{subfigure}

\caption{The positive scores of the bar-plots denote that the current LLMs are relatively good with those language families / tasks.}
\label{fig:easiness}
\end{figure*}

Figure~\ref{fig:easiness} gives the distribution of the $\Delta_{(i,j)}$ scores for Language Families and Tasks. We observe that languages in IE:Germanic Family, which ranks at the top, attain a significantly higher score that the mean, while at the the opposite end, Bantu and Afro-Asiatic languages significantly under-perform the mean across models and datasets. We also find that the models tested are significantly better at tasks such as MCQ Reading Comprehension and Parts of Speech Tagging (across all languages), than more open tasks such as Q\&A and text Summarization.

%% file: content/contamination.tex
\section{Contamination Analysis}

\label{sec:contamination}

\subsection{Commercial Model Contamination Study}
In our work, we follow the method described by \citet{golchin2023data} where we try to quantify contamination for commercial models such as PaLM2 and GPT-4. First, we prompt the model to generate three perturbations of the test set data points. Next, we provide these perturbations appended with the original text as four options to the model, and prompt it to pick a preferred option. We measure contamination as the chance adjusted accuracy using Cohen's Kappa ($\kappa$) and account for LLM's position bias towards a particular option by adjusting the calculation of $\kappa$, called $\kappa_{fixed}$.

We study contamination on GPT-4 and PaLM2 for 5 datasets: PAWS-X, UDPOS, TyDiQA, XNLI, and XCOPA, on 100 data points per language in each dataset. Our results show that all datasets are highly contaminated except for UDPOS, and for all datasets, contamination is higher for GPT-4, than for PaLM2. Contamination values for all datasets across different languages are reported in Appendix \ref{sec:appendix_contamination}. Contamination values differ significantly across languages for the same dataset, which could be due to bad perturbations generated by models owing to their varying performance in different languages. Another limitation of this approach is that \citet{golchin2023data} study position bias only for GPT models and append the original text as the fourth option based on their observations. However, this could vary for different models.

\subsection{Open-Source Model Contamination study}
We follow the Black Box test for contamination study of open-source model described by \citet{oren2023proving}. This test is statistical test which provides provable guarantees that a given test set is contaminated. To achieve these guarantees, they exploit the fact that many datasets have a property known as \textit{exchangeability}, where the order of examples in the dataset can be shuffled without affecting its join distribution. If a model has seen a benchmark dataset, it will have a preference for the canonical order (i.e. the order that examples are given in the public repositories) over randomly shuffled example orderings. If the difference between the said canonical order and the shuffled order is statistically significant, then the dataset is considered to be contaminated according to this method.

We conducted tests on the 7B instruction-tuned variants of Llama2, Mistral, and Gemma across the following evaluation datasets: PAWS-X, XCOPA, XNLI, XQUAD, XRiSAWOZ, and XstoryCloze. The significance level for our analysis was set at $0.001$. We observed (Table \ref{tab:opensource_contamination}) that all models, except for the Gemma base model, exhibited contamination. Specifically, datasets such as PAWS-X, XCOPA, XQUAD, and XRiSAWOZ were found to have their p-values less than the significant value for Gemma 7B Instruct, Llama2 7B Instruct and Mistral 7B Instruct indicating contamination.

%% file: content/discussion.tex
\section{Discussion}
\label{sec:discussion}

In this work, we benchmark 22 datasets covering 83 languages across several models -- GPT-3.5-Turbo, GPT-4, PaLM2, Gemini-Pro, Gemma, Llama2, Mistral as well as multimodal models. We find similar trends across most datasets we study - larger commercial models such as GPT-4 and Gemini-pro outperform smaller models like Gemma, Llama and Mistral models, particularly on low-resource languages. This suggests that multilingual performance is a challenge for smaller models, and directions such as language-specific models, language family-based models and fine-tuning should be explored for better multilingual performance. 

GPT-4, PaLM2 and Gemini-Pro excel on different datasets, with GPT-4 showing superior performance overall on multilingual datasets compared to both PaLM2 and Gemini-Pro. GPT-4-Vision outperforms LLaVA and Gemini-Pro-Vision on the multimodal datasets we study. Tokenizer fertility is correlated with Language Model performance \cite{rust-etal-2021-good, ali2023tokenizer}. We plot the fertility analysis of all the tokenizers (Figure: \ref{fig:fertility}) for the models that we studied in this work. We noticed that on average, Latin script languages such as Spanish, English had lower fertility as compared to languages that are morphologically complex languages like Telugu, Malay and Malayalam having high fertility amongst all the tokenizers.

Dataset contamination is a critical issue that affects English and non-English language benchmarking studies. Our contamination analysis on open source and commercial models shows that almost all models are contaminated with datasets included in \systemname. New multilingual evaluation datasets are difficult to create due to resource and funding constraints, hence, care should be taken to make sure that they are not included in the training data of LLMs. To achieve this, we need to enhance our ability to identify instances of contamination, as well as implement measures to avoid future contamination.

%% file: content/limitations.tex
\section{Limitations}
\label{sec:limitations}

Our work is subject to the following limitations: 

\paragraph{Model comparison} We have covered a wide array of Large Language Models. We realize that access to the commercial models (GPT, PaLM2, etc.) is via an API endpoint. These models might be running various post-processing modules and classifiers resulting in an inflated performance as compared to the Open-source models (LLaVA, Llama, Mistral).

\paragraph{Dataset contamination} We perform the dataset contamination exercise on a few set of datasets for PaLM2 and GPT-4 on a granular level. We also perform a thorough analysis of the open-source models covered in \systemname. However, there were certain limitations that we discuss in depth in Section \ref{sec:contamination}. We were also limited by the compute and time, therefore we did not perform the contamination study on all our datasets and only covered the 7B variants of our open-source models.

\paragraph{Prompt tuning} LLMs are sensitive to prompting, and we do not perform extensive prompt tuning for the new datasets. We also do not experiment with prompting variations, such as translate-test and zero-shot cross-lingual prompting, or more complex strategies such as Chain of Thought prompting due to resource constraints. 

\paragraph{Experiments on limited data and datasets} Due to resource constraints, we perform experiments on partial datasets when indicated, and do not evaluate all models on all datasets. We plan to do so in future work.

\paragraph{Focus on task accuracy} We perform limited experiments on RAI datasets and do not perform experiments on other important dimensions such as fairness, bias, robustness, efficiency, etc., mainly due to the lack of such datasets for non-English languages. This is an important future research direction.

%% file: appendix/tasks_and_datasets.tex
\section{Appendix}
\label{sec:appendix}

\subsection{Tasks and Datasets}
We benchmark 22 datasets encompassing 83 languages. A breakdown of this is described here in Table \ref{tab:datasets}
\begin{table}[h]
     \centering
     \tiny
     \begin{tabular}{lll}
     \toprule
    Dataset&Task&Languages\\
    \midrule
    XNLI&Natural Language Inference& 15\\
    Indic-XNLI&Natural Language Inference&11\\
    GLUECoS&Natural Language Inference&2\\
    PAWS-X&Paraphrase Identification&7\\
    XCOPA&Commonsense Reasoning& 10\\
    XStoryCloze&Commonsense Reasoning& 11 \\
    TyDiQA-GoldP&Question Answering& 9 \\
    MLQA&Question Answering&6\\
    XQuAD&Question Answering& 11\\
    IndicQA&Question Answering& 10\\
    AfriQA&Question Answering& 10\\
    MaRVL&Visual Question Answering&5\\
    UDPOS&Part of Speech Tagging& 38\\
    PANX&Name Entity Recognition&48\\
    XRiSAWOZ&Task Oriented Dialogue&6\\
    WinoMT&Responsible AI& 8 \\
    GLUECoS&Sentiment Analysis& 2 \\
    Jigsaw&Toxicity Classification& 6\\
    XLSum&Summarization&  44\\
    IN22&Machine Translation&14\\
    XM-3600&Image Captioning&20\\ 
    BeleBele&Multiple Choice Reading Comprehension&23\\
    
  \bottomrule
     \end{tabular}
     \caption{Dataset and language coverage}
     \label{tab:datasets}
     \vspace{-0.4cm}
 \end{table}

\subsection{Prompts}
\label{sec:prompt_templates}
Figures \ref{fig:AfriQA prompt} to \ref{fig:RG} shows the various prompts used in our benchmarking study.

\subsection{Results for Fertility Analysis}
Figure \ref{fig:fertility} shows fertility analysis.

% \subsection{Examples for Caption Generation}
% Figure \ref{fig:vision_visualization} shows an example for caption generation.

 \subsection{Results - Figures}
Figures \ref{fig:afriqa} to \ref{fig:xlsum} show our results on various models, languages, and datasets.

\subsection{Results - Tables}
Tables \ref{tab:results_summary_xnli} to \ref{tab:detailed_xrisawoz} show our results on various models, languages, and datasets.

\subsection{Contamination}
\label{sec:appendix_contamination}
Tables \ref{tab:cont_tydiqa} to \ref{tab:cont_xnli} show the contamination values for the various datasets for the commercial models. For the p-values of the statistic test performed on the open-source models, please refer to Table \ref{tab:opensource_contamination}.

%% file: appendix/prompts.tex
% \subsection{Prompts}
% \label{sec:prompt_templates}

\begin{figure}[!h]
\centering
\begin{promptbox}
\justify
\noindent Task Instruction $\mathcal{I}$: You are an NLP assistant trained to answer questions directly. For each question provided, respond with the most accurate and concise answer. The answer should be in the same language as the question.\\[5pt]
\noindent Template $f_{temp}$:\\
    Q: \texttt{\{question\}}\\ 
    A: \texttt{\{answer\}}

\end{promptbox}
\caption{AfriQA Prompt}
\label{fig:AfriQA prompt}
\end{figure}

\paragraph{Prompt:} For Belebele fig: \ref{fig:Belebele MRC prompt}, we evaluated our models on zero-shot prompting using instructions proposed by \citet{bandarkar2023belebele} 
\footnote{\url{https://github.com/EleutherAI/lm-evaluation-harness/pull/885}}.\\
For chat-based (e.g. Llama2 chat) models and the X-RiSAWOZ prompt (fig: \ref{fig:xRiSAWOZ_prompt}), we drop the ``Learning example$\dots$'' and ``Target example$\dots$'' and use the ChatGPT-like prompt format with task prompt in the ``system'' prompt, \{Turn ID, Database, Context\} in the ``user'' prompt and ``Answer'' in the ``assistant'' prompt. We use the dataset provided by \citet{moradshahi-etal-2023-x} in which the context is preprocessed to include all the relevant information (e.g. previous dialogue acts or states) for a task.
\begin{figure}[!h]
\centering
\begin{promptbox}
\justify
\noindent Task Instruction $\mathcal{I}$:You are an AI assistant whose purpose is to perform reading comprehension task. Given the following passage, query, and answer choices, output the letter corresponding to the correct answer. \\[2pt]

\noindent Template $f_{temp}$:\\
\texttt{\{instruction\}}\\
\#\#\# \\
Passage: \\
\texttt{\{passage\}} \\ 
\#\#\# \\
Query: \\
\texttt{\{query\}} \\
\#\#\# \\
Choices: \\
(A) \texttt{\{A\}} \\
(B) \texttt{\{B\}} \\
(C) \texttt{\{C\}} \\
(D) \texttt{\{D\}} \\
\#\#\# \\
Answer:

\end{promptbox}
\caption{Belebele MRC Prompt}
\label{fig:Belebele MRC prompt}
\end{figure}

\begin{figure}[!h]
\centering
\begin{promptbox}
You are an AI assistant whose purpose is to perform translation. Given the following sentence in \{source\}, translate it to \{target\}.
\end{promptbox}
\caption{Translation Prompt}
\label{fig:translation_prompt}
\end{figure}

\begin{figure}[!h]
\centering
\begin{promptbox}
Is the below statement in \{language\} correct with respect to the left and right images? Return `TRUE` if it is true, else `FALSE`.

CAPTION: \{caption\}
\end{promptbox}
\caption{MaRVL Prompt}
\label{fig:marvl_prompt}
\end{figure}

\begin{figure}[!h]
\centering
\begin{promptbox}
Generate a **brief** coco style caption for the given image in \{language\}.
\end{promptbox}
\caption{XM-3600 Prompt}
\label{fig:marvl_prompt}
\end{figure}

\begin{figure}[!h]
\centering
\begin{promptbox}
\justify
\noindent $\langle$ TASK PROMPT. Refer to each task below. $\rangle$ \\
\{ \\
Learning example \#i: \\
Turn ID: turn\_id \\
Database: db\_id \\
Context: gold\_context \\
Answer: gold\_answer \\
\} for i in range(k) \# (in-context examples) \\
Target example \#i: \\
Turn ID: turn\_id \\
Database: db\_id \\
Context: gold\_context \\
Answer: $\langle$model-completion-here$\rangle$
\end{promptbox}
\caption{General prompt structure for X-RiSAWOZ}
\label{fig:xRiSAWOZ_prompt}
\end{figure}

\begin{figure}[!h]
\centering
\begin{promptbox}
\justify
You are a helpful NLP assistant solving the ``Task Oriented Dialogue" problem. In particular, you are solving the "Dialogue State Prediction" subtask. In Dialogue State Prediction, you must describe what is the state of the dialogue given the history using SQL-like structure. The syntax can be understood from the examples below. Based on the learning examples given below, complete the ``Answer" part of the target example. Do not print any additional information.
\end{promptbox}
\caption{Task prompt for ``DST'' subtask in X-RiSAWOZ}
\label{fig:DST}
\end{figure}

\begin{figure}[!h]
\centering
\begin{promptbox}
\justify
You are a helpful NLP assistant solving the ``Task Oriented Dialogue" problem. In particular, you are solving the "API Call Detection" subtask. In API call detection, your task is to identify whether the dialogue can be continued with whatever context we already have. "yes" here means that additional data must be queried using an API for continuing the dialog while "no" means that API call is not required. Based on the learning examples given below, complete the ``Answer" part of the target example. Do not print any additional information.
\end{promptbox}
\caption{Task prompt for ``API'' subtask in X-RiSAWOZ}
\label{fig:API}
\end{figure}

\begin{figure}[!h]
\centering
\begin{promptbox}
\justify
You are a helpful NLP assistant solving the ``Task Oriented Dialogue" problem. In particular, you are solving the "Dialogue Act Prediction" subtask. In Dialogue Act Generation, you must generate the next dialogue action based on the given context. This will be an SQL-like structure. The syntax can be understood from the examples below. Based on the learning examples given below, complete the ``Answer" part of the target example. Do not print any additional information.
\end{promptbox}
\caption{Task prompt for ``DA'' subtask in X-RiSAWOZ}
\label{fig:DA}
\end{figure}

\begin{figure}[!h]
\centering
\begin{promptbox}
\justify
You are a helpful NLP assistant solving the ``Task Oriented Dialogue" problem. In particular, you are solving the "Response Generation" subtask. In Response Generation, your task is to produce a natural language response from the chatbot given the context of the conversation. Based on the learning examples given below, complete the ``Answer" part of the target example. Do not print any additional information.
\end{promptbox}
\caption{Task prompt for ``RG'' subtask in X-RiSAWOZ}
\label{fig:RG}
\end{figure}

%% file: appendix/fertility.tex
\clearpage
% \subsection{Results for Fertility Analysis}
% Figure \ref{fig:fertility} shows figure for fertility analysis.

\begin{figure*}[h]
\centering

    \includegraphics[width=\textwidth]{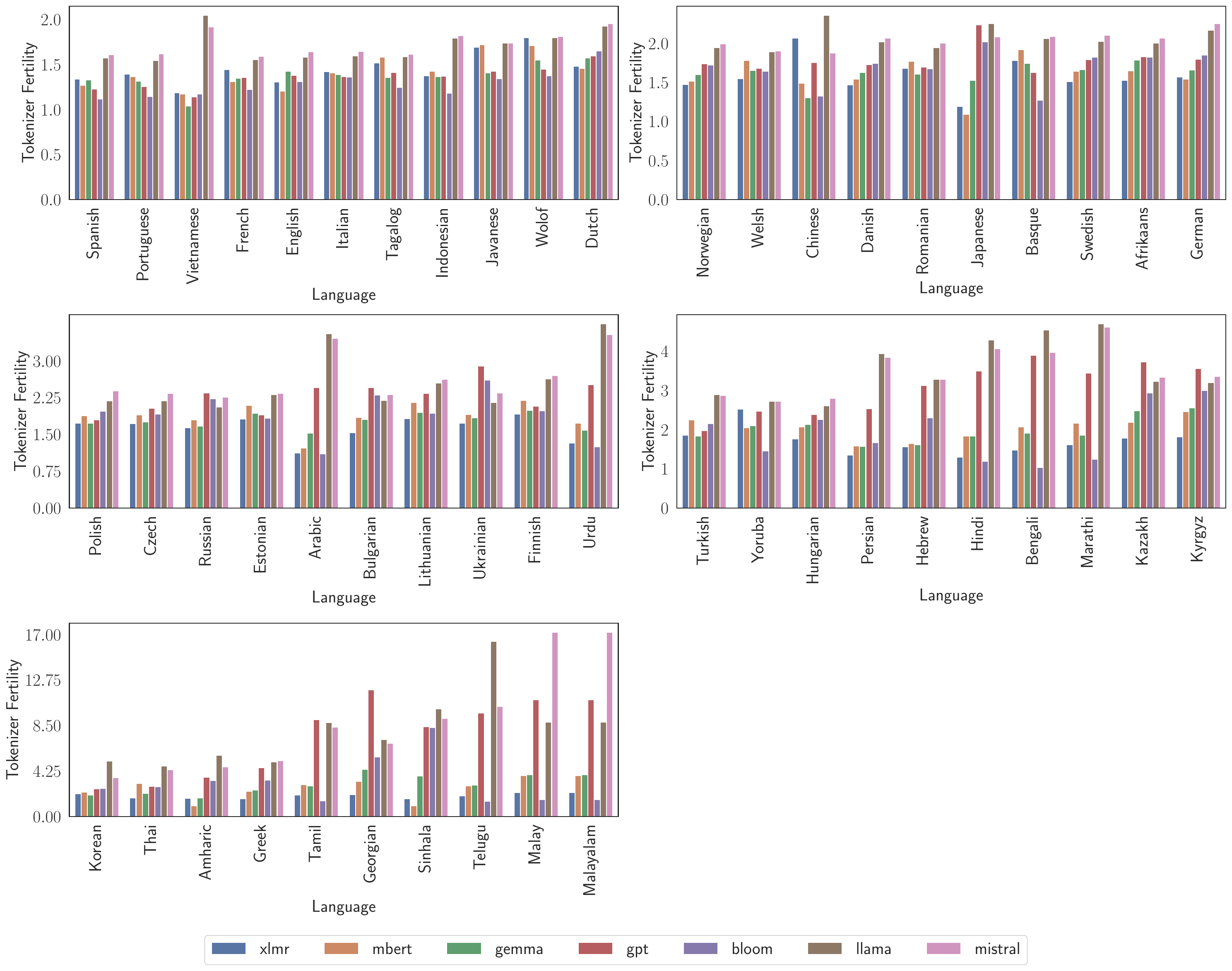}
    \caption{Fertility analysis was performed for all assessed models, with the exception of PaLM2 and Gemini, which was excluded due to a lack of available information about its tokenizer. \cite{acs:2019}}
    \label{fig:fertility}
\end{figure*}

%% file: appendix/vision_visualization.tex
% \begin{figure*}[]
% \centering
% \fbox{
%     \includegraphics[width=\textwidth]{figures/abc.drawio.pdf}}
%     \caption{Captions generated by GPT4-Vision and LLaVA-v1.5 on XM-3600. The top boxes correspond to GPT4-Vision and the bottom ones to LLaVA.}
%     \label{fig:vision_visualization}
% \end{figure*}

%% file: appendix/results_figures.tex
\clearpage
% \subsection{Results - Figures}
% Figures \ref{fig:afriqa} to \ref{fig:xlsum} show our results on various models, languages, and datasets.

\begin{figure*}[h]
\centering
\begin{subfigure}{\textwidth}
    \includegraphics[width=\textwidth]{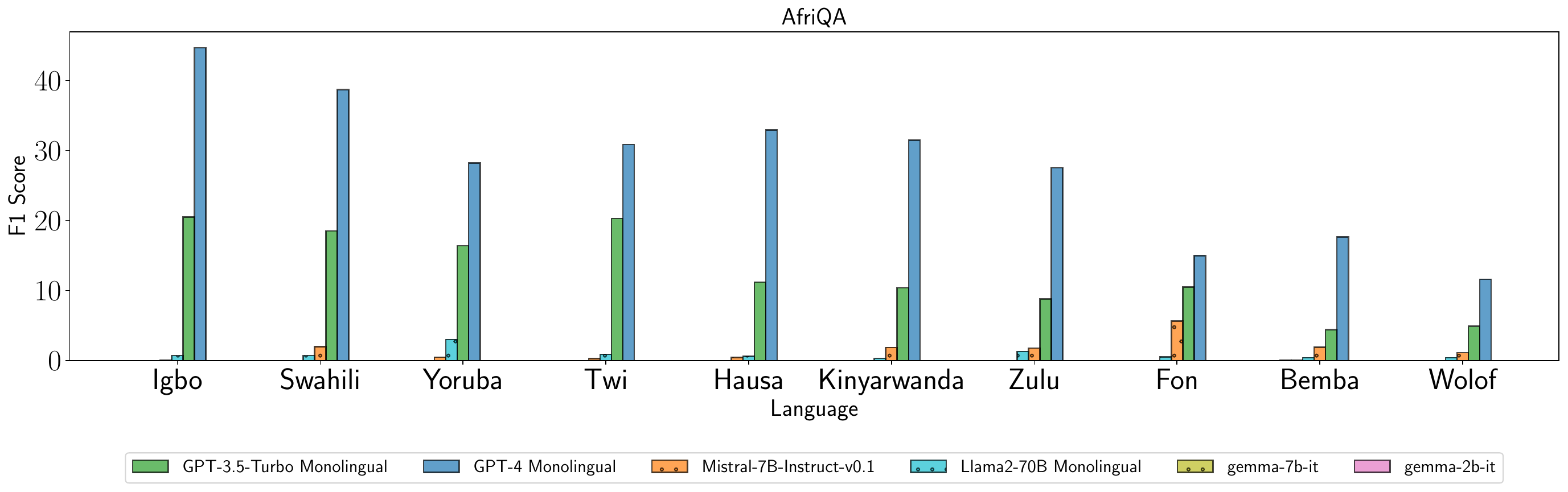}
\end{subfigure}

    \caption{Results for AfriQA across all languages and models for monolingual prompting}
    \label{fig:afriqa}
\end{figure*}

\begin{figure*}[h]
\centering
\begin{subfigure}{\textwidth}
    \includegraphics[width=\textwidth]{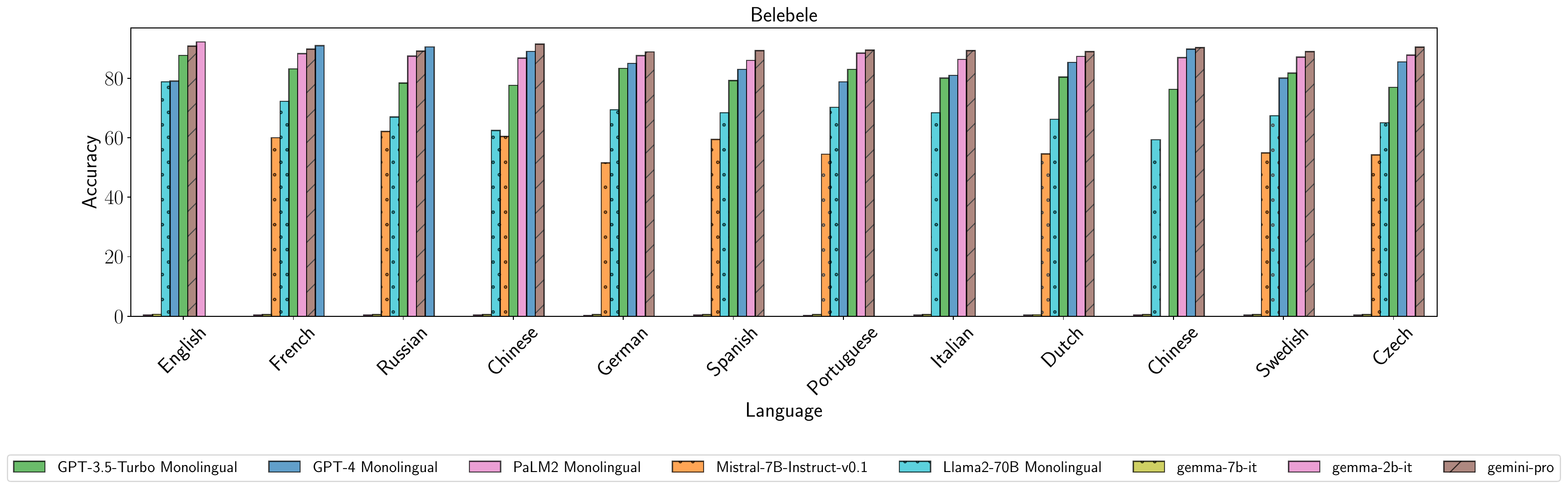}
\end{subfigure}
\begin{subfigure}{\textwidth}
    \includegraphics[width=\textwidth]{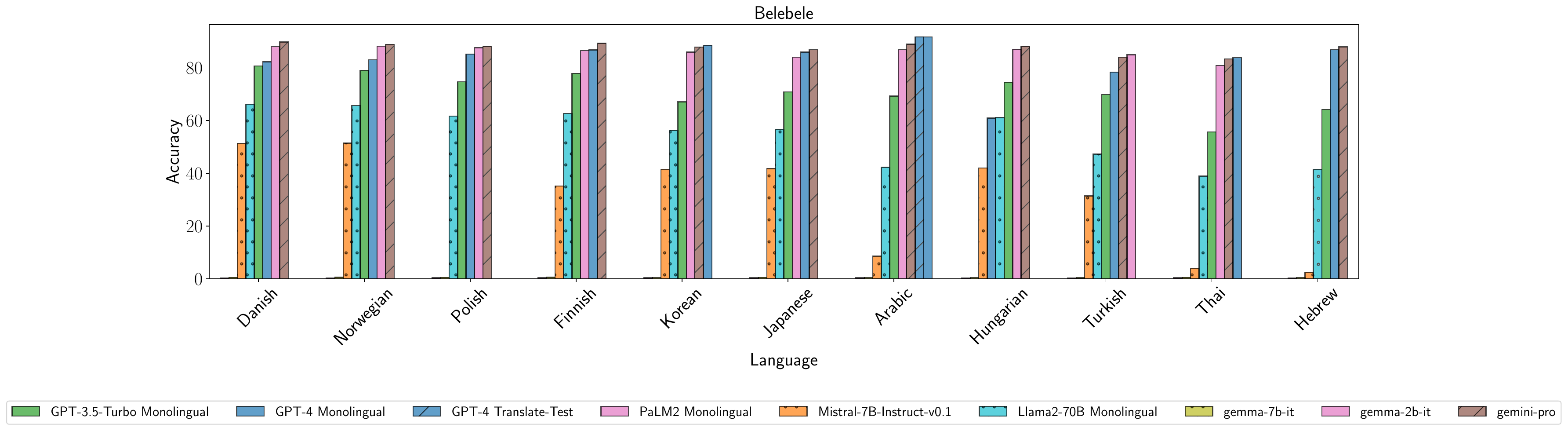}
\end{subfigure}
    \caption{Results for Belebele across all languages and models for monolingual prompting}
    \label{fig:belebele}
\end{figure*}

\begin{figure*}
\centering
    \includegraphics[width=\textwidth]{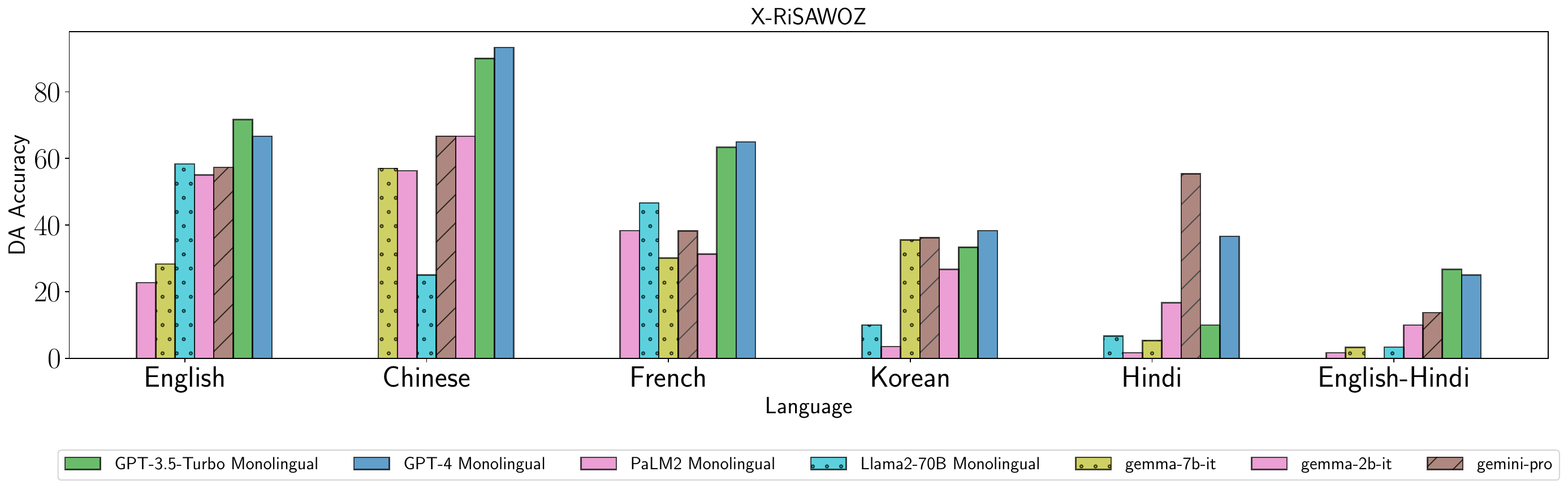}
    \caption{Results for X-RiSAWOZ across all languages and models for monolingual prompting}
    \label{fig:xrisawoz_plot}
\end{figure*}

\begin{figure*}
\centering
\begin{subfigure}{\textwidth}
    \includegraphics[width=\textwidth]{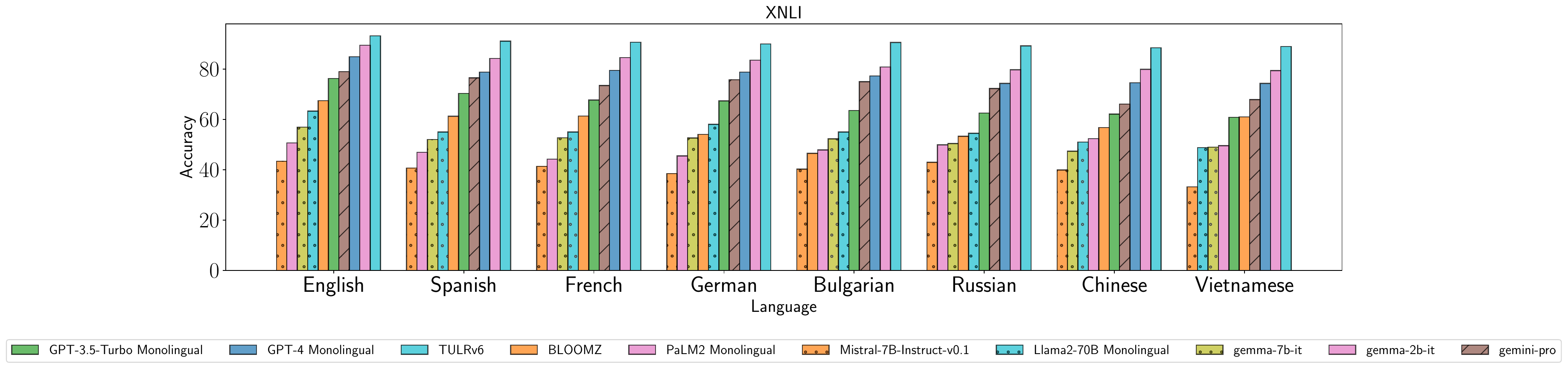}
\end{subfigure}
\begin{subfigure}{\textwidth}
    \includegraphics[width=\textwidth]{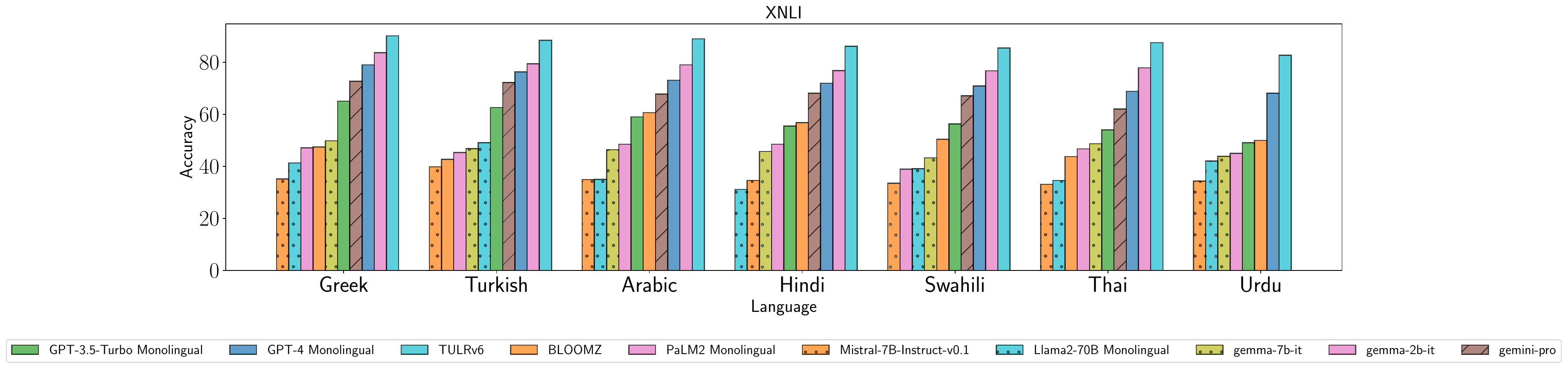}
\end{subfigure}
% \begin{subfigure}{\textwidth}
%     \includegraphics[width=\textwidth]{figures/XNLI_lang_nums_14.pdf}
% \end{subfigure}
    % \includegraphics[width=\textwidth]{figures/XNLI_lang_nums.pdf}
    \caption{Results for XNLI across all languages and models for monolingual prompting}
    \label{fig:xnli}
\end{figure*}

\begin{figure*}
\centering
\begin{subfigure}{\textwidth}
    \includegraphics[width=\textwidth]{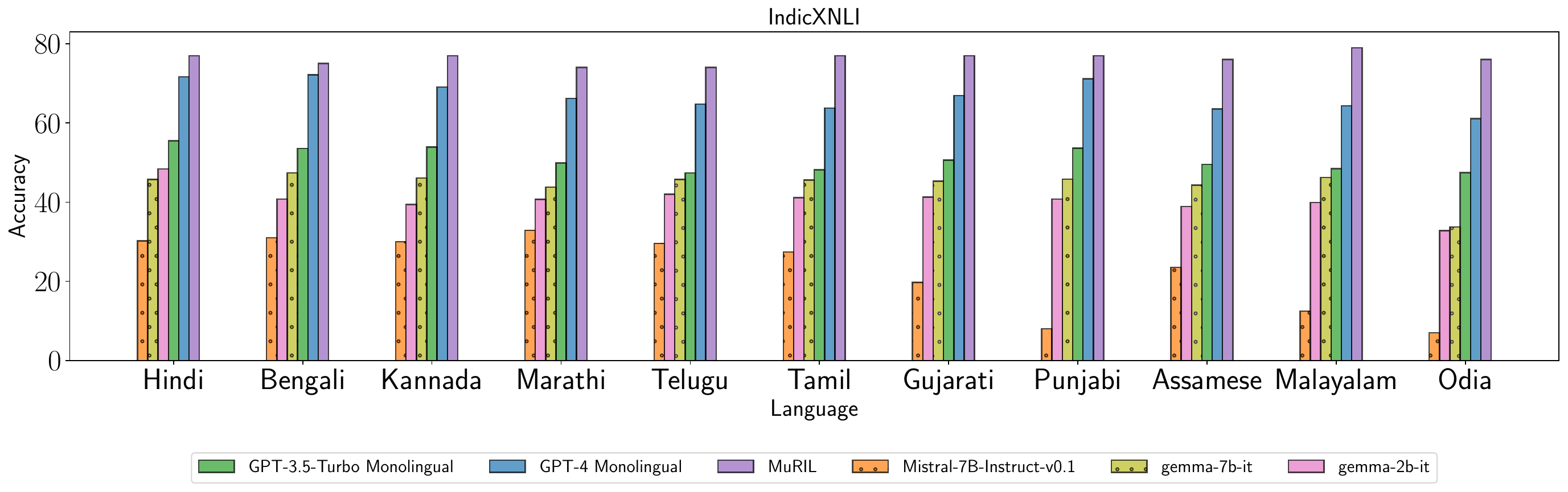}
\end{subfigure}
    \caption{Results for IndicXNLI across all languages and models for monolingual prompting}
    \label{fig:indicxnli}
\end{figure*}

\begin{figure*}
\centering
    \includegraphics[width=\textwidth]{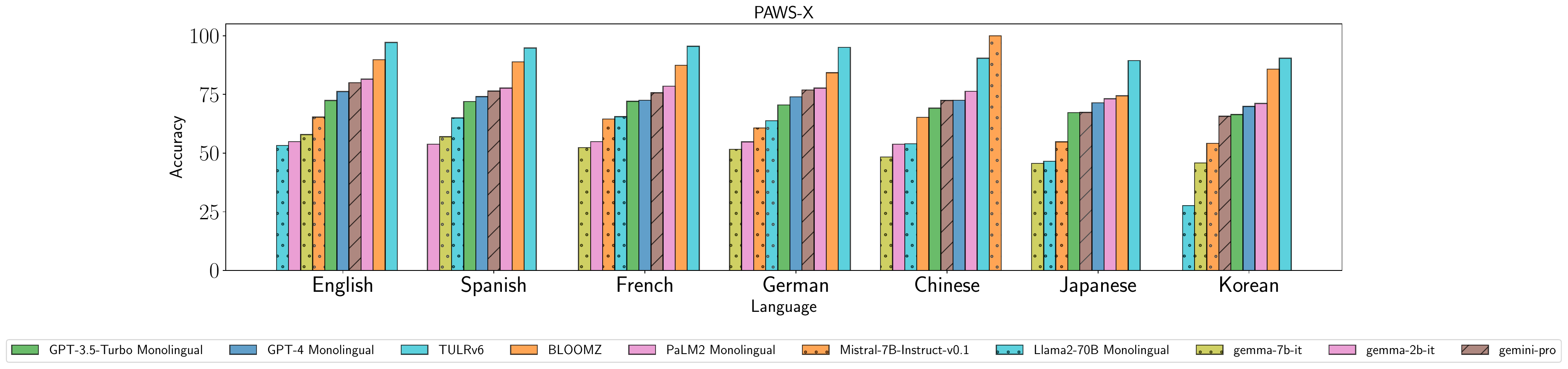}
    \caption{Results for PAWSX across all languages and models for monolingual prompting}
    \label{fig:pawsx}
\end{figure*}

\begin{figure*}
\centering
\begin{subfigure}{\textwidth}
    \includegraphics[width=\textwidth]{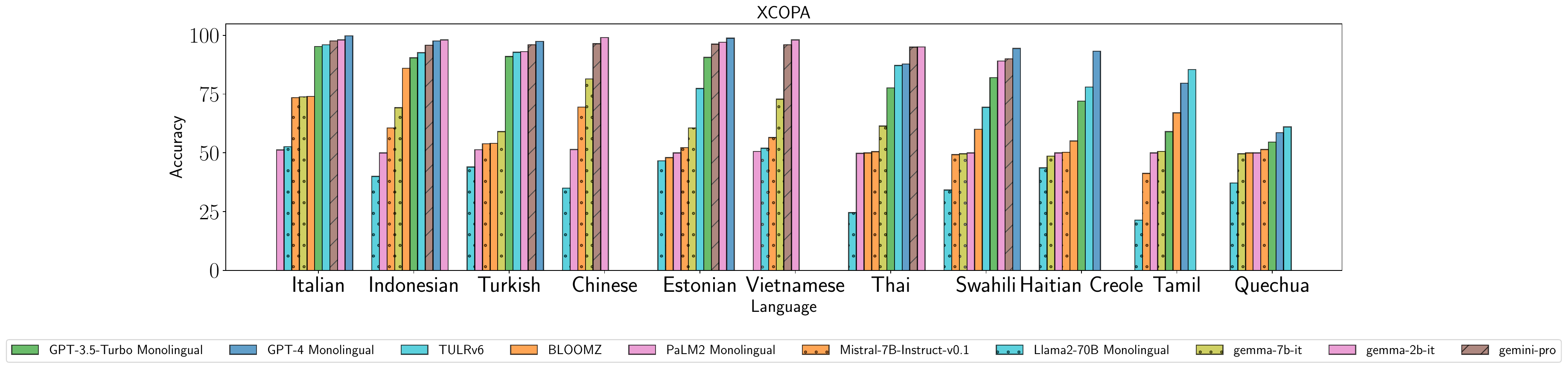}
\end{subfigure}

    \caption{Results for XCOPA across all languages and models for monolingual prompting}
    \label{fig:xcopa}
\end{figure*}

\begin{figure*}
\centering
\begin{subfigure}{\textwidth}
    \includegraphics[width=\textwidth]{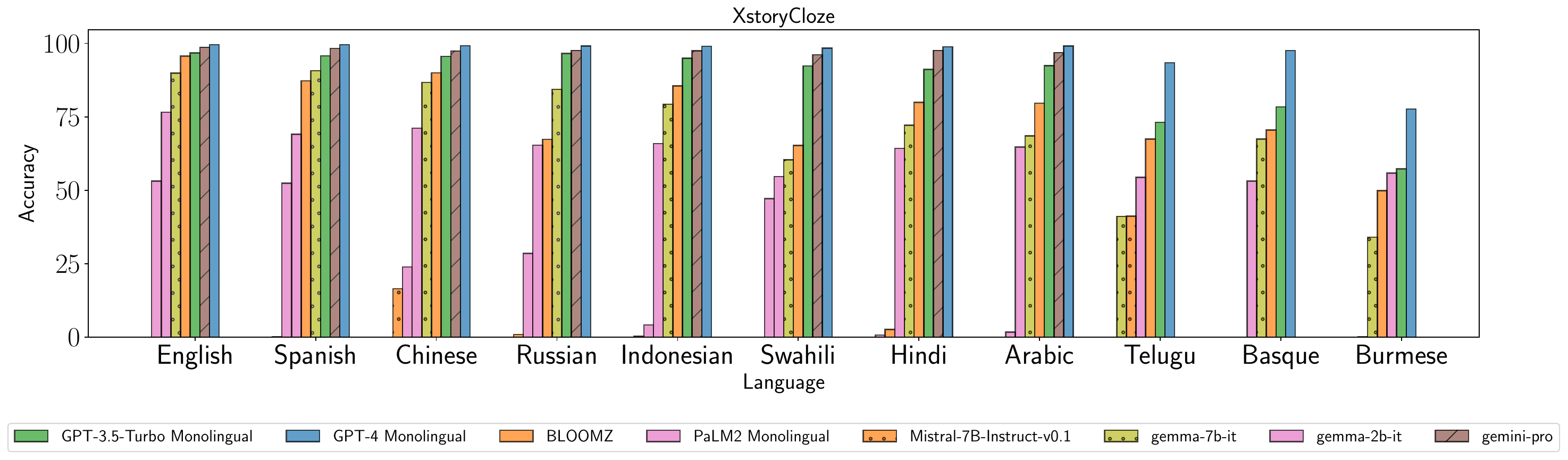}
\end{subfigure}
    \caption{Results for XStoryCloze across all languages and models for monolingual prompting}
    \label{fig:xstorycloze}
\end{figure*}

\begin{figure*}
\centering
\begin{subfigure}{\textwidth}
    \includegraphics[width=\textwidth]{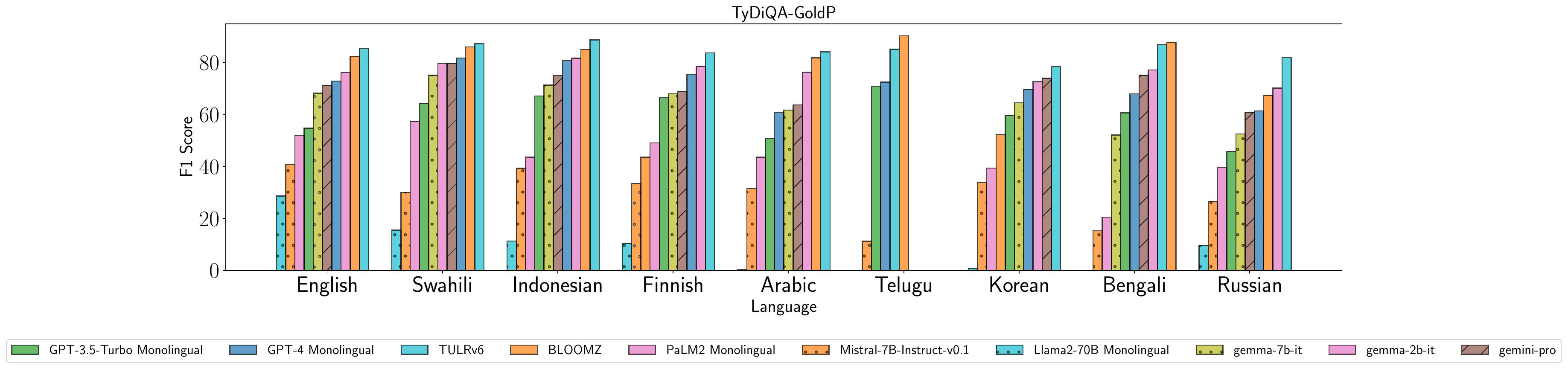}
\end{subfigure}
    \caption{Results for TyDiQA across all languages and models for monolingual prompting}
    \label{fig:tydiqa}
\end{figure*}

\begin{figure*}
\centering
    \includegraphics[width=\textwidth]{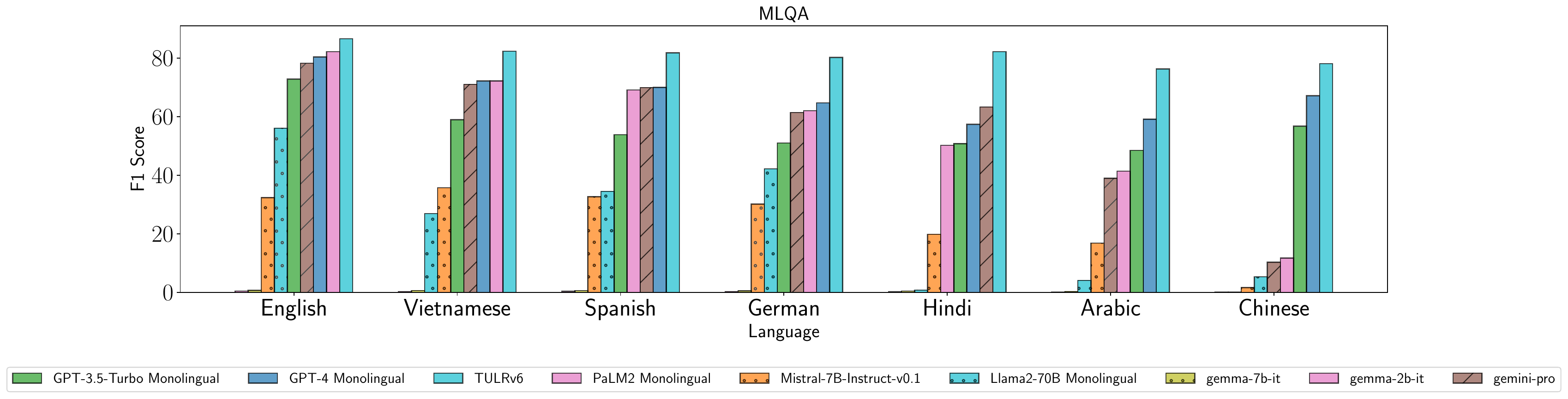}
    \caption{Results for MLQA across all languages and models for monolingual prompting}
    \label{fig:mlqa}
\end{figure*}

\begin{figure*}
\centering
\begin{subfigure}{\textwidth}
    \includegraphics[width=\textwidth]{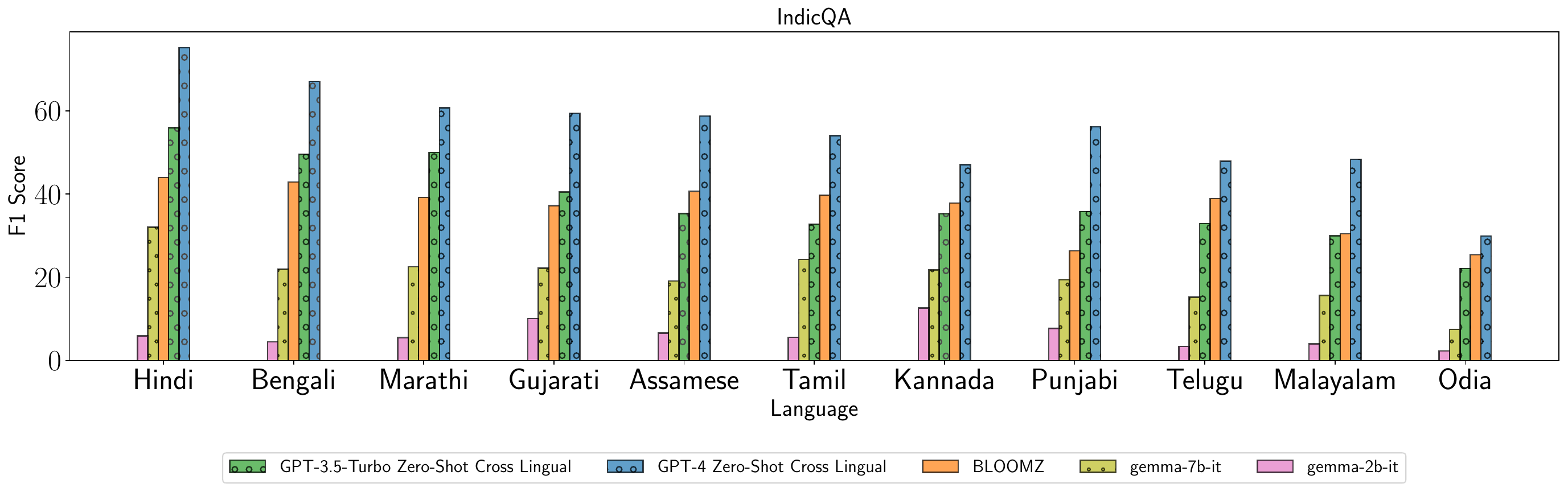}
\end{subfigure}

    \caption{Results for IndicQA across all languages and models with zero-shot cross-lingual prompting}
    \label{fig:indicqa}
\end{figure*}

\begin{figure*}
    \centering
    \begin{subfigure}{\textwidth}
        \includegraphics[width=\textwidth]{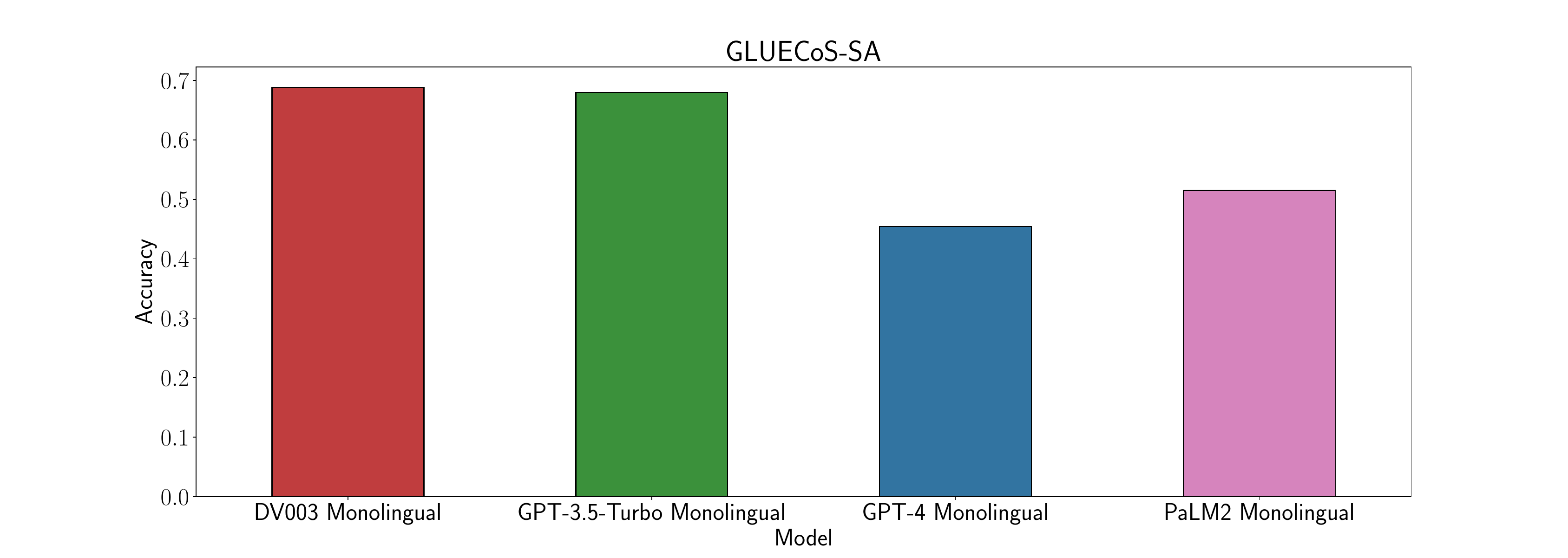}
    \end{subfigure}
    % \vspace{5em}
        \begin{subfigure}{\textwidth}
        \includegraphics[width=\textwidth]{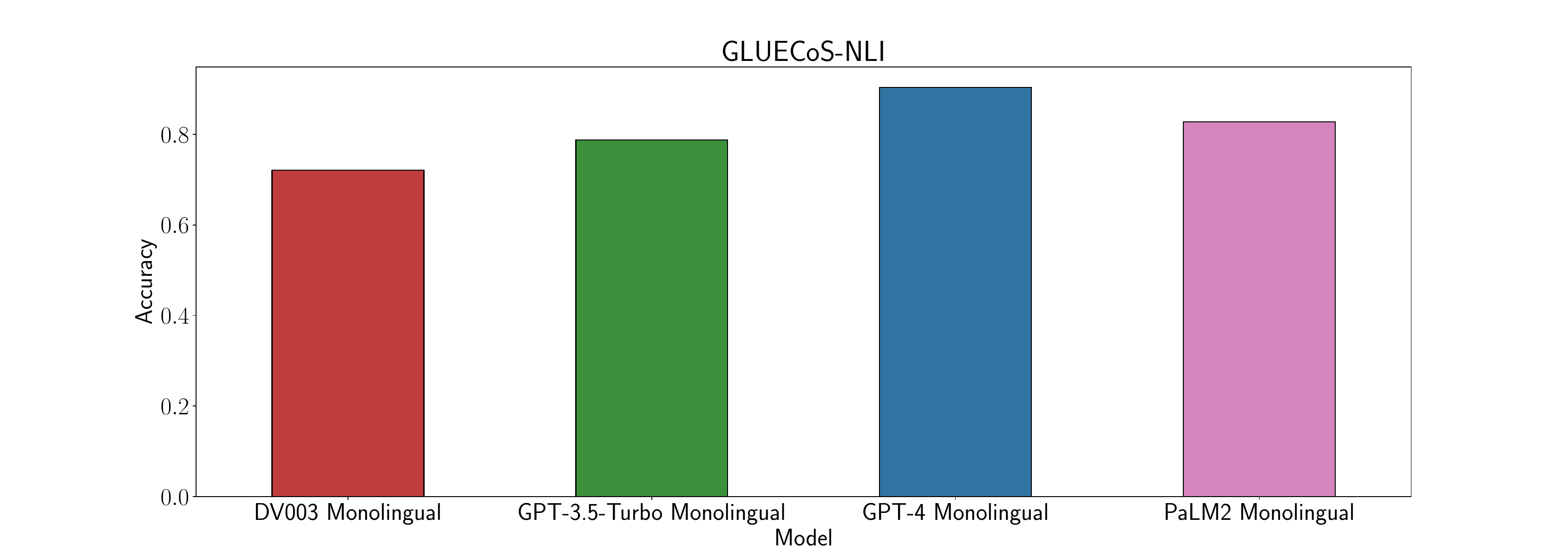}
    \end{subfigure}
    \caption{Results for the GLUECoS dataset on the Sentiment Classification (English-Spanish, En-Es-CS) and the NLI (English-Hindi) task}
    \label{fig:gluecos}
\end{figure*}

\begin{figure*}
\centering
    \includegraphics[width=\textwidth]{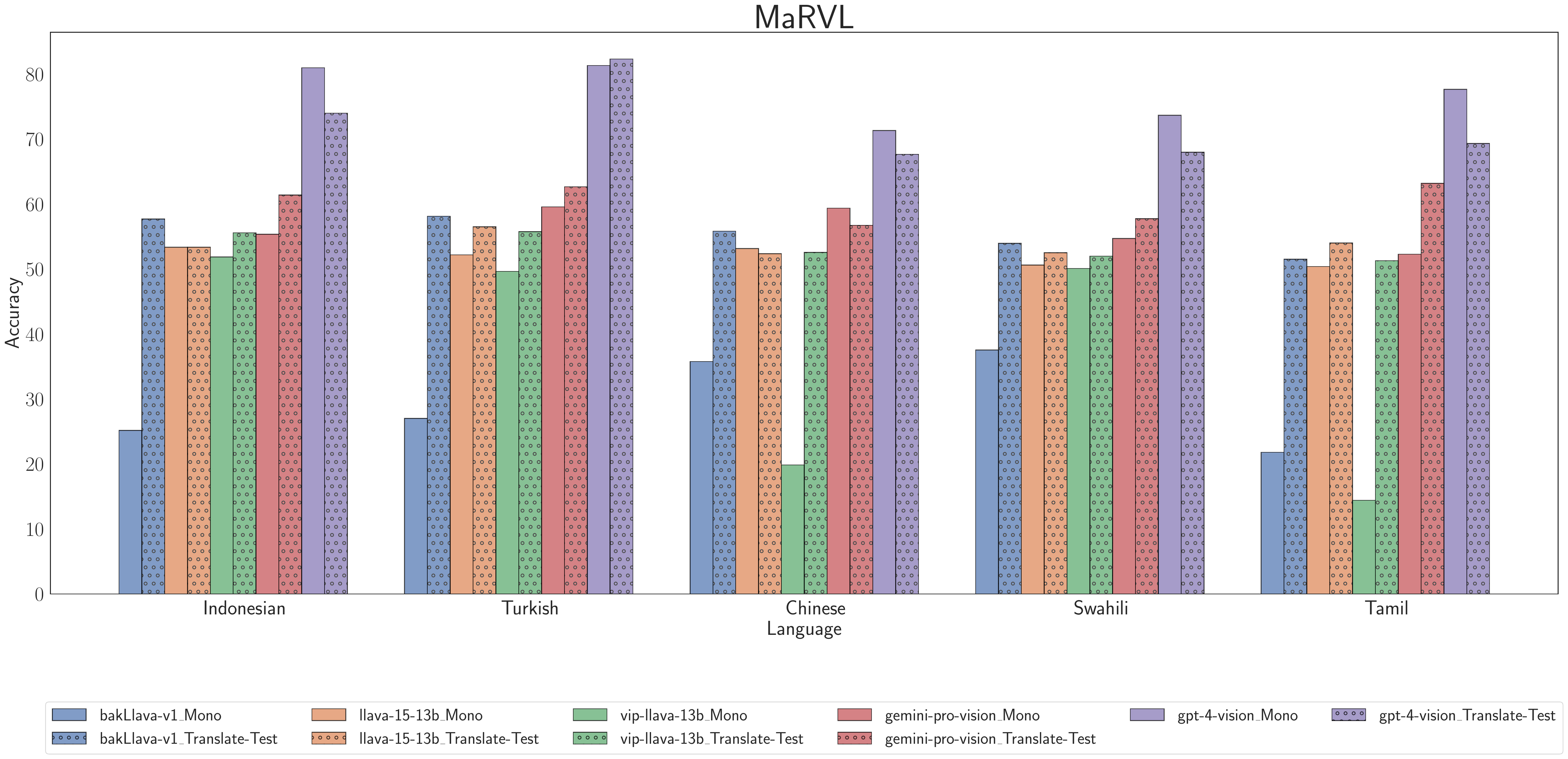}
    \caption{Accuracy scores for the LLaVA models, GPT4-Vision, and Gemini-Pro-Vision on MaRVL. We used two prompting strategies, monolingual and translate-test.}
    \label{fig:marvl}
\end{figure*}

\begin{figure*}
\centering
    \includegraphics[width=\textwidth]{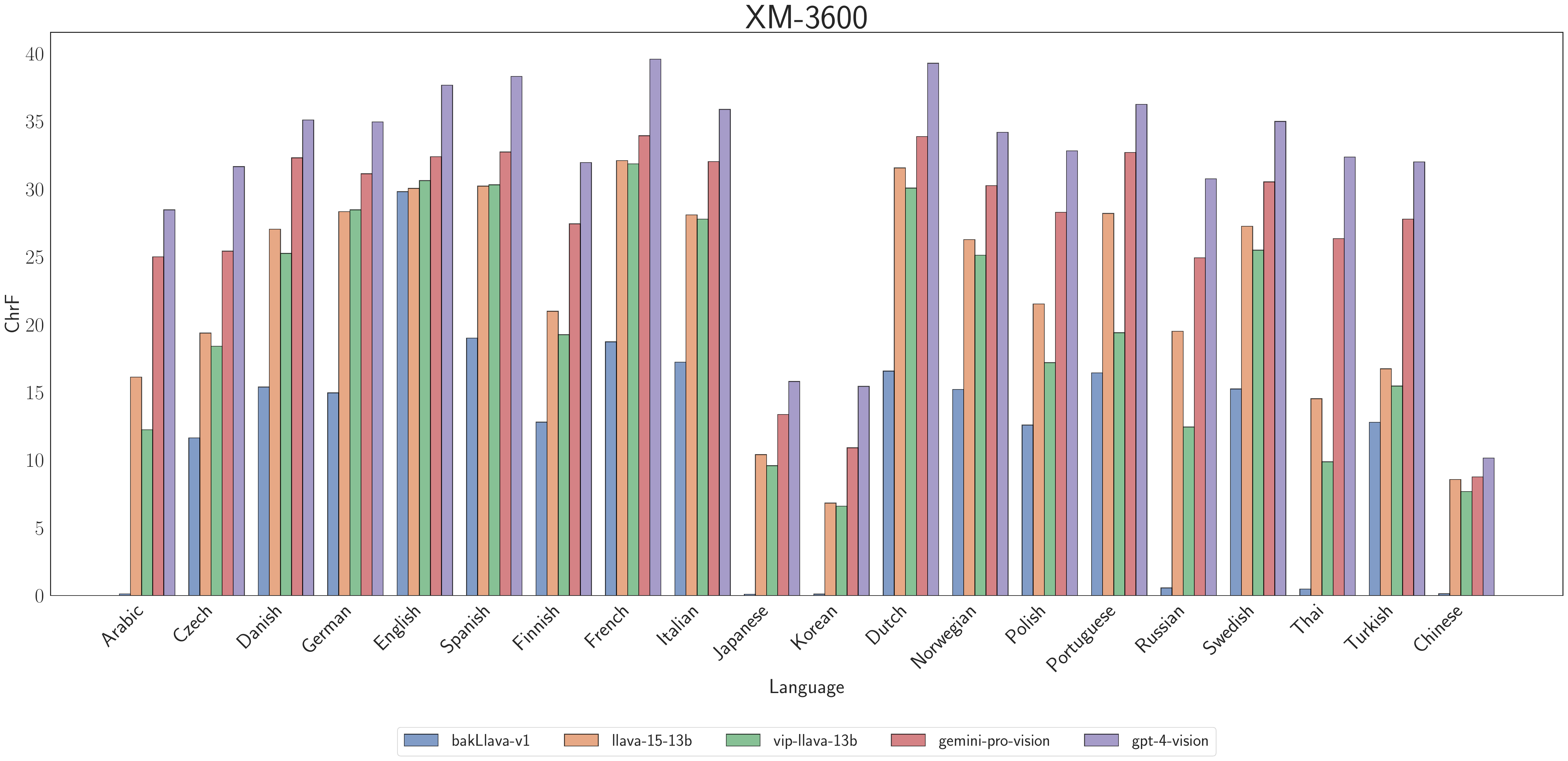}
    \caption{chrF scores for the LLaVA models, GPT4-Vision, and Gemini-Pro-Vision on XM-3600. We use monolingual prompting as the prompting strategy.}
    \label{fig:xm3600}
\end{figure*}

\begin{figure*}
\centering
% \begin{subfigure}{\textwidth}
        \includegraphics[width=\textwidth]{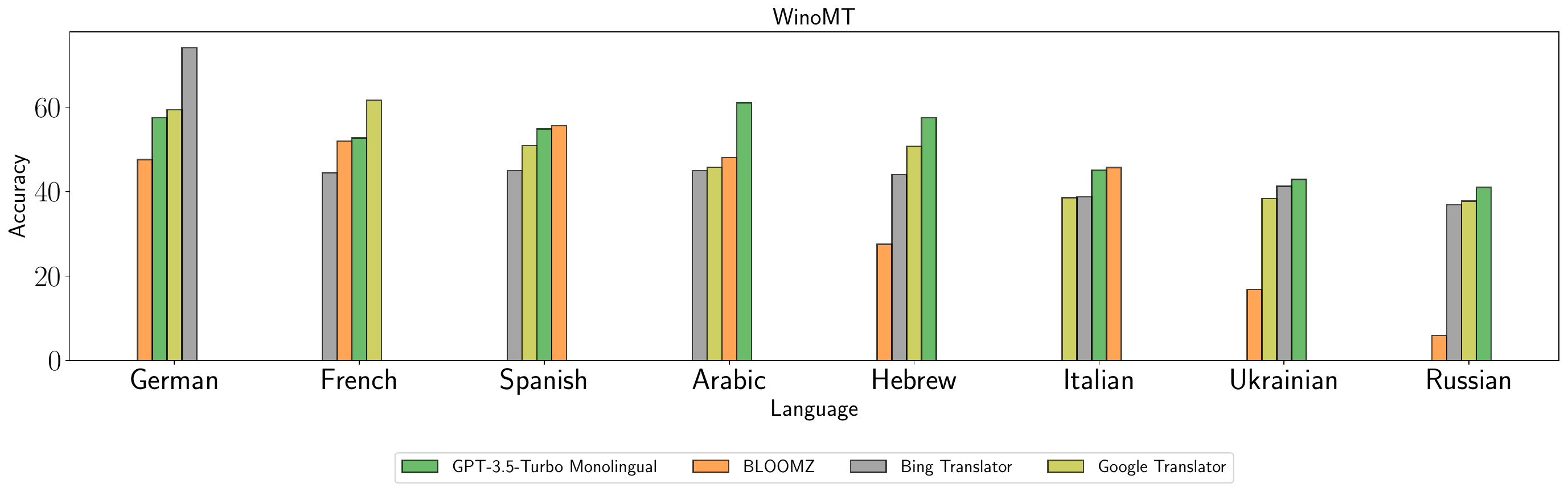}
    % \end{subfigure}
% \begin{subfigure}{\textwidth}
%         \includegraphics[width=\textwidth]{figures/WinoMT_lang_nums_0.pdf}
%     \end{subfigure}

%     \begin{subfigure}{\textwidth}
%         \includegraphics[width=\textwidth]{figures/WinoMT_lang_nums_7.pdf}
%     \end{subfigure}
    \caption{Results for WinoMT across all languages and models for monolingual prompting}
    \label{fig:winomt}
\end{figure*}

\begin{figure*}
\centering
\includegraphics[width=\textwidth]{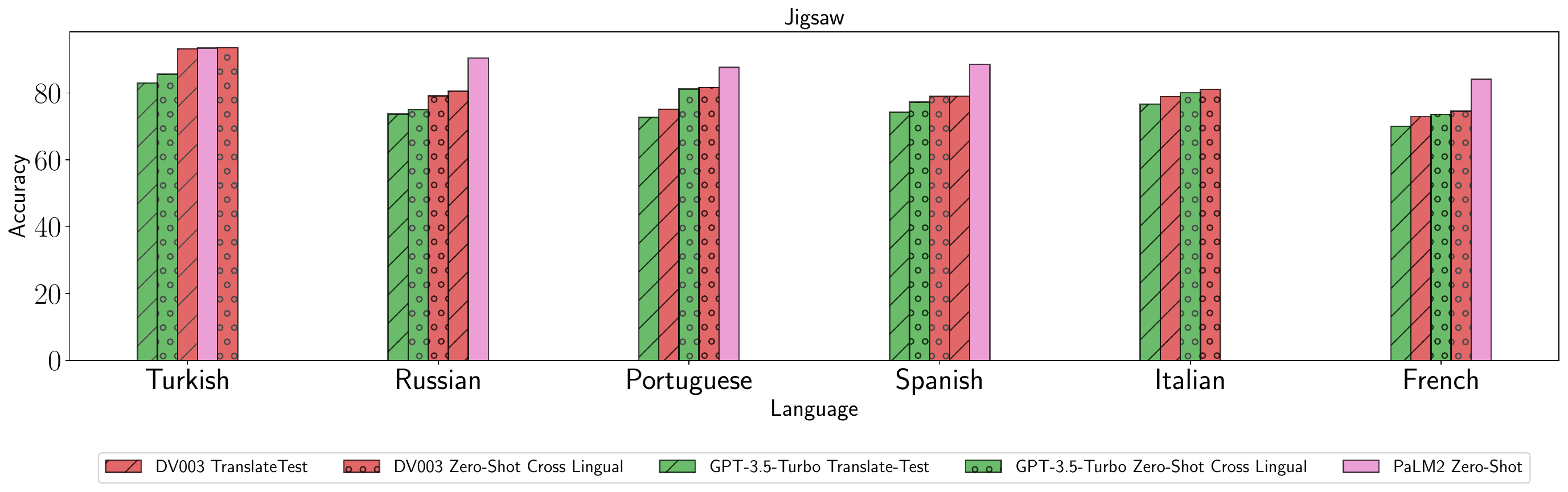}
    \caption{Results for Jigsaw across all languages and models for monolingual prompting}
    \label{fig:jigsaw}
\end{figure*}

\begin{figure*}
\centering
\begin{subfigure}{\textwidth}
\includegraphics[width=\textwidth]{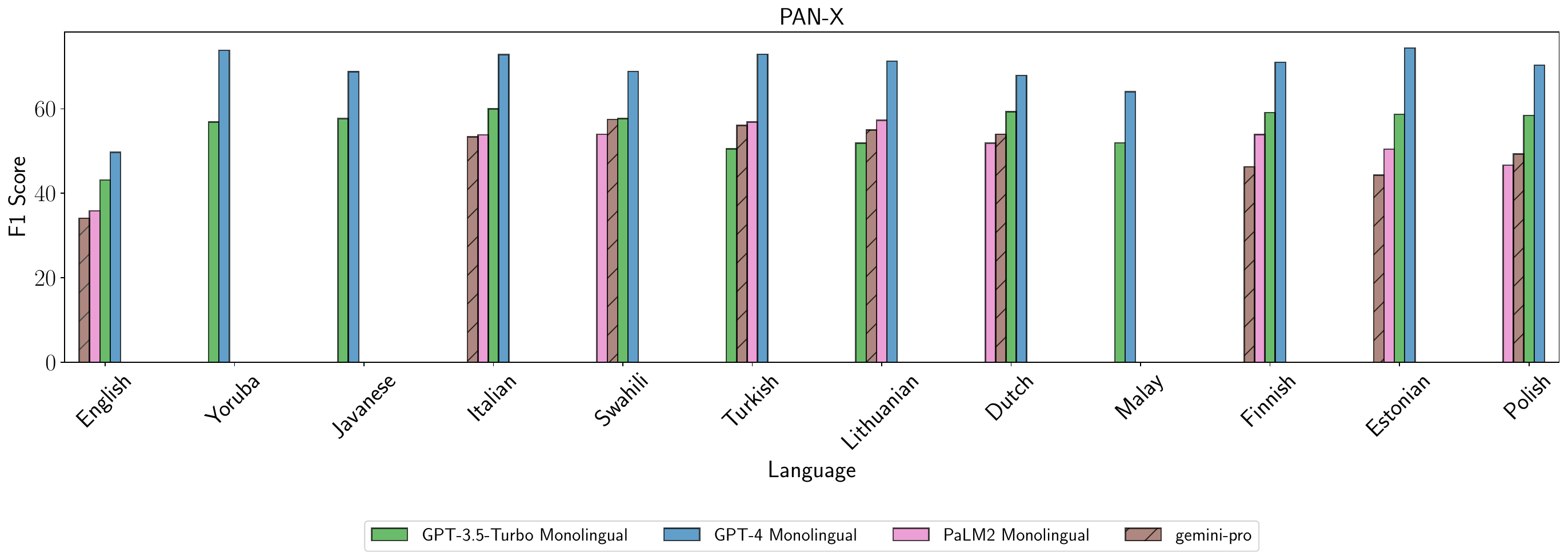}
\end{subfigure}
\begin{subfigure}{\textwidth}
\includegraphics[width=\textwidth]{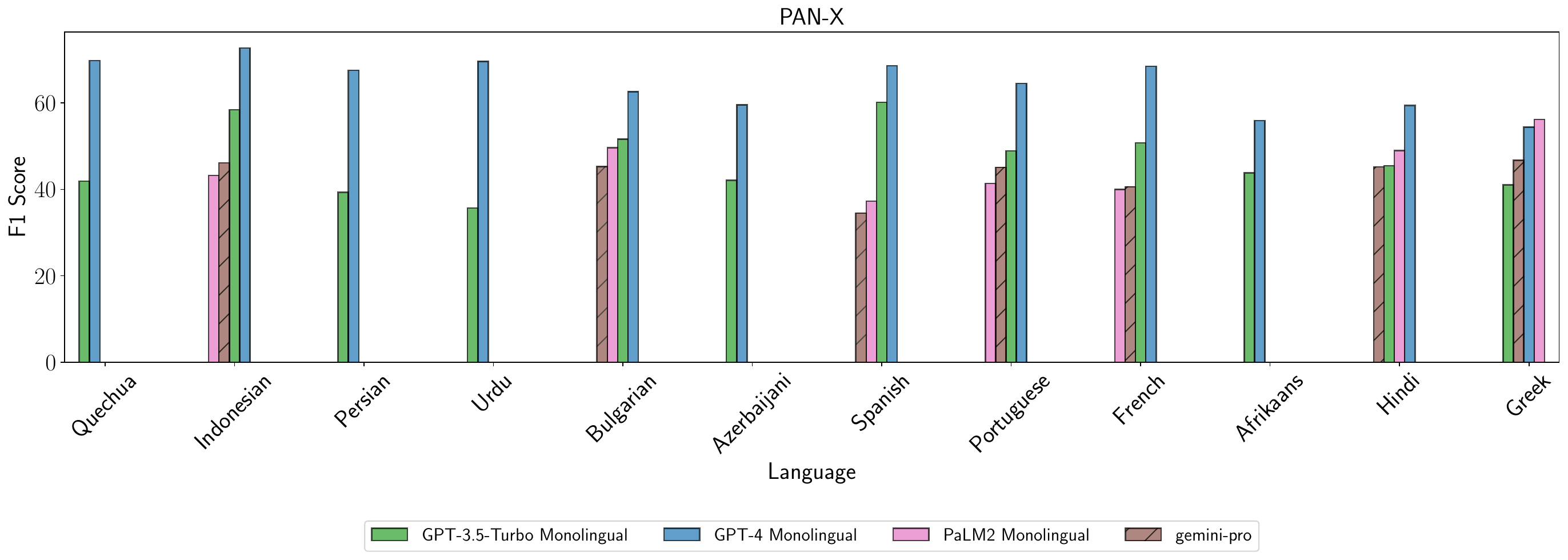}
\end{subfigure}

\begin{subfigure}{\textwidth}
\includegraphics[width=\textwidth]{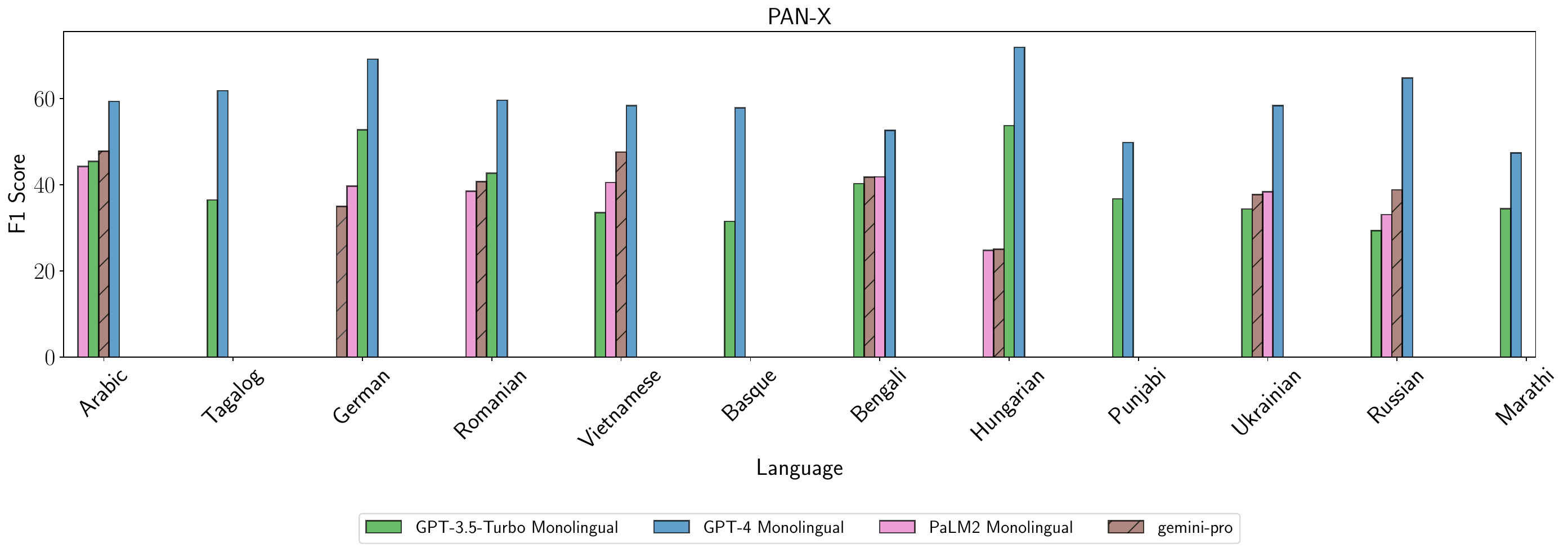}
\end{subfigure}

% \centering
    \caption{Results for PAN-X across all languages with monolingual prompting}
    \label{fig:panx}
\end{figure*}

\begin{figure*}
\centering
\begin{subfigure}{\textwidth}
\includegraphics[width=\textwidth]{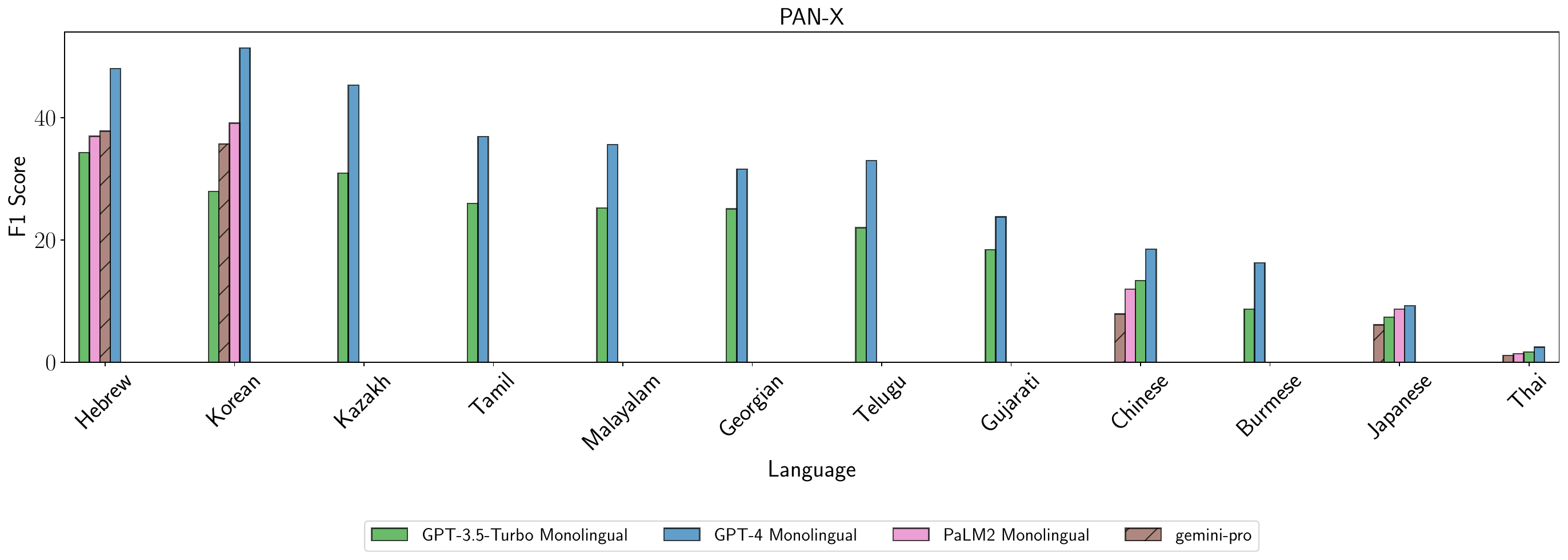}
\end{subfigure}

% \centering
    \caption{Results for PAN-X across all languages with monolingual prompting}
    \label{fig:panx_2}
\end{figure*}

\begin{figure*}
\centering
\begin{subfigure}{\textwidth}
\includegraphics[width=\textwidth]{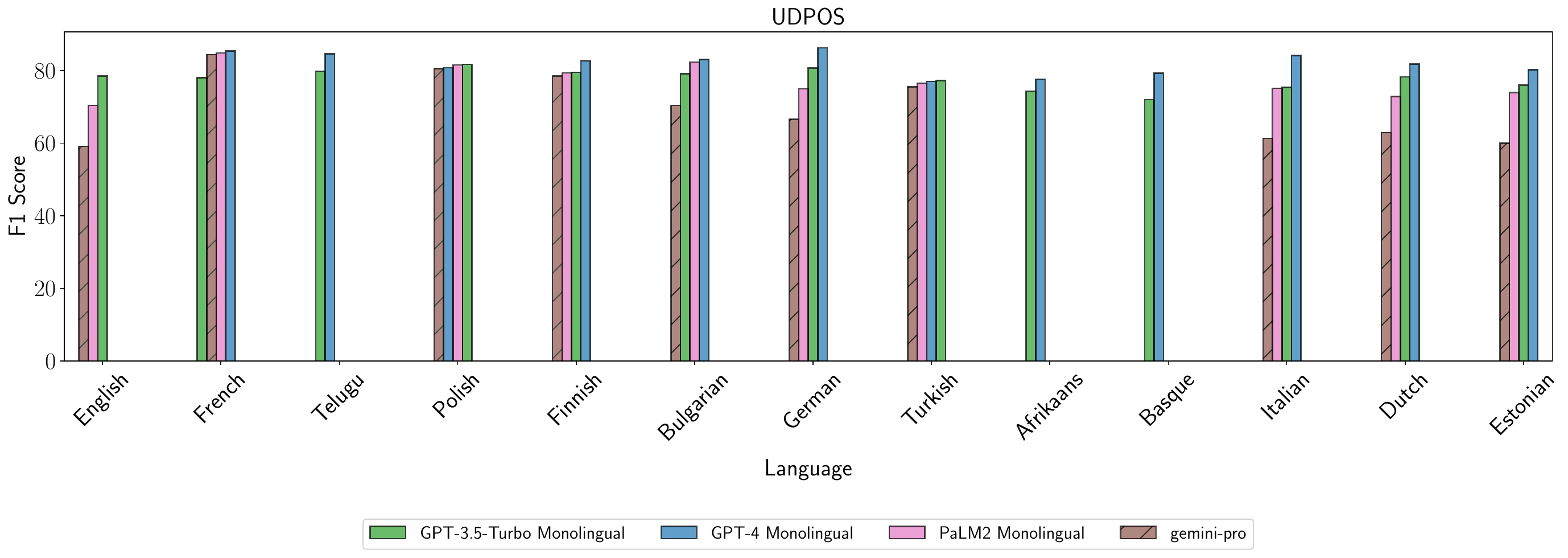}
\end{subfigure}

\begin{subfigure}{\textwidth}
\includegraphics[width=\textwidth]{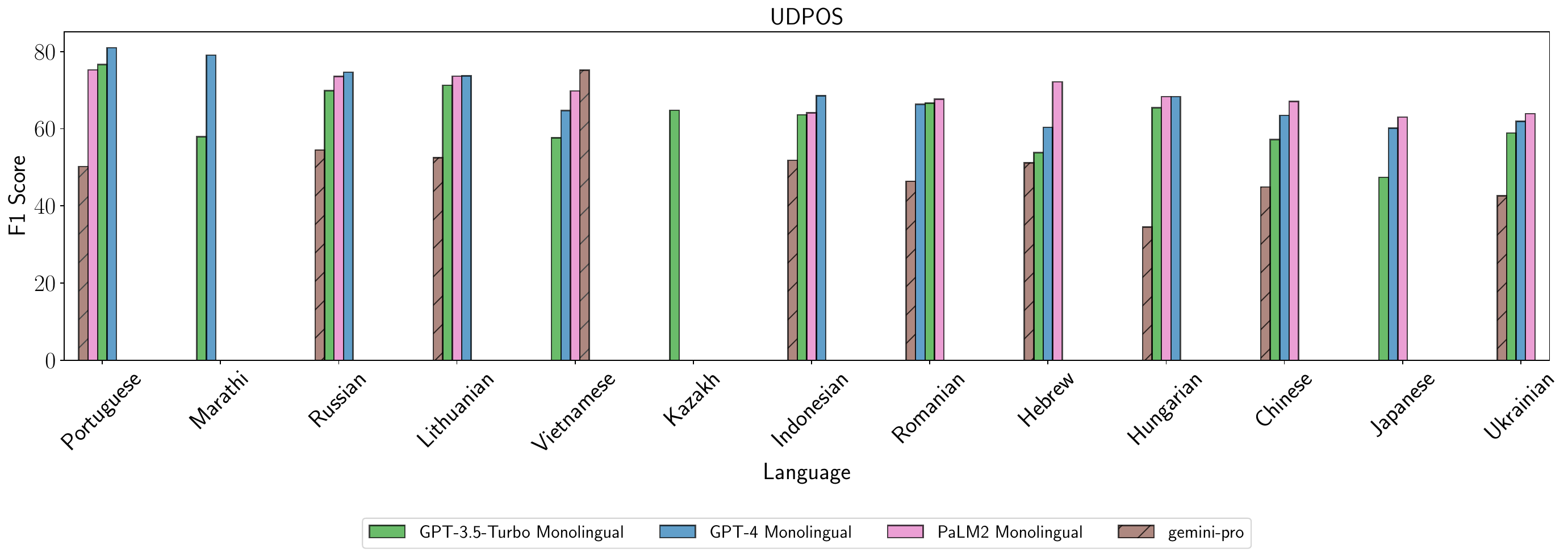}
\end{subfigure}

\begin{subfigure}{\textwidth}
\includegraphics[width=\textwidth]{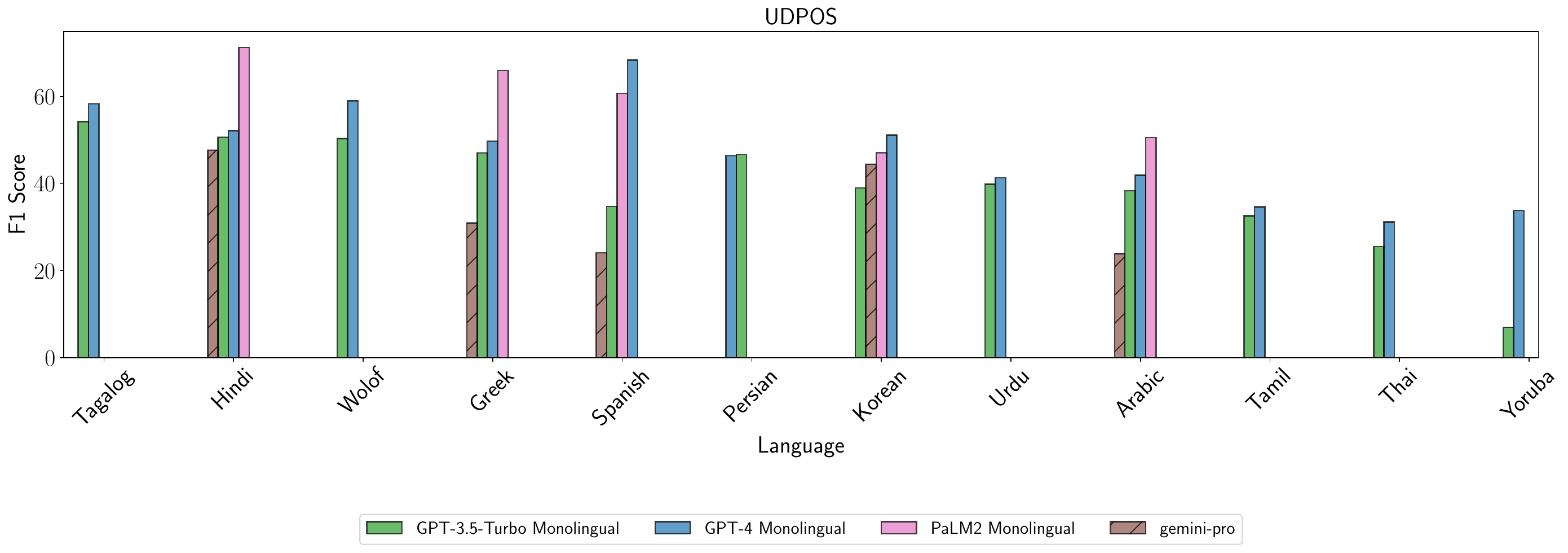}
\end{subfigure}

    \caption{Results for UDPOS across all languages with monolingual prompting}
    \label{fig:udpos}
\end{figure*}

\begin{figure*}
    \centering
    \begin{subfigure}{\textwidth}
        \includegraphics[width=\textwidth]{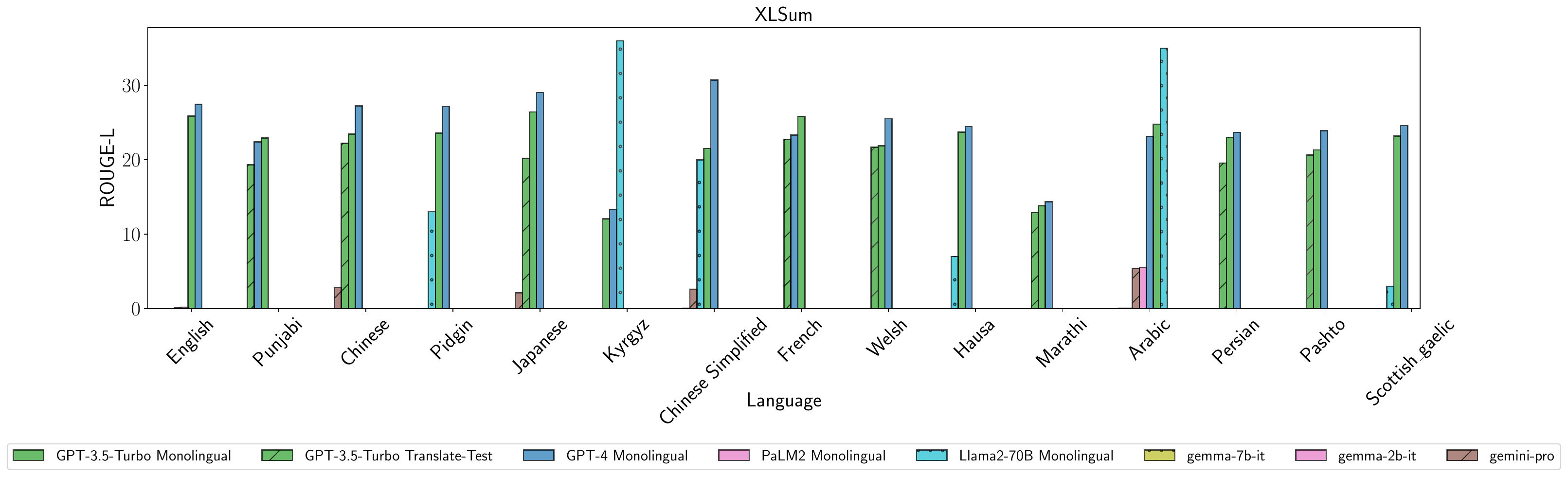}
    \end{subfigure}
    \begin{subfigure}{\textwidth}
        \includegraphics[width=\textwidth]{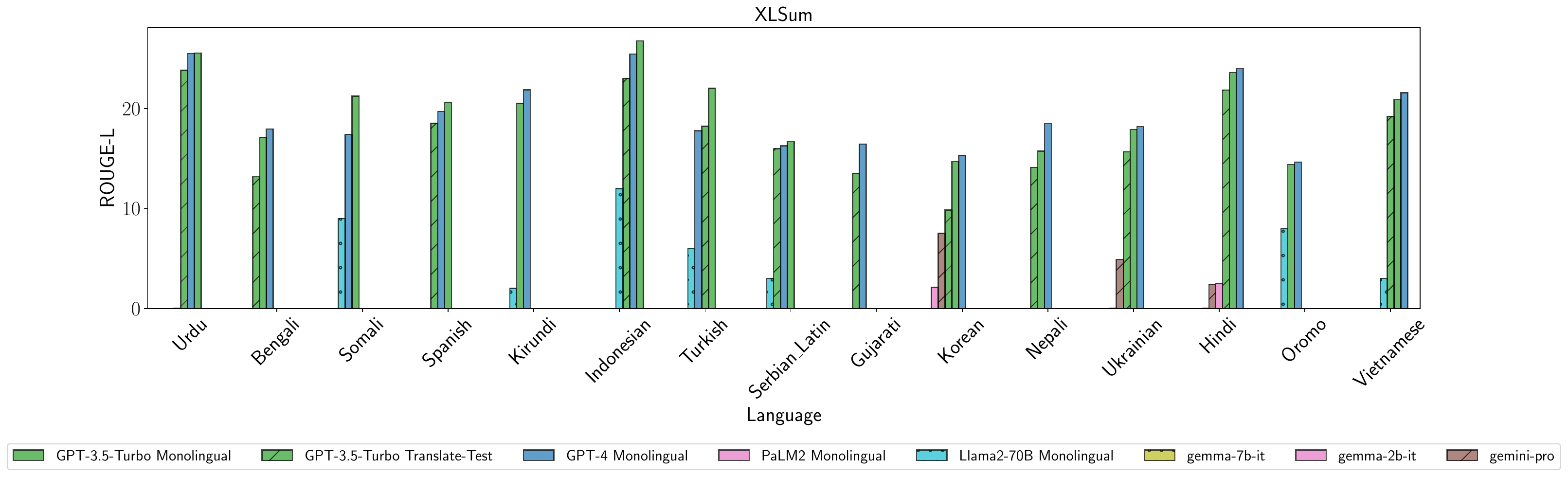}
    \end{subfigure}
    \begin{subfigure}{\textwidth}
        \includegraphics[width=\textwidth]{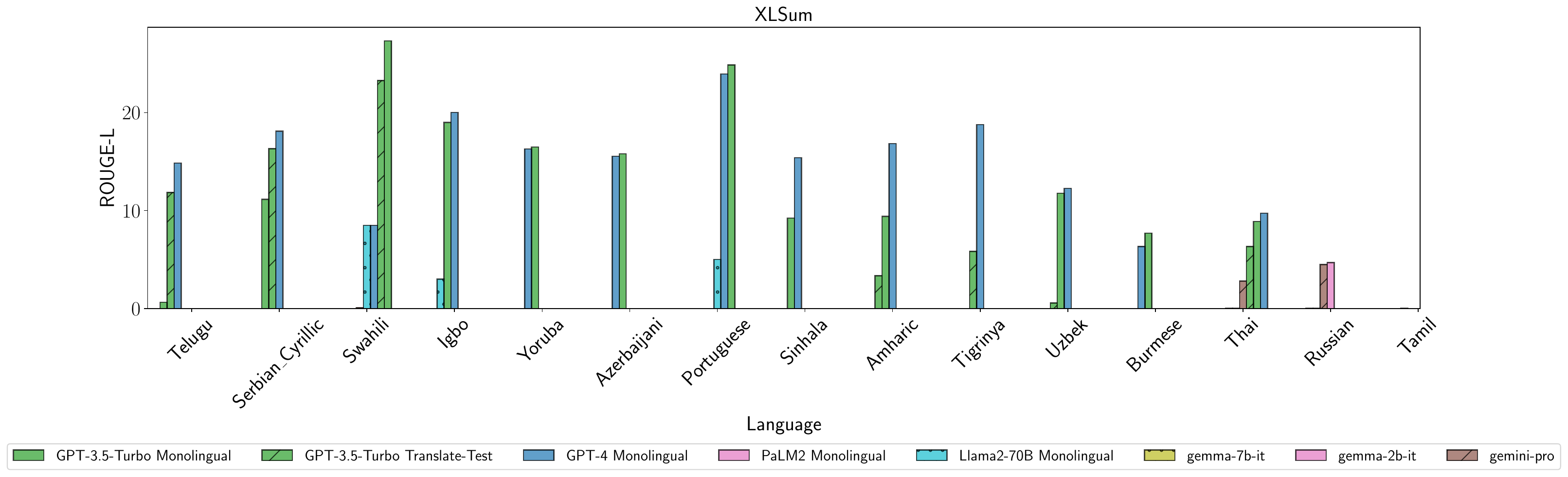}
    \end{subfigure}
    \caption{Results for XLSUM across all languages and models with monolingual prompting}
    \label{fig:xlsum}
\end{figure*}

\begin{figure*}[h]
\centering
  \begin{subfigure}{0.8\textwidth}
    \includegraphics[width=\linewidth]{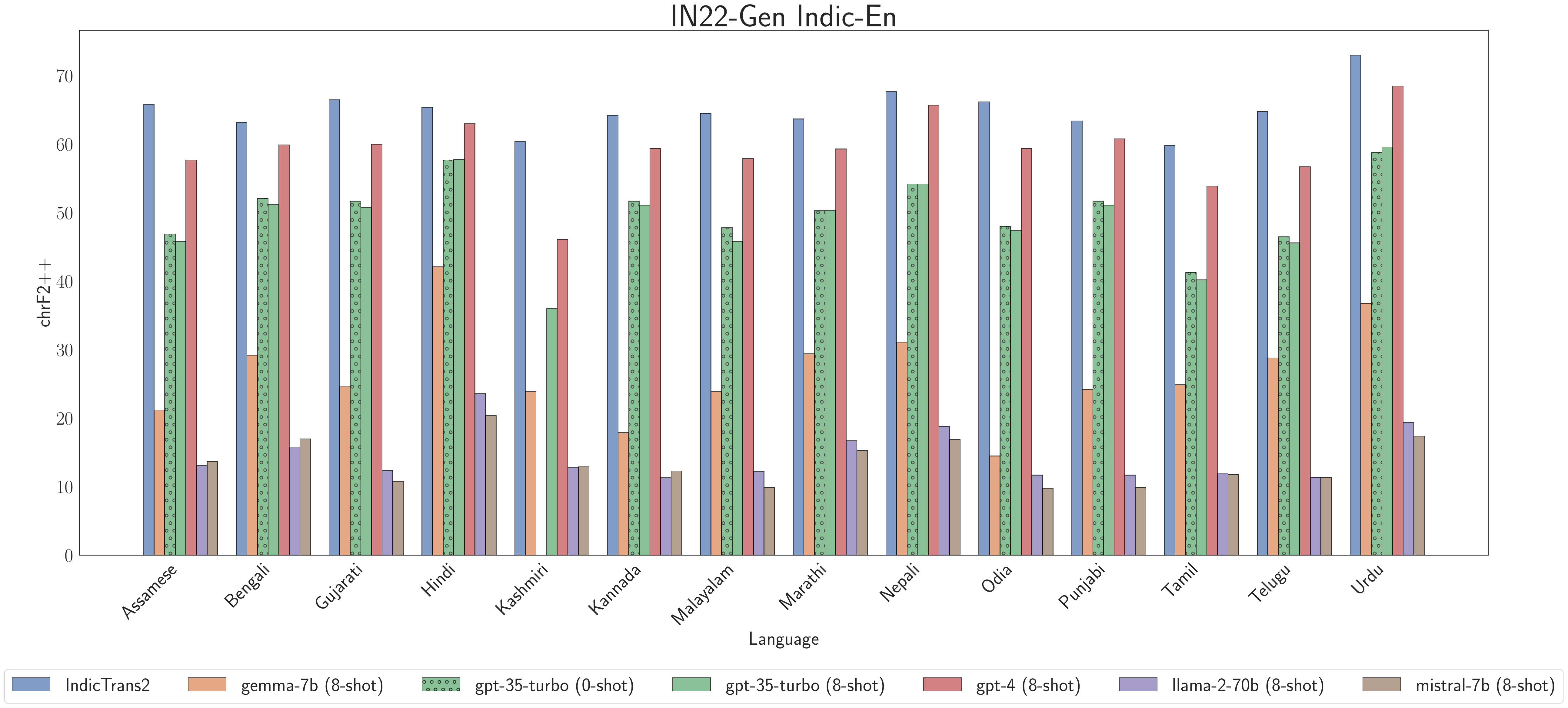}
  \end{subfigure}
  % \\ \\
  \begin{subfigure}{0.8\textwidth}
    \includegraphics[width=\linewidth]{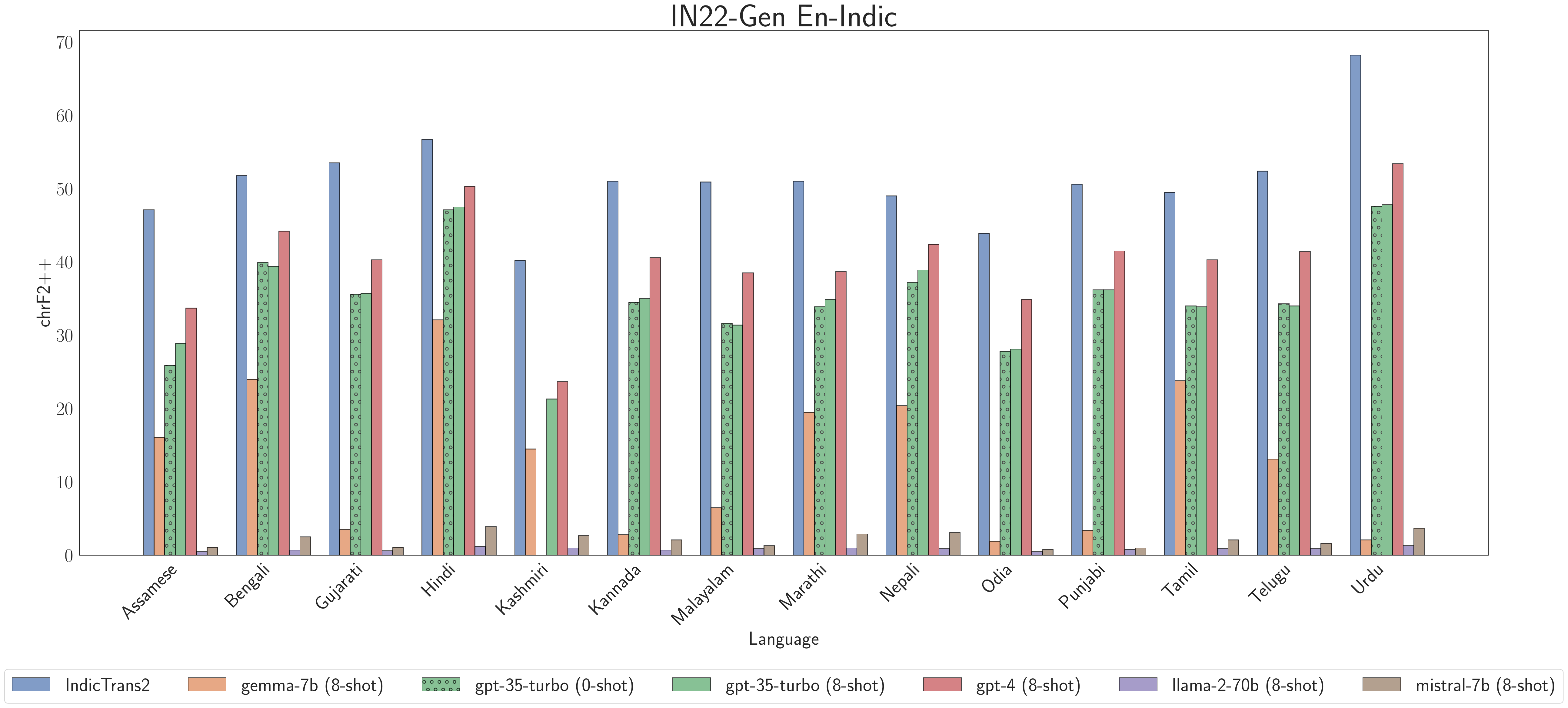}
  \end{subfigure}
  % \\ \\
 \begin{subfigure}{0.8\textwidth}
    \includegraphics[width=\linewidth]{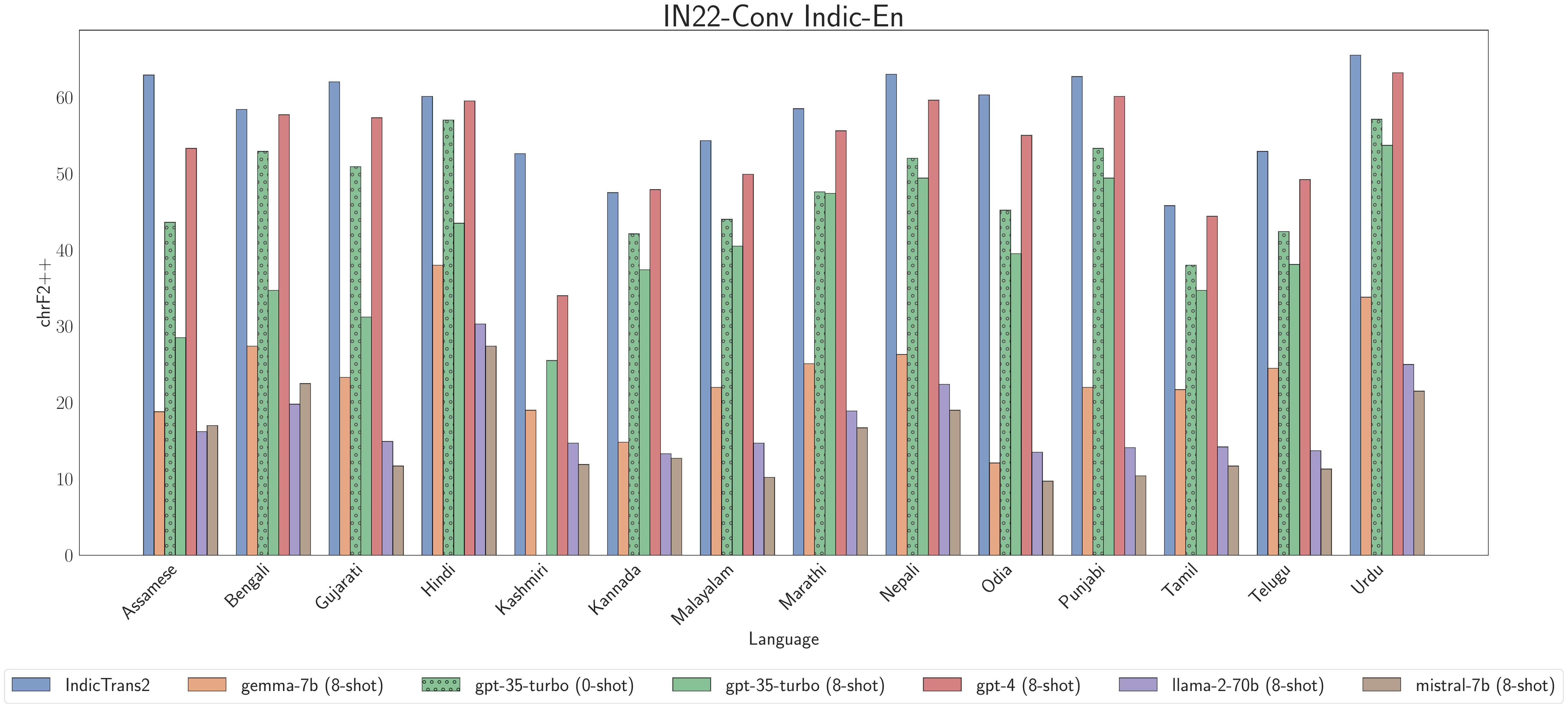}
  \end{subfigure}
  % \\ \\
  \begin{subfigure}{0.8\textwidth}
    \includegraphics[width=\linewidth]{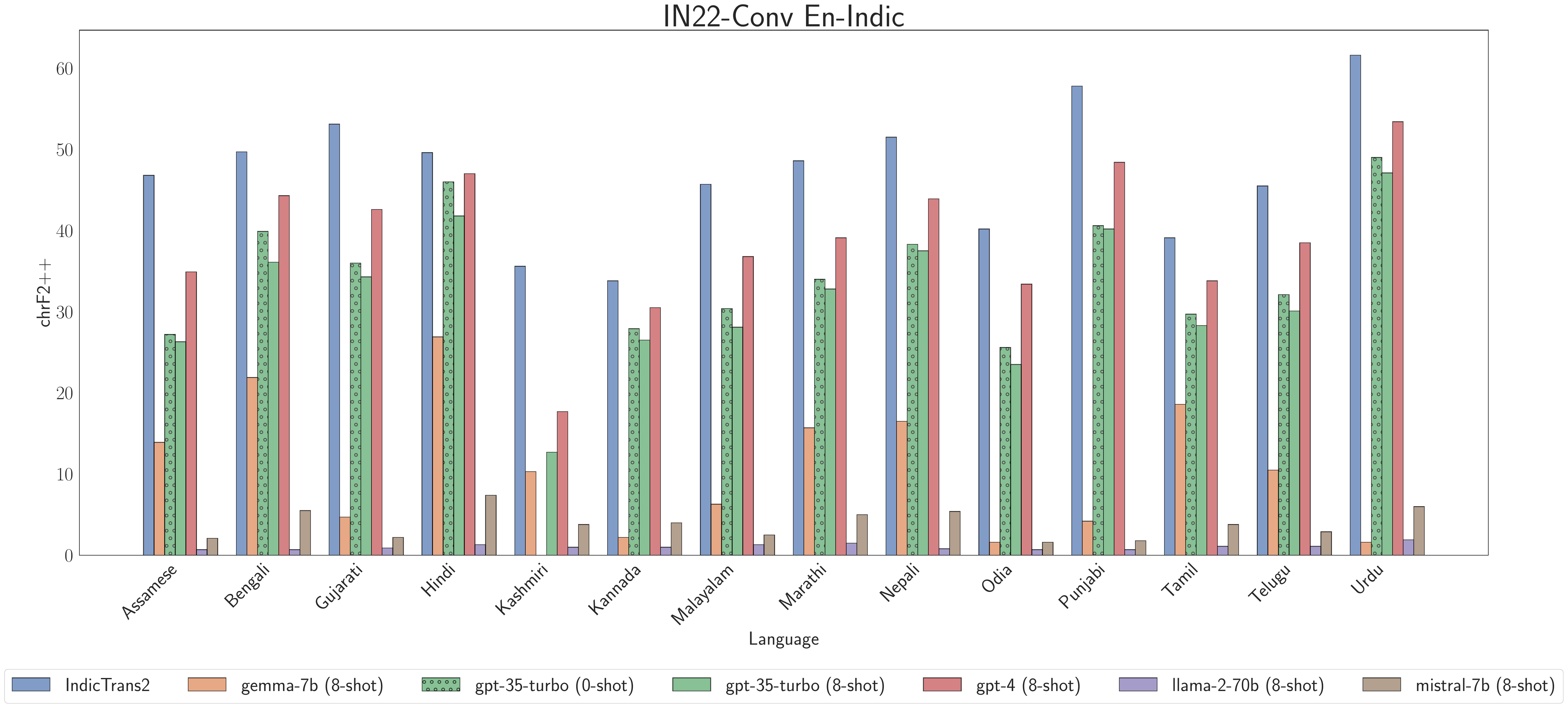}
  \end{subfigure}
  \caption{chrF++ scores of IN22. Note that, Kashmiri 0-shot was not covered in \citet{gala2023indictrans}}
  \label{fig:in22}
\end{figure*}

%% file: appendix/results_tables.tex
\clearpage
% \subsection{Results - Tables}
% Tables \ref{tab:results_summary_xnli} to \ref{tab:detailed_xrisawoz} show our results on various models, languages, and datasets.
\begin{table*}[h]
\centering 
\tiny
\begin{tabular}{l c c c c c c c c c c c c c c c c}
\toprule
Model & en & ar & bg & de & el & es & fr & hi & ru & sw & th & tr & ur & vi & zh & \textbf{avg} \\ \midrule
\multicolumn{10}{l}{\textit{Fine-tuned Baselines}} \\
\midrule
mBERT & 80.8 & 64.3 & 68.0 & 70.0 & 65.3 & 73.5 & 73.4 & 58.9 & 67.8 & 49.7 & 54.1 & 60.9 & 57.2 & 69.3 & 67.8 & 65.4 \\
mT5-Base & 84.7 & 73.3 & 78.6 & 77.4 & 77.1 & 80.3 & 79.1 & 70.8 & 77.1 & 69.4 & 73.2 & 72.8 & 68.3 & 74.2 & 74.1 & 75.4\\
XLM-R Large & 88.7 & 77.2 & 83.0 & 82.5 & 80.8 & 83.7 & 82.2 & 75.6 & 79.1 & 71.2 & 77.4 & 78.0 & 71.7 & 79.3 & 78.2 & 79.2 \\
TuLRv6 - XXL & \textbf{93.3} & \textbf{89.0} & \textbf{90.6} & \textbf{90.0} & \textbf{90.2} & \textbf{91.1} & \textbf{90.7} & \textbf{86.2} & \textbf{89.2} & \textbf{85.5} & \textbf{87.5} & \textbf{88.4} & \textbf{82.7} & \textbf{89.0} & \textbf{88.4} & \textbf{88.8} \\
\midrule
\multicolumn{10}{l}{\textit{Prompt-Based Baselines}} \\
\midrule
BLOOMZ & 67.5 & 60.7 & 46.5 & 54.0 & 47.4 & 61.2 & 61.4 & 56.8 & 53.3 & 50.4 & 43.8 & 42.7 & 50.0 & 61.0 & 56.7 & 54.2\\
\bottomrule
\end{tabular}
\caption{Comparing performance of different models on all languages in XNLI. Metric: Accuracy. Unsupported languages are marked with `-'. Averages are calculated only from supported languages.}
\label{tab:results_summary_xnli}
\end{table*}

% IndicXNLI
\begin{table*}[h]
\footnotesize
\centering
\begin{tabular}{l c c c c c c c c c c c c c }
\toprule
Model & as & bn & gu & hi & kn & ml & mr & or & pa & ta & te & \textbf{avg} \\ \midrule
\multicolumn{10}{l}{\textit{Fine-tuned Baselines}} \\
\midrule
MuRIL & \textbf{76.0} & \textbf{75.0} & \textbf{77.0} & \textbf{77.0} & \textbf{77.0} & \textbf{79.0} & \textbf{74.0} & \textbf{76.0} & \textbf{77.0} & \textbf{77.0} & \textbf{74.0} & \textbf{76.0}\\
\midrule
\multicolumn{10}{l}{\textit{Open AI Models}} \\
\midrule
\texttt{gpt-3.5-turbo} & 49.5 & 53.6 & 50.6 & 55.5 & 53.9 & 48.4 & 49.9 & 47.4 & 53.6 & 48.2 & 47.4 & 50.7 \\
\texttt{gpt-3.5-turbo} (TT) & 54.3 & 61.6 & 61.8 & 59.6 & 60.8 & 59.9 & 58.7 & 58.5 & 62.3 & 58.3 & 60.8 & 59.7 \\
\texttt{text-davinci-003} & 48.6 & 52.6 & 51.2 & 56.9 & 49.1 & 48.2 & 49.4 & 46.4 & 50.4 & 45.5 & 47.2 & 49.6 \\
\texttt{text-davinci-003} (TT) & 56.0 & 66.0 & 64.7 & 62.6 & 63.9 & 61.8 & 60.9 & 60.8 & 64.7 & 61.8 & 63.1 & 62.4 \\
\texttt{gpt-4-32k} & 63.5 & 72.2 & 66.9 & 71.7 & 69.0 & 64.3 & 66.2 & 61.1 & 71.1 & 63.7 & 64.8 & 66.8 \\
\midrule
\multicolumn{10}{l}{\textit{Open Source Models}} \\
\midrule
\texttt{Mistral 7B} & 34.4 & 45.5 & 32.2 & 46 & 37.3 & 21.5 & 38.8 & 0 & 9.9 & 41.2 & 13.3 & 29.1 \\
\texttt{Mistral 7B Instruct} & 23.5 & 30.7 & 19.7 & 30.2 & 30 & 12.5 & 32.9 & 7 & 8 & 27.4 & 29.6 & 28.3
\\
\midrule
\multicolumn{10}{l}{\textit{Google Models}} \\
\midrule
Gemma 2B Instruct & 38.9 & 40.8 & 41.3 & 48.4 & 39.4 & 39.9 & 40.7 & 32.8 & 40.8 & 41.1 & 42 & 40.5 \\
Gemma 7B Instruct & 44.3 & 47.4 & 45.3 & 45.7 & 46.1 & 46.2 & 43.8 & 33.7 & 45.8 & 45.6 & 45.7 & 44.5
\\
\bottomrule
\end{tabular}
\caption{Comparing performance of different models on all languages in IndicXNLI. Metric: Accuracy.}
\label{tab:results_summary_indicxnli}
\end{table*}

% PAWS-X
\begin{table*}[h]
\centering
\footnotesize
\begin{tabular}{l c c c c c c c c}
\toprule
Model & en & de & es & fr & ja & ko & zh & \textbf{avg} \\ \midrule
\multicolumn{9}{l}{\textit{Fine-tuned Baselines}} \\
\midrule
mBERT & 94.0 & 85.7 & 87.4 & 87.0 & 73.0 & 69.6 & 77.0 & 81.9 \\
mT5-Base & 95.4 & 89.4 & 89.6 & 91.2 & 79.8 & 78.5 & 81.1 & 86.4 \\
XLM-R Large & 94.7 & 89.7 & 90.1 & 90.4 & 78.7 & 79.0 & 82.3 & 86.4 \\
TuLRv6 - XXL & \textbf{97.2} & \textbf{95.1} & \textbf{94.8} & \textbf{95.6} & \textbf{89.4} & \textbf{90.4} & \textbf{90.4} & \textbf{93.2} \\
\midrule
\multicolumn{9}{l}{\textit{Prompt-Based Baselines}} \\
\midrule
BLOOMZ & 89.8 & 84.3 & 88.9 & 87.5 & 74.4 & 85.8 & 65.2 & 82.3\\
Llama 2 7B & 68.4 & 65.1 & 67 & 67 & 56 & 53.8 & 60.5 & 62.5\\
Llama 2 13B & 63.3 & 52.3 & 57.7 & 54 & 0 & 0 & 6 & 33.3\\
Llama 2 70B & 53.2 & 63.8 & 65 & 65.6 & 46.5 & 27.7 & 54 & 53.7\\
Mistral 7B & 64.2 & 68.6 & 67.4 & 63.6 & 53.4 & 49.3 & 53.3 & 60 \\
Mistral 7B Instruct & 65.3 & 60.7 & 64.6 & 64.5 & 54.8 & 54.1 & 55.1 & 59.9 \\
\midrule
\multicolumn{9}{l}{\textit{Google Models}} \\
\midrule
PaLM 2 & 81.5 & 77.7 & 77.7 & 78.5 & 73.2 & 71.2 & 76.4 & 76.6\\
gemini-pro & 80.0 & 76.9 & 76.4 & 75.7 & 67.3 & 65.7 & 72.4 & 73.5\\
Gemma 2B Instruct & 54.9 & 54.8 & 53.8 & 54.9 & 53.4 & 55.1 & 53.8 & 54.4 \\
Gemma 7B Instruct & 57.9 & 51.6 & 57.0 & 52.3 & 45.6 & 45.8 & 48.3 & 51.2 \\
\midrule
\multicolumn{9}{l}{\textit{Open AI Models}} \\
\midrule
\texttt{gpt-3.5-turbo} & 72.4 & 70.6 & 72.0 & 72.1 & 67.2 & 66.5 & 69.2 & 70.0 \\
\texttt{gpt-3.5-turbo} (TT) & 72.4 & 70.8 & 69.7 & 70.1 & 61.9 & 62.5 & 63.1 & 67.2 \\
\texttt{text-davinci-003} & 72.5 & 70.6 & 72.7 & 70.7 & 60.6 & 61.8 & 60.8 & 67.1 \\
\texttt{text-davinci-003} (TT) & 72.5 & 69.8 & 70.1 & 71.3 & 65.4 & 65.8 & 65.2 & 68.6 \\
\texttt{gpt-4-32k} & 76.2 & 74.0 & 74.1 & 72.6 & 71.5 & 69.9 & 72.6 & 73.0\\
\bottomrule
\end{tabular}
\caption{Comparing performance of different models on all languages in PAWS-X. Metric: Accuracy.}
\label{tab:results_summary_pawsx}
\end{table*}

% XCOPA
\begin{table*}[h]
\footnotesize
\centering
\begin{tabular}{l c c c c c c c c c c c}
\toprule
Model & en & et & ht & id & it & qu & sw & ta & th & tr &\textbf{avg} \\ \midrule
\multicolumn{10}{l}{\textit{Fine-tuned Baselines}} \\
\midrule
%mBERT & \\
mT5-Base & - & 50.3 & 49.9 & 49.2 & 49.6 & 50.5 & 50.4 & 49.2 & 50.7 & 49.5 & 49.9
\\
%XLM-R Large & \\
TuLRv6 - XXL & - & 77.4 & 78.0 & 92.6 & 96.0 & 61.0 & 69.4 & 85.4 & 87.2 & 92.8 & 74.0 \\
\midrule
\multicolumn{10}{l}{\textit{Prompt-Based Baselines}} \\
\midrule
BLOOMZ & 88.0 & 48.0 & 55.0 & 86.0 & 74.0 & 50.0 & 60.0 & 67.0 & 50.0 & 54.0 & 63.2 \\
XGLM & - & 65.9 & 58.9 & 68.9 & 69.2 & 47.1 & 62.9 & 56.3 & 62.0 & 58.5 & 61.1\\
Llama 2 7B & 74 & 50.6 & 51.2 & 59 & 70.6 & 50.4 & 50.6 & 49.6 & 52 & 52.6 & 56.0 \\
\bottomrule
\end{tabular}
\caption{Comparing performance of different models on all languages in XCOPA. Metric: Accuracy. Unsupported languages are marked with `-'. Averages are calculated only from supported languages.}
\label{tab:results_summary_xcopa}
\end{table*}

% XQuAD
\begin{table*}[h]
\resizebox{\textwidth}{!}{%
% \footnotesize
\centering
\begin{tabular}{l c c c c c c c c c c c c}
\toprule
Model & en & ar & de & el & es & hi & ru & th & tr & vi & zh & \textbf{avg} \\ \midrule
\multicolumn{10}{l}{\textit{Fine-tuned Baselines}} \\
\midrule
mBERT & 83.5 / 72.2 & 61.5 / 45.1 & 70.6 / 54.0 & 62.6 / 44.9 & 75.5 / 56.9 & 59.2 / 46.0 & 71.3 / 53.3 & 42.7 / 33.5 & 55.4 / 40.1 & 69.5 / 49.6 & 58.0 / 48.3 & 64.5 / 49.4 \\
mT5-Base & 84.6 / 71.7 & 63.8 / 44.3 & 73.8 / 54.5 & 59.6 / 35.6 & 74.8 / 56.1 & 60.3 / 43.4 & 57.8 / 34.7 & 57.6 / 45.7 & 67.9 / 48.2 & 70.7 / 50.3 & 66.1 / 54.1 &  67.0 / 49.0\\
XLM-R Large & 86.5 / 75.7 & 68.6 / 49.0 & 80.4 / 63.4 & 79.8 / 61.7 & 82.0 / 63.9 & 76.7 / 59.7 & 80.1 / 64.3 & 74.2 / 62.8 & 75.9 / 59.3 & 79.1 / 59.0 & 59.3 / 50.0 & 76.6 / 60.8 \\
TuLRv6 - XXL & 90.1 / 80.6 & \textbf{85.4 / 69.6} & \textbf{86.1 / 70.4} & \textbf{86.3 / 70.4} & \textbf{87.6 / 71.0} & \textbf{85.9 / 70.5} & \textbf{86.8 / 73.2} & \textbf{87.0 / 81.1} & \textbf{84.3 / 71.0} & \textbf{87.6 / 71.3} & 79.2 / 73.2 & \textbf{86.0 / 72.9} \\
\midrule
\multicolumn{10}{l}{\textit{Prompt-Based Baselines}} \\
\midrule
BLOOMZ & \textbf{92.1 / 83.8} & 82.8 / 69.7 & 76.3 / 60.4 & 49.7 / 37.6 & 86.8 / 71.4 & 83.4 / 72.9 & 65.7 / 47.2 & 20.5 / 15.5 & 51.4 / 37.2 & 86.9 / 72.7 & \textbf{82.4 / 78.6} & 70.7 / 58.8 \\
\bottomrule
\end{tabular}}
\caption{Comparing performance of different models on all languages in XQUAD. Metric: F1 Score / Exact Match.}
\label{tab:results_summary_xquad}
\end{table*}

% TyDiQA-GoldP
\begin{table*}[h]
\centering
\tiny
\begin{tabular}{l c c c c c c c c c c}
\toprule
Model & en & ar & bn & fi & id & ko & ru & sw & te & \textbf{avg} \\ \midrule
\multicolumn{10}{l}{\textit{Fine-tuned Baselines}} \\
\midrule
mBERT & 75.3 / 63.6 & 62.2 / 42.8 & 49.3 / 32.7 & 59.7 / 45.3 & 64.8 / 45.8 & 58.8 / 50.0 & 60.0 / 38.8 & 57.5 / 37.9 & 49.6 / 38.4 & 59.7 / 43.9 \\
mT5-Base & 71.8 / 60.9 & 67.1 / 50.4 & 40.7 / 22.1 & 67.0 / 52.2 & 71.3 / 54.5 & 49.5 / 37.7 & 54.9 / 32.6 & 60.4 / 43.9 & 40.6 / 31.1 & 58.1 / 42.8 \\
XLM-R Large & 71.5 / 56.8 & 67.6 / 40.4 & 64.0 / 47.8 & 70.5 / 53.2 & 77.4 / 61.9 & 31.9 / 10.9 & 67.0 / 42.1 & 66.1 / 48.1 & 70.1 / 43.6 & 65.1 / 45.0 \\
TuLRv6 - XXL & \textbf{85.4 / 76.4} & \textbf{84.1 / 70.4} & 86.9 / 79.6 & \textbf{83.8 / 72.8} & \textbf{88.8 / 77.9} & \textbf{78.5 / 67.8} & \textbf{81.9 / 68.6} & \textbf{87.2 / 79.6} & 85.2 / 71.6 & \textbf{84.6 / 73.8} \\
\midrule
\multicolumn{10}{l}{\textit{Prompt-Based Baselines}} \\
\midrule
BLOOMZ & 82.4 / 70.9 & 81.9 / 62.2 & \textbf{87.8 / 82.3} & 43.6 / 28.6 & 85.0 / 71.0 & 52.3 / 43.1 & 67.4 / 51.5 & 86.0 / 77.2 & \textbf{90.3 / 81.6} & 75.2 / 63.2 \\
\bottomrule
\end{tabular}
\caption{Comparing performance of different models on all languages in TyDiQA. Metric: F1 Score / Exact Match. Unsupported languages are marked with `-'. Averages are calculated only from supported languages.}
\label{tab:results_summary_tydiqa}
\end{table*}

\begin{table*}[h]
\centering
\tiny
\begin{tabular}{lcccccccc}
\hline
\textbf{Model}                                  & \textbf{en} & \textbf{ar} & \textbf{de} & es          & hi          & vi          & zh          & \textbf{avg} \\ \hline
\multicolumn{9}{l}{\textit{Fine-tuned Baselines}}                                                                                                                \\ \hline
mBERT                                           & 80.2 / 67.0 & 52.3 / 34.6 & 59.0 / 43.8 & 67.4 / 49.2 & 50.2 / 35.3 & 61.2 / 40.7 & 59.6 / 38.6 & 61.4 / 44.2  \\
mT5-Base                                        & 81.7 / 66.9 & 57.1 / 36.9 & 62.1 / 43.2 & 67.1 / 47.2 & 55.4 / 37.9 & 65.9 / 44.1 & 61.6 / 38.6 & 64.4 / 45.0  \\
XLM-R Large                                     & 83.5 / 70.6 & 66.6 / 47.1 & 70.1 / 54.9 & 74.1 / 56.6 & 70.6 / 53.1 & 74.0 / 52.9 & 62.1 / 37.0 & 71.6 / 53.2  \\
TuLRv6 - XXL &
  \textbf{86.6 / 74.4} &
  \textbf{76.2 / 56.5} &
  \textbf{80.2 / 67.0} &
  \textbf{81.7 / 65.1} &
  \textbf{82.2 / 64.8} &
  \textbf{82.3 / 63.2} &
  \textbf{78.1 / 56.5} &
  \textbf{81.0 / 63.9} \\ \hline
\multicolumn{9}{l}{\textit{Prompt-Based Baselines}}                                                                                                              \\ \hline
Llama 2 7B                                      & 69.6 / 50.7 & 5.2 / 1.2   & 44.5 / 27.3 & 51.9 / 29.8 & 0.5 / 0.0   & 42.5 / 27.4 & 7.0 / 4.2   & 31.6 / 20.1  \\
\hline
\end{tabular}
\caption{Comparing performance of different models on all languages in MLQA. Metric: F1 Score / Exact Match.}
\label{tab:results_summary_mlqa}
\end{table*}

% gemini-pro & 
% \\
% Gemma 2B Instruct & \\
% Gemma 7B Instruct & \\

% IndicQA
\begin{table*}[h]
\resizebox{\textwidth}{!}{%
\begin{tabular}{l c c c c c c c c c c c c c }
\toprule
Model & as & bn & gu & hi & kn & ml & mr & or & pa & ta & te & \textbf{avg} \\ \midrule
\multicolumn{10}{l}{\textit{Fine-tuned Baselines}} \\
\midrule
BLOOMZ & 40.6 / 31.7 & 42.9 / 36.6 & 37.2 / 29.9 & 44.0 / 45.1 & 37.8 / 26.6 & 30.5 / 28.4 & 39.2 / 33.0 & 25.4 / 22.0 & 26.4 / 33.5 & 39.7 / 35.9 & 38.9 / 34.7 & 36.6 / 32.5 \\
\midrule
\multicolumn{10}{l}{\textit{Google Models}} \\
\midrule
Gemma 2B Instruct & 6.6 / 5.5 & 4.5 / 2.9 & 10.1 / 9.6 & 5.9 / 1.8 & 12.6 / 11.6 & 4 / 3 & 5.5 / 3.9 & 2.3 / 2.3 & 7.7 / 5.1 & 5.6 / 2.4 & 3.4 / 2.5 & 6.2 / 4.6\\
Gemma 7B Instruct & 19.1 / 12.6 & 21.9 / 14.9 & 22.2 / 20.1 & 32 / 21.8 & 21.8 / 17.4 & 15.6 / 12.6 & 22.5 / 16.4 & 7.5 / 7.4 & 19.4 / 15 & 24.3 / 18.1 & 15.2 / 12.6 & 20.1 / 15.4\\
\midrule
\multicolumn{10}{l}{\textit{Open AI Models}} \\
\midrule
\texttt{gpt-3.5-turbo} & 35.3 / 21.4 & 49.5 / 30.2 & 40.5 / 25.5 & 55.9 / 39.3 & 35.3 / 20.4 & 30.0 / 19.2 & 50.0 / 32.0 & 22.1 / 12.7 & 35.8 / 15.1 & 32.7 / 21.6 & 32.9 / 19.7 & 38.2 / 23.4 \\
\texttt{text-davinci-003} & 6.7 / 3.2 & 10.3 / 5.8 & 5.4 / 3.5 & 16.8 / 11.8 & 7.1 / 3.9 & 3.6 / 2.3 & 14.6 / 8.5 & 6.9 / 3.4 & 10.7 / 4.1 & 4.2 / 2.5 & 6.8 / 3.6 & 8.4 / 4.8 \\
\texttt{gpt-4-32k} & \textbf{58.8 / 40.4} & \textbf{67.1 / 47.4} & \textbf{59.4 / 42.4} & \textbf{75.2 / 62.2} & \textbf{47.1 / 31.6} & \textbf{48.3 / 33.7} & \textbf{60.7 / 43.1} & \textbf{29.9 / 16.7} & \textbf{56.1 / 34.1} & \textbf{54.0 / 39.7} & \textbf{47.9 / 27.8} & \textbf{55.0 / 38.1}
\\
\bottomrule
\end{tabular}}
\caption{Comparing performance of different models on all languages in IndicQA. Metric: F1 Score / Exact Match.}
\label{tab:results_summary_indicqa}
\end{table*}

% AfriQA (Note - no PaLM 2 results)
\begin{table*}[h]
\centering
\begin{tabular}{l c c c c c c c c c c c c}
\toprule
Model & bem & fon & hau & ibo & kin & swa & twi & wol & yor & zul & \textbf{avg} \\ \midrule
\multicolumn{10}{l}{\textit{Open Source Baselines}} \\
\midrule
Llama 2 7B & 0.4 & 0.5 & 0.6 & 0.7 & 0.3 & 0.7 & 0.9 & 0.4 & 2.9 & 1.3 & 0.9 \\
Llama 2 13B & 0.4 & 0.5 & 0.6 & 0.7 & 0.3 & 0.7 & 0.9 & 0.4 & 3 & 1.3 & 0.9 \\
Llama 2 70B & 0.4 & 0.5 & 0.6 & 0.7 & 0.3 & 0.7 & 0.9 & 0.4 & 3 & 1.3 & 0.9\\
Mistral 7B & 1.9 & 0 & 1.7 & 1.4 & 1.9 & 4.3 & 12 & 0.8 & 1.4 & 2.9 & 2.83 \\
Mistral 7B Instruct & 1.9 & 5.6 & 0.4 & 0.1 & 1.9 & 2 & 0.3 & 1.1 & 0.5 & 1.8 & 1.56 \\
\midrule
\multicolumn{10}{l}{\textit{Google Models}} \\
\midrule
Gemma 2B Instruct & 0.9 & 0.0 & 0.0 & 0.0 & 0.2 & 0.4 & 0.4 & 0.0 & 0.5 & 0.0 & 0.2 \\
Gemma 7B Instruct & 0.9 & 0.0 & 0.4 & 0.0 & 0.3 & 0.5 & 0.0 & 0.0 & 0.0 & 0.0 & 0.2 \\
\midrule
\multicolumn{10}{l}{\textit{Open AI Models}} \\
\midrule
\texttt{gpt-3.5-turbo} & 4.4 & 10.5 & 11.2 & 20.5 & 10.4 & 18.5 & 20.3 & 4.9 & 16.4 & 8.8 & 12.6 \\
\texttt{gpt-4-32k} & \textbf{17.7} & \textbf{15.0} & \textbf{32.9} & \textbf{44.7} & \textbf{31.5} & \textbf{38.7} & \textbf{30.8} & \textbf{11.6} & \textbf{28.2} & \textbf{27.5} & \textbf{27.9} \\
\bottomrule
\end{tabular}
\caption{Comparing performance of different models on all languages in AfriQA. Metric: F1 Score.}
\label{tab:results_summary_afriqa}
\end{table*}

\begin{table*}[h]
\resizebox{\textwidth}{!}{%
\begin{tabular}{l c c c c c c c c c c c c c c c c c c c c}
\toprule
Model & en & af & ar & bg & de & el & es & et & eu & fa & fi & fr & he & hi & hu & id & it & ja  & kk \\ \midrule
\multicolumn{10}{l}{\textit{Fine-tuned Baselines}} \\
\midrule
mBERT & 96.4 & 86.7 & 50.0 & 84.7 & 88.7 & 80.9 & 86.6 & 79.9 & 62.1 & 65.5 & 73.3 & 81.2 & 55.5 & 66.0 & 78.6 & 74.2 & 87.8 & 47.2 & 70.4 \\
XLM-R Large  & \textbf{97.0} & \textbf{89.2} & \textbf{63.0} & \textbf{88.3} & \textbf{91.2} & \textbf{86.5} & \textbf{89.2} & \textbf{87.3} & 74.9 & \textbf{70.8} & \textbf{82.7} & \textbf{86.7} & 67.5 & \textbf{75.2} & \textbf{83.4} & \textbf{75.7} & \textbf{89.2} & 29.3 & \textbf{78.3} \\
\midrule
\multicolumn{10}{l}{\textit{Google Models}} \\
\midrule
PaLM 2 & 70.4 & - & 50.6 & 82.3 & 74.9 & 66.0 & 60.7 & 73.9 & - & - & 79.4 & 84.9 & \textbf{72.2} & 71.3 & 68.3 & 64.2 & 75.1 & 63.1 & - \\
gemini-pro & 59.1 & - & 23.9 & 70.4 & 66.6 & 30.9 & 24.1 & 60.0 & - & - & 78.5 & 84.4 & 51.2 & 47.7 & 34.5 & 51.8 & 61.3 & - & -\\
\midrule
\multicolumn{10}{l}{\textit{Open AI Models}} \\
\end{tabular}}
\caption{Comparing performance of different models on all languages in UDPOS. Metric: F1 Score. All numbers are Monolingual results except the ones marked with $\dagger$ symbol which indicate Zero-Shot Cross-Lingual results (due to the absence of training data in those languages. Unsupported languages are marked with `-'. Averages are calculated only from supported languages.}
\label{tab:results_summary_udpos}
\end{table*}

% NER
\begin{table*}[h]
\resizebox{\textwidth}{!}{%
\begin{tabular}{l c c c c c c c c c c c c c c c c c c c c c c c c c}
\toprule
Model & en & af & ar & az & bg & bn & de & el & es & et & eu & fa & fi & fr & gu & he & hi & hu & id & it & ja & jv & ka & kk \\ \midrule
\multicolumn{10}{l}{\textit{Fine-tuned Baselines}} \\
\midrule
mBERT & \textbf{86.4} & 76.1 & 42.9 & 65.5 & 76.7 & 69.7 & 79.5 & 70.9 & \textbf{75.3} & 75.8 & \textbf{64.4} & 40.0 & 76.6 & 79.6 & 51.3 & \textbf{56.2} & 65.9 & 76.1 & 61.0 & \textbf{81.3} & \textbf{29.2} & 62.4 & 65.1 & 50.3\\
XLM-R Large & 85.4 & \textbf{78.6} & 47.3 & \textbf{69.4} & \textbf{80.9} & \textbf{74.7} & \textbf{80.7} & \textbf{79.2} & 71.8 & \textbf{78.7} & 61.6 & 55.2 & \textbf{79.6} & \textbf{79.8} & \textbf{62.7} & 55.5 & \textbf{70.9} & \textbf{80.2} & 51.8 & 80.3 & 18.5 & 61.9 & \textbf{70.9} & \textbf{54.4} \\
\midrule
\multicolumn{10}{l}{\textit{Google Models}} \\
\midrule
PaLM 2 & 35.8 & - & 44.2 & - & 49.6 & 41.8 & 39.7 & 56.2 & 37.3 & 50.5 & - & - & 53.9 & 40.0 & - & 37.0 & 49.0 & 24.8 & 43.2 & 53.9 & 8.7 & - & - & - \\
gemini-pro & 34.1 & - & 47.8 & - & 45.3 & 41.8 & 35.0 & 46.7 & 34.5 & 44.3 & - & - & 46.3 & 40.6 & - & 37.8 & 45.2 & 25.0 & 46.1 & 53.4 & 6.1 & - & - & - & \\
\midrule
\multicolumn{10}{l}{\textit{Open AI Models}} \\
\bottomrule
\end{tabular}}
\caption{Comparing performance of different models on all languages in PAN-X. Metric: F1 Score. Unsupported languages are marked with `-'. Averages are calculated only from supported languages.}
\label{tab:results_summary_panx}
\end{table*}

% XStoryCloze (Note - no Llama 2 results)
\begin{table*}[h]
\centering 
\small
\begin{tabular}{l c c c c c c c c c c c c }
\toprule
Model & ar & en & es & eu & hi & id & my & ru & sw & te & zh &
\textbf{avg} \\ \midrule
%\multicolumn{10}{l}{\textit{Fine-tuned Baselines}} \\
%\midrule
%mBERT & \\
%mT5-Base & \\
%XLM-R Large &  \\
%\midrule
\multicolumn{10}{l}{\textit{Prompt-Based Baselines}} \\
\midrule
BLOOMZ & 79.7 & 95.7 & 87.3 & 70.5 & 79.9 & 85.6 & 49.9 & 67.3 & 65.3 & 67.4 & 90.0 & 76.2 \\
XGLM & 59.8 & 75.9 & 69.2 & 63.8 & 62.5 & 70.8 & 61.2 & 72.4 & 65.2 & 63.4 & 67.7 & 66.5 \\
\midrule
\multicolumn{10}{l}{\textit{Google Models}} \\
\midrule
PaLM 2 & 1.7 & 53.1 & 52.4 & - & 0.7 & 4.2 & - & 28.5 & 47.1 & - & 23.9 & 18.8 \\
gemini-pro & 96.9 & 98.7 & 98.3 & - & 97.6 & 97.5 & - & 97.6 & 96.1 & - & 97.4 & \textbf{97.5} \\
Gemma 2B Instruct & 64.7 & 76.6 & 69.1 & 53.1 & 64.3 & 65.9 & 55.8 & 65.3 & 54.7 & 54.4 & 71.1 & 63.2 \\
Gemma 7B Instruct & 68.5 & 89.9 & 90.7 & 67.4 & 72.1 & 79.3 & 34.0 & 84.4 & 60.4 & 41.1 & 86.7 & 70.4 \\
\midrule
\multicolumn{10}{l}{\textit{Open AI Models}} \\
\midrule
\texttt{gpt-3.5-turbo} & 92.5 & 96.8 & 95.8 & 78.4 & 91.1 & 95.0 & 57.2 & 96.6 & 92.3 & 73.1 & 95.6 & 87.7 \\
\texttt{gpt-3.5-turbo} (TT) & 94.3 & 96.8 & 96.1 & 92.5 & 94.7 & 95.2 & 88.6 & 96.2 & 88.7 & 93.6 & 95.6 & 93.9 \\
\texttt{text-davinci-003} & 87.4 & 98.3 & 97.6 & 78.1 & 77.8 & 96.4 & 47.4 & 94.2 & 78.1 & 57.6 & 95.0 & 82.5 \\
\texttt{text-davinci-003} (TT) & 95.0 & 98.3 & 96.2 & 94.1 & 95.1 & 95.9 & 90.1 & 96.9 & 90.7 & 94.3 & 96.2 & 94.8 \\
\texttt{gpt-4-32k} & \textbf{99.1} & \textbf{99.6} & \textbf{99.5} & \textbf{97.6} & \textbf{98.8} & \textbf{99.0} & 77.6 & 99.1 & \textbf{98.4} & 93.4 & \textbf{99.2} & 96.5 \\
\texttt{gpt-4-32k} (TT) & 97.7 & \textbf{99.6} & 98.7 & 96.8 & 97.9 & 98.1 & \textbf{93.2} & \textbf{99.2} & 93.6 & \textbf{96.4} & 98.3 & 97.0
\\
\bottomrule
\end{tabular}
\caption{Comparing performance of different models on all languages in XStoryCloze. Metric: Accuracy. Unsupported languages are marked with `-'. Averages are calculated only from supported languages.}
\label{tab:results_summary_xstorycloze}
\end{table*}

% GLUECoS (Note - no Llama 2 results)
\begin{table*} [h]
\centering
\begin{tabular}{l c c}
\toprule
Model & NLI En-Hi & Sentiment En-Es \\
\midrule
\multicolumn{3}{l}{\textit{Fine-tuned Baselines}} \\
\midrule
mBERT & 63.1 & \textbf{69.31} \\
\midrule
\multicolumn{3}{l}{\textit{Google Models}} \\
\midrule
PaLM 2 & 82.8 & 51.5 \\
gemini-pro & 80.8 & 29.4 \\
\midrule
\multicolumn{3}{l}{\textit{Open AI Models}} \\
\midrule
\texttt{text-davinci-003} & 72.1 & 68.8 \\
\texttt{gpt-3.5-turbo} & 78.8 & 68.0 \\
\texttt{gpt-4-32k} & \textbf{90.4} & 45.5 \\
\bottomrule
\end{tabular}
\caption{Comparing performance of different models on code-mixing datasets from \citet{khanuja2020gluecos}. Metric: Accuracy.}
\label{tab:cm_results}
\end{table*}

% XLSum
\begin{table*}[h]
\resizebox{\textwidth}{!}{%
\begin{tabular}{l c c c c c c c c c c c c c c c c c c c c c c c }
\toprule
Model & en & am & ar & az & bn & cy & es & fa & fr & gd & gu & ha & hi & id & ig & ja & ko & ky & mr & my & ne & om & pa \\
\midrule
\multicolumn{10}{l}{\textit{Fine-tuned Baselines}} \\
\midrule
mT5-RL & \textbf{35.1} & \textbf{24.6} & 31.2 & 24.4 & \textbf{24.2} & \textbf{29.9} & \textbf{27.4} & \textbf{33.5} & \textbf{31.3} & \textbf{26.3} & \textbf{22.7} & \textbf{35.3} & \textbf{33.5} & \textbf{34.4} & \textbf{25.6} & \textbf{38.7} & \textbf{27.1} & 17.7 & \textbf{22.9} & \textbf{17.3} & \textbf{27.9} & \textbf{22.3} & \textbf{28.8} \\
\midrule
\multicolumn{10}{l}{\textit{Prompt-Based Baselines}} \\
\midrule
Llama 2 7B & 0.0 & 0.0 & 0.0 & \textbf{32.0} & 0.0 & 23.8 & 0.0 & 0.0 & 0.0 & 7.1 & 0.0 & 7.7 & 0.0 & 2.7 & 10.5 & 0.0 & 0.0 & 0.0 & 0.0 & 0.0 & 0.0 & 1.1 & 0.0 \\
Llama 2 13B & 0.0 & 0.0 & 0.0 & 25.0 & 0.0 & 27.0 & 0.0 & 0.0 & 0.0 & 8.0 & 0.0 & 4.0 & 0.0 & 2.0 & 11.0 & 0.0 & 0.0 & 0.0 & 0.0 & 0.0 & 0.0 & 1.0 & 0.0 \\
Llama 2 70B & 0.0 & 0.0 & \textbf{35.0} & 0.0 & 0.0 & 0.0 & 0.0 & 0.0 & 0.0 & 3.0 & 0.0 & 7.0 & 0.0 & 12.0 & 3.0 & 0.0 & 0.0 & \textbf{36.0} & 0.0 & 0.0 & 0.0 & 8.0 & 0.0\\
\midrule
\multicolumn{10}{l}{\textit{Google Models}} \\
\bottomrule
\end{tabular}}
\caption{Comparing performance of different models on all languages in XLSum. Metric: Rouge - L.\\
sr*: \texttt{serbian\_cyrillic}, sr**: \texttt{serbian\_latin}. Unsupported languages are marked with `-'. Averages are calculated only from supported languages.}
\label{tab:results_summary_xlsum}
\end{table*}

\begin{table*}[h]
\centering
\small
\begin{tabular}{@{}lcccccc@{}}
\toprule
Model & id & tr & zh & sw & ta & \textbf{avg} \\ \midrule
\textit{LLaVA Models} & & & & & & \\ \midrule
bakllava-v1 & 0.25 & 0.27 & 0.36 & 0.38 & 0.22 & 0.30 \\
bakllava-v1 (TT) & 0.58 & 0.58 & 0.56 & 0.54 & 0.52 & 0.56 \\
llava-v1.5-13B & 0.53 & 0.52 & 0.53 & 0.51 & 0.50 & 0.52 \\
llava-v1.5-13B (TT) & 0.53 & 0.57 & 0.52 & 0.53 & 0.54 & 0.54 \\
vip-llava-13B & 0.52 & 0.50 & 0.20 & 0.50 & 0.14 & 0.37 \\
vip-llava-13B (TT) & 0.56 & 0.56 & 0.53 & 0.52 & 0.51 & 0.54 \\ \midrule
\textit{Google Models} & & & & & & \\ \midrule
\texttt{gemini-pro-vision} & 0.55 & 0.60 & 0.59 & 0.55 & 0.52 & 0.56 \\
\texttt{gemini-pro-vision (TT)} & 0.61 & 0.63 & 0.57 & 0.58 & 0.63 & 0.60 \\ \midrule
\textit{OpenAI Models} & & & & & & \\ \midrule
\texttt{gpt-4-vision} & 0.81 & 0.81 & 0.71 & 0.74 & 0.78 & 0.77 \\
\texttt{gpt-4-vision (TT)} & 0.74 & 0.82 & 0.68 & 0.68 & 0.69 & 0.72 \\ \bottomrule
\end{tabular}
\caption{Comparing performance of Multimodal models on MaRVL. Metric: Accuracy}
\label{tab:MARVL_results}
\end{table*}

% % XRisaWoz (Note - No baselines)
% \begin{table*}[h]
% \centering
% \begin{tabular}{l c c c c c c c }
% \toprule
% Model & en & en\_hi & fr & hi & ko & zh & \textbf{avg} \\ 
% \midrule
% \multicolumn{7}{l}{\textit{Prompt-Based Baselines}} \\
% \midrule
% PaLM 2 & 55.0 & 10.0 & 38.3 & 16.7 & 26.7 & 66.7 & 35.6 \\
% Llama 2 70B & 58.3 & 3.3 & 46.7 & 6.7 & 10.0 & 25.0 & 25.0 \\
% \midrule
% \multicolumn{10}{l}{\textit{Open AI Models}} \\
% \midrule
% \texttt{gpt-3.5-turbo} & \textbf{71.7} & \textbf{26.7} & 63.3 & 10.0 & 33.3 & 90.0 & 49.2 \\
% \texttt{gpt-4-32k} & 66.7 & 25.0 & \textbf{65.0} & \textbf{36.7} & \textbf{38.3} & \textbf{93.3} & \textbf{54.2} \\
% \bottomrule
% \end{tabular}
% \caption{Comparing performance of different models on all languages in XRiSAWOZ. Metric: Average.}
% \label{tab:results_summary}
% \end{table*}

% BeleBele
\begin{table*}[h]
\centering 
\small
\begin{tabular}{l c c c c c c c c c c c c }
\toprule
Model & ar & cs & da & de & en & es & fi & fr & he & hu & it & ja \\
\midrule
\multicolumn{10}{l}{\textit{Prompt-Based Baselines}} \\
\midrule
Llama 2 70B & 42.3 & 65.0 & 66.2 & 69.4 & 78.8 & 68.4 & 62.7 & 72.2 & 41.4 & 61.1 & 68.4 & 56.6 \\
Mistral 7B Instruct & 1 & 54.2 & 51.3 & 51.5 & 65.6 & 59.4 & 35.2 & 60 & 0 & 42 & 51.4 & 41.7 \\
\midrule
\multicolumn{10}{l}{\textit{Google Models}} \\
\midrule
PaLM 2 & 86.9 & 87.8 & 88.1 & 87.6 & \textbf{92.2} & 86.0 & 86.6 & 88.2 & 0.0 & 87.0 & 86.3 & 84.1 \\
gemini-pro & 89.0 & \textbf{90.4} & \textbf{89.8} & \textbf{88.8} & 90.8 & \textbf{89.3} & \textbf{89.3} & 89.8 & \textbf{88.0} & \textbf{88.2} & \textbf{89.3} & \textbf{86.9}
\\
Gemma 2B Instruct & 36.1 & 35.0 & 34.0 & 34.7 & 42.2 & 39.2 & 34.8 & 40.0 & 31.7 & 29.1 & 37.3 & 35.8\\
\end{tabular}
\caption{Comparing performance of different models on all languages in BeleBele. Metric: Accuracy.}
\label{tab:results_summary_belebele}
\end{table*}

\begin{table*}[h]
\centering 
\small
\begin{tabular}{@{}llcccclllll@{}}
\toprule
Model & \multicolumn{1}{c}{ar} & cs & da & de & en & es & fi & fr & it & ja \\ \midrule
\textit{LLaVA Models} & \multicolumn{1}{c}{} &  &  &  &  &  &  &  &  &  \\ \midrule
bakllava-v1 & \multicolumn{1}{c}{0.12} & 11.64 & 15.39 & 14.97 & 29.81 & 19.00 & 12.80 & 18.73 & 17.23 & 0.10 \\
llava-v1.5-13B & \multicolumn{1}{c}{16.13} & 19.37 & 27.04 & 28.34 & 30.06 & 30.22 & 20.99 & 32.10 & 28.10 & 10.40 \\
vip-llava-13B & \multicolumn{1}{c}{12.24} & 18.41 & 25.26 & 28.47 & 30.63 & 30.32 & 19.26 & 31.86 & 27.79 & 9.59 \\ \midrule
\textit{Google Models} & \multicolumn{1}{c}{} &  &  &  &  &  &  &  &  &  \\ \midrule
\texttt{gemini-pro-vision} & \multicolumn{1}{c}{25.00} & 25.42 & 32.31 & 31.14 & 32.39 & 32.74 & 27.44 & 33.95 & 32.03 & 13.37 \\ \midrule
\textit{OpenAI Models} & \multicolumn{1}{c}{} &  &  &  &  &  &  &  &  &  \\ \midrule
\texttt{gpt-4-vision} & \multicolumn{1}{c}{28.47} & 31.67 & 35.11 & 34.96 & 37.67 & 38.32 & 31.96 & 39.59 & 35.88 & 15.81 \\ \midrule
 & ko & nl & no & pl & pt & ru & sv & th & tr & zh \\ \midrule
\textit{LLaVA Models} &  &  &  &  &  &  &  &  &  &  \\ \midrule
bakllava-v1 & 0.11 & 16.57 & 15.21 & 12.58 & 16.44 & 0.57 & 15.25 & 0.49 & 12.79 & 0.13 \\
llava-v1.5-13B & 6.82 & 31.56 & 26.28 & 21.53 & 28.20 & 19.51 & 27.26 & 14.53 & 16.74 & 8.57 \\
vip-llava-13B & 6.60 & 30.08 & 25.12 & 17.19 & 19.40 & 12.44 & 25.49 & 9.87 & 15.47 & 7.67 \\ \midrule
\textit{Google Models} &  &  &  &  &  &  &  &  &  &  \\ \midrule
\texttt{gemini-pro-vision} & 10.91 & 33.88 & 30.26 & 28.29 & 32.70 & 24.93 & 30.53 & 26.34 & 27.79 & 8.76 \\ \midrule
\textit{OpenAI Models} &  &  &  &  &  &  &  &  &  &  \\ \midrule
\texttt{gpt-4-vision} & 15.44 & 39.29 & 34.19 & 32.82 & 36.26 & 30.76 & 35.00 & 32.37 & 32.00 & 10.15 \\ \bottomrule
\end{tabular}
\caption{Comparing performance of Multimodal models on XM3600. Metric: chrF}
\label{tab:xm3600_results}
\end{table*}

% Jigsaw (Note - no Llama 2 results)
\begin{table*}[h]
\footnotesize
\centering
\begin{tabular}{@{}lcccccccl@{}} 
\toprule
Model                           & es                                        & fr                                        & it                                        & pt                                        & ru                                        & tr                                        & \textbf{avg} &   \\ 
\midrule
\multicolumn{9}{l}{\textit{Prompt-Based Baselines}}                                                                                                                                                                                                                                                                          \\ 
\midrule
PaLM (0-Shot)                   & 79.83                                     & 78.99                                     & -                                         & 77.58                                     & 80.35                                     & 84.1                                      & 80.17        &   \\
PaLM (10-Shot Monolingual)      & \textbf{91.23}                                     & 86.16                                     & -                                         & 90.99                                     & 92.47                                     & 84.5                                      & 89.07        &   \\
PaLM-2 (0-Shot)                 & 88.6                                      & 84.11                                     & -                                         & 87.68                                     & 90.5                                      & 93.42                                     & 88.86        &   \\
\bottomrule
\end{tabular}
\caption{Comparing performance of different models on all languages in Jigsaw. Metric: Accuracy. PaLM 2 does not support all languages; unsupported languages are marked with `-'. Averages for PaLM 2 are calculated only from supported languages.}
\label{tab:jigsaw_results_summary}
\end{table*}

% WinoMT (Note - No Palm 2 & Llama 2 results)
\begin{table*}[h]
\centering 
\tiny
\begin{tabular}{ccccccccccccclccccc}
\hline
 &
  \multicolumn{3}{c}{Google} &
  \multicolumn{3}{c}{Microsoft} &
  \multicolumn{3}{c}{Amazon} &
  \multicolumn{3}{c}{Systran} &
  \multicolumn{3}{c}{GPT Turbo 3.5} &
  \multicolumn{3}{c}{Bloomz} \\
 &
  Acc &
  $\Delta_G$\ &
  $\Delta_S$\ &
  Acc &
  $\Delta_G$\ &
  $\Delta_S$\ &
  Acc &
  $\Delta_G$\ &
  $\Delta_S$\ &
  Acc &
  $\Delta_G$\ &
  $\Delta_S$\ &
  \multicolumn{1}{c}{Acc} &
  $\Delta_G$\ &
  $\Delta_S$\ &
  Acc &
  $\Delta_G$\ &
  $\Delta_S$\ \\ \hline
es &
  50.9 &
  23.2 &
  20.9 &
  45 &
  36.5 &
  22.9 &
  \textbf{57.2} &
  15.3 &
  21.7 &
  42.5 &
  46.2 &
  15.6 &
  54.9 &
  22.7 &
  26.2 &
  55.6 &
  17.2 &
  32.5 \\
fr &
  \textbf{61.6} &
  6.1 &
  22.3 &
  44.5 &
  34.2 &
  15.8 &
  54.2 &
  16.4 &
  15 &
  43.4 &
  41.8 &
  -0.1 &
  52.7 &
  21.4 &
  26.1 &
  52 &
  17.8 &
  24.6 \\
it &
  38.6 &
  32.9 &
  18.6 &
  38.8 &
  41.8 &
  10.5 &
  40.2 &
  26.8 &
  14.7 &
  38.1 &
  47.3 &
  6.3 &
  45.1 &
  21.9 &
  26.7 &
  \textbf{45.7} &
  9 &
  18.5 \\ \hline
ru &
  37.8 &
  36.7 &
  11.4 &
  36.9 &
  42 &
  8.4 &
  39.8 &
  34.8 &
  9.4 &
  37.3 &
  44.1 &
  9.2 &
  \textbf{41} &
  31.6 &
  10.2 &
  5.9 &
  INV &
  0 \\
uk &
  38.4 &
  43.5 &
  10.7 &
  41.3 &
  46.8 &
  11.9 &
  - &
  - &
  - &
  28.9 &
  22.4 &
  12.9 &
  \textbf{42.9} &
  34.2 &
  12.1 &
  16.8 &
  22.7 &
  2.2 \\ \hline
he &
  50.8 &
  11.7 &
  35.5 &
  44 &
  22 &
  29.8 &
  48 &
  13.6 &
  45.9 &
  43.1 &
  26.9 &
  23.1 &
  \textbf{57.5} &
  7.6 &
  40.8 &
  27.5 &
  31.4 &
  5 \\
ar &
  45.8 &
  42.5 &
  16.2 &
  45 &
  47.1 &
  14.2 &
  48.3 &
  37.8 &
  18.8 &
  45.6 &
  49.4 &
  -4.1 &
  \textbf{61.1} &
  13.9 &
  27.9 &
  48.1 &
  23 &
  25.6 \\ \hline
de &
  59.4 &
  12.5 &
  12.6 &
  \textbf{74.1} &
  0 &
  8.8 &
  62.4 &
  12 &
  16.7 &
  48.5 &
  34.5 &
  10 &
  57.5 &
  19.5 &
  14.2 &
  47.6 &
  56.2 &
  6.6 \\ \hline
\end{tabular}%
\caption{Performance of commercial MT systems and LLMs on the WinoMT corpus on 8 target languages. Results are categorized by language family. Acc indicates overall gender accuracy (\% of instances the translation had the correct gender), $\Delta_G$ denotes the difference in performance (F1 score) between masculine and feminine scores, and $\Delta_S$ is the difference in performance (F1 score) between pro-stereotypical and anti-stereotypical gender role assignments (higher numbers in the two latter metrics indicate stronger biases). Numbers in bold indicate best accuracy for the language across all systems. \footnotesize {Notes: [1. For Google, Microsoft, Amazon, and Systran we use the translations provided by \cite{stanovsky2019evaluating}. Some values differ from the original paper due to updated Spcay modules. 2. For Ru in Bloomz, Precision in male predictions is 0 leading to Invalid (INV) in $\Delta_G$]}}
\label{tab:wino-mt}
\end{table*}

\begin{table*}[h]
\centering 
\tiny
\begin{tabular}{@{}lcccccccccccccc@{}}
\toprule
Model & as & bn & gu & hi & ka & kn & ml & mr & np & or & pa & ta & te & ur \\ \midrule
\textit{SOTA model} & \multicolumn{1}{l}{} & \multicolumn{1}{l}{} & \multicolumn{1}{l}{} & \multicolumn{1}{l}{} & \multicolumn{1}{l}{} & \multicolumn{1}{l}{} & \multicolumn{1}{l}{} & \multicolumn{1}{l}{} & \multicolumn{1}{l}{} & \multicolumn{1}{l}{} & \multicolumn{1}{l}{} & \multicolumn{1}{l}{} & \multicolumn{1}{l}{} & \multicolumn{1}{l}{} \\ \midrule
IT2 & 46.8 & 49.7 & 53.1 & 49.6 & 35.6 & 33.8 & 45.7 & 48.6 & 51.5 & 40.2 & 57.8 & 39.1 & 45.5 & 61.6 \\ \midrule
\textit{Prompt-Based Baselines} & \multicolumn{1}{l}{} & \multicolumn{1}{l}{} & \multicolumn{1}{l}{} & \multicolumn{1}{l}{} & \multicolumn{1}{l}{} & \multicolumn{1}{l}{} & \multicolumn{1}{l}{} & \multicolumn{1}{l}{} & \multicolumn{1}{l}{} & \multicolumn{1}{l}{} & \multicolumn{1}{l}{} & \multicolumn{1}{l}{} & \multicolumn{1}{l}{} & \multicolumn{1}{l}{} \\ \midrule
Llama 2 70B & 0.7 & 0.7 & 0.9 & 1.3 & 1 & 1 & 1.3 & 1.5 & 0.8 & 0.7 & 0.7 & 1.1 & 1.1 & 1.9 \\
Mistral 7B Instruct & 2.1 & 5.5 & 2.2 & 7.4 & 3.8 & 4 & 2.5 & 5 & 5.4 & 1.6 & 1.8 & 3.8 & 2.9 & 6 \\
Gemma 7B Instruct & 13.9 & 21.9 & 4.7 & 26.9 & 10.3 & 2.2 & 6.3 & 15.7 & 16.5 & 1.6 & 4.2 & 18.6 & 10.5 & 1.6 \\ \midrule
\textit{OpenAI Models} & \multicolumn{1}{l}{} & \multicolumn{1}{l}{} & \multicolumn{1}{l}{} & \multicolumn{1}{l}{} & \multicolumn{1}{l}{} & \multicolumn{1}{l}{} & \multicolumn{1}{l}{} & \multicolumn{1}{l}{} & \multicolumn{1}{l}{} & \multicolumn{1}{l}{} & \multicolumn{1}{l}{} & \multicolumn{1}{l}{} & \multicolumn{1}{l}{} & \multicolumn{1}{l}{} \\
\bottomrule
\end{tabular}
\caption{Performance of various models on IN22-Conv in the En-Indic direction. All the LLMs are prompted with 8 few-shot examples. However, we also report the 0-shot performance from \citet{gala2023indictrans}}
\label{tab:in22-conv-en-indic}
\end{table*}

\begin{table*}[h]
\centering
\tiny
\begin{tabular}{@{}lcccccccccccccc@{}}
\toprule
Model & as & bn & gu & hi & ka & kn & ml & mr & np & or & pa & ta & te & ur \\ \midrule
\textit{SOTA model} & \multicolumn{1}{l}{} & \multicolumn{1}{l}{} & \multicolumn{1}{l}{} & \multicolumn{1}{l}{} & \multicolumn{1}{l}{} & \multicolumn{1}{l}{} & \multicolumn{1}{l}{} & \multicolumn{1}{l}{} & \multicolumn{1}{l}{} & \multicolumn{1}{l}{} & \multicolumn{1}{l}{} & \multicolumn{1}{l}{} & \multicolumn{1}{l}{} & \multicolumn{1}{l}{} \\ \midrule
IT2 & 62.9 & 58.4 & 62 & 60.1 & 52.6 & 47.5 & 54.3 & 58.5 & 63 & 60.3 & 62.7 & 45.8 & 52.9 & 65.5 \\ \midrule
\textit{Prompt-Based Baselines} & \multicolumn{1}{l}{} & \multicolumn{1}{l}{} & \multicolumn{1}{l}{} & \multicolumn{1}{l}{} & \multicolumn{1}{l}{} & \multicolumn{1}{l}{} & \multicolumn{1}{l}{} & \multicolumn{1}{l}{} & \multicolumn{1}{l}{} & \multicolumn{1}{l}{} & \multicolumn{1}{l}{} & \multicolumn{1}{l}{} & \multicolumn{1}{l}{} & \multicolumn{1}{l}{} \\ \midrule
Llama 2 70B & 16.2 & 19.8 & 14.9 & 30.3 & 14.7 & 13.3 & 14.7 & 18.9 & 22.4 & 13.5 & 14.1 & 14.2 & 13.7 & 25 \\
Mistral 7B Instruct & 17 & 22.5 & 11.7 & 27.4 & 11.9 & 12.7 & 10.2 & 16.7 & 19 & 9.7 & 10.4 & 11.7 & 11.3 & 21.5 \\
Gemma 7B Instruct & 18.8 & 27.4 & 23.3 & 38 & 19 & 14.8 & 22 & 25.1 & 26.3 & 12.1 & 22 & 21.7 & 24.5 & 33.8 \\ \midrule
\textit{OpenAI Models} & \multicolumn{1}{l}{} & \multicolumn{1}{l}{} & \multicolumn{1}{l}{} & \multicolumn{1}{l}{} & \multicolumn{1}{l}{} & \multicolumn{1}{l}{} & \multicolumn{1}{l}{} & \multicolumn{1}{l}{} & \multicolumn{1}{l}{} & \multicolumn{1}{l}{} & \multicolumn{1}{l}{} & \multicolumn{1}{l}{} & \multicolumn{1}{l}{} & \multicolumn{1}{l}{} \\
\bottomrule
\end{tabular}
\caption{Performance of various models on IN22-Conv in the Indic-En direction. All the LLMs are prompted with 8 few-shot examples. However, we also report the 0-shot performance from \citet{gala2023indictrans}}
\label{tab:in22-conv-indic-en}
\end{table*}

\begin{table*}[h]
\centering
\tiny
\begin{tabular}{@{}lcccccccccccccc@{}}
\toprule
Model & as & bn & gu & hi & ka & kn & ml & mr & np & or & pa & ta & te & ur \\ \midrule
\textit{SOTA model} & \multicolumn{1}{l}{} & \multicolumn{1}{l}{} & \multicolumn{1}{l}{} & \multicolumn{1}{l}{} & \multicolumn{1}{l}{} & \multicolumn{1}{l}{} & \multicolumn{1}{l}{} & \multicolumn{1}{l}{} & \multicolumn{1}{l}{} & \multicolumn{1}{l}{} & \multicolumn{1}{l}{} & \multicolumn{1}{l}{} & \multicolumn{1}{l}{} & \multicolumn{1}{l}{} \\ \midrule
IT2 & 47.1 & 51.8 & 53.5 & 56.7 & 40.2 & 51 & 50.9 & 51 & 49 & 43.9 & 50.6 & 49.5 & 52.4 & 68.2 \\ \midrule
\textit{Prompt-Based Baselines} & \multicolumn{1}{l}{} & \multicolumn{1}{l}{} & \multicolumn{1}{l}{} & \multicolumn{1}{l}{} & \multicolumn{1}{l}{} & \multicolumn{1}{l}{} & \multicolumn{1}{l}{} & \multicolumn{1}{l}{} & \multicolumn{1}{l}{} & \multicolumn{1}{l}{} & \multicolumn{1}{l}{} & \multicolumn{1}{l}{} & \multicolumn{1}{l}{} & \multicolumn{1}{l}{} \\ \midrule
Llama 2 70B & 0.5 & 0.7 & 0.6 & 1.2 & 1 & 0.7 & 0.9 & 1 & 0.9 & 0.5 & 0.8 & 0.9 & 0.9 & 1.3 \\
Mistral 7B Instruct & 1.1 & 2.5 & 1.1 & 3.9 & 2.7 & 2.1 & 1.3 & 2.9 & 3.1 & 0.8 & 1 & 2.1 & 1.6 & 3.7 \\
Gemma 7B Instruct & 16.1 & 24 & 3.5 & 32.1 & 14.5 & 2.8 & 6.5 & 19.5 & 20.4 & 1.9 & 3.4 & 23.8 & 13.1 & 2.1 \\ \midrule
\textit{OpenAI Models} & \multicolumn{1}{l}{} & \multicolumn{1}{l}{} & \multicolumn{1}{l}{} & \multicolumn{1}{l}{} & \multicolumn{1}{l}{} & \multicolumn{1}{l}{} & \multicolumn{1}{l}{} & \multicolumn{1}{l}{} & \multicolumn{1}{l}{} & \multicolumn{1}{l}{} & \multicolumn{1}{l}{} & \multicolumn{1}{l}{} & \multicolumn{1}{l}{} & \multicolumn{1}{l}{} \\
\bottomrule
\end{tabular}
\caption{Performance of various models on IN22-Gen in the En-Indic direction. All the LLMs are prompted with 8 few-shot examples. However, we also report the 0-shot performance from \citet{gala2023indictrans}}
\label{tab:in22-gen-en-indic}
\end{table*}

\begin{table*}[h]
\centering
\tiny
\begin{tabular}{@{}lcccccccccccccc@{}}
\toprule
Model & as & bn & gu & hi & ka & kn & ml & mr & np & or & pa & ta & te & ur \\ \midrule
\textit{SOTA model} & \multicolumn{1}{l}{} & \multicolumn{1}{l}{} & \multicolumn{1}{l}{} & \multicolumn{1}{l}{} & \multicolumn{1}{l}{} & \multicolumn{1}{l}{} & \multicolumn{1}{l}{} & \multicolumn{1}{l}{} & \multicolumn{1}{l}{} & \multicolumn{1}{l}{} & \multicolumn{1}{l}{} & \multicolumn{1}{l}{} & \multicolumn{1}{l}{} & \multicolumn{1}{l}{} \\ \midrule
IT2 & 65.8 & 63.2 & 66.5 & 65.4 & 60.4 & 64.2 & 64.5 & 63.7 & 67.7 & 66.2 & 63.4 & 59.8 & 64.8 & 73 \\ \midrule
\textit{Prompt-Based Baselines} & \multicolumn{1}{l}{} & \multicolumn{1}{l}{} & \multicolumn{1}{l}{} & \multicolumn{1}{l}{} & \multicolumn{1}{l}{} & \multicolumn{1}{l}{} & \multicolumn{1}{l}{} & \multicolumn{1}{l}{} & \multicolumn{1}{l}{} & \multicolumn{1}{l}{} & \multicolumn{1}{l}{} & \multicolumn{1}{l}{} & \multicolumn{1}{l}{} & \multicolumn{1}{l}{} \\ \midrule
Llama 2 70B & 13.1 & 15.8 & 12.4 & 23.6 & 12.8 & 11.3 & 12.2 & 16.7 & 18.8 & 11.7 & 11.7 & 12 & 11.4 & 19.4 \\
Mistral 7B Instruct & 13.7 & 17 & 10.8 & 20.4 & 12.9 & 12.3 & 9.9 & 15.3 & 16.9 & 9.8 & 9.9 & 11.8 & 11.4 & 17.4 \\
Gemma 7B Instruct & 21.2 & 29.2 & 24.7 & 42.1 & 23.9 & 17.9 & 23.9 & 29.4 & 31.1 & 14.5 & 24.2 & 24.9 & 28.8 & 36.8 \\ \midrule
\textit{OpenAI Models} & \multicolumn{1}{l}{} & \multicolumn{1}{l}{} & \multicolumn{1}{l}{} & \multicolumn{1}{l}{} & \multicolumn{1}{l}{} & \multicolumn{1}{l}{} & \multicolumn{1}{l}{} & \multicolumn{1}{l}{} & \multicolumn{1}{l}{} & \multicolumn{1}{l}{} & \multicolumn{1}{l}{} & \multicolumn{1}{l}{} & \multicolumn{1}{l}{} & \multicolumn{1}{l}{} \\
\bottomrule
\end{tabular}
\caption{Performance of various models on IN22-Gen in the Indic-En direction. All the LLMs are prompted with 8 few-shot examples. However, we also report the 0-shot performance from \citet{gala2023indictrans}}
\label{tab:in22-gen-indic-en}
\end{table*}

% X-RiSAWOZ
\begin{table*}[h]
\footnotesize
% \resizebox{\textwidth}{!}{%
\centering
\begin{tabular}{l c c c c c c c}
\toprule
Model & en & en-hi & fr & hi & ko & zh & \textbf{avg}\\ \midrule
\multicolumn{8}{l}{\textit{Fine-tuned Baselines}} \\
\midrule
mBART & \textbf{84.6} & \textbf{60.7} & \textbf{73.1} & \textbf{75.3} & \textbf{71.2} & \textbf{91.7} & \textbf{76.1} \\
\midrule
\multicolumn{8}{l}{\textit{Prompt-Based Baselines}} \\
\midrule
Llama 2 70B & 59.0 & 44.0 & 51.8 & 38.4 & 51.1 & 73.0 & 52.9 \\ 
Mistral 7B & 49.8 & 38.1 & 44.3 & 35.5 & 41.4 & 51.8 & 43.5 \\ 
Mistral 7B Instruct & 48.2 & 38.1 & 42.4 & 31.9 & 41.4 & 50.8 & 42.1 \\ 
\midrule
\multicolumn{8}{l}{\textit{Google Models}} \\
\midrule
PaLM 2 & 62.2 & 45.3 & 51.5 & 52.1 & 55.4 & 73.9 & 56.7 \\ 
\texttt{gemini-pro} & 64.4 & 49.3 & 52.2 & 53.0 & 53.2 & 73.6 & 57.6 \\
Gemma 2B Instruct & 49.3 & 33.5 & 52.2 & 38.7 & 51.4 & 58.4 & 47.2 \\
Gemma 7B Instruct & 50.3 & 33.7 & 52.0 & 39.5 & 52.1 & 58.0 & 47.8 \\
\midrule
\multicolumn{8}{l}{\textit{Open AI Models}} \\
\midrule
\texttt{gpt-3.5-turbo} & 71.0 & 50.2 & 60.3 & 57.3 & 59.9 & 81.4 & 63.4 \\
\texttt{gpt-4-32k} & 75.6 & 57.7 & 69.0 & 63.2 & 69.1 & 85.3 & 70.0 \\ 
\bottomrule
\end{tabular}
\caption{Comparing performance of different models on all languages in X-RiSAWOZ. Metric: Dialogue Action Accuracy. The fine-tuned baseline is taken from \citet{moradshahi-etal-2023-x}.}
\label{tab:results_xrisawoz}
\end{table*}

\begin{table*}[!ht]
    \centering
    \small
    \begin{tabular}{llccccccc}
    \toprule
        \multirow{2}{*}{Metric} & \multirow{2}{*}{Model} & \multicolumn{6}{c}{Languages} & \multirow{2}{*}{\textbf{avg}} \\
        \cline{3-8}
        ~ & ~ & en & en-hi & fr & hi & ko & zh & ~ \\ 
    \midrule
        \multirow{6}{*}{BLEU ($\uparrow$)} & Llama 2 & 12.8 & 6.6 & 13.6 & 3.4 & 2.1 & 1.7 & 6.7 \\
        ~ & Mistral 7B & 24.9 & 7.7 & 22.3 & 4.7 & 9.0 & 12.3 & 13.5 \\ 
        ~ & Mistral 7B Instruct & 30.7 & 5.5 & 23.8 & 3.4 & 9.6 & 15.1 & 14.7 \\
        ~ & PaLM 2 & 35.9 & 21.9 & \textbf{31.9} & \textbf{26.4} & 21.7 & 24.3 & 27.0\\
        ~ & \texttt{gemini-pro} & \textbf{36.3} & \textbf{25.7} & 33.3 & 16.7 & 23.8 & \textbf{26.8} & \textbf{27.1}\\
        ~ & Gemma 2B Instruct & 26.0 & 21.2 & 13.7 & 6.4 & 25.9 & 24.3 & 19.6\\
\bottomrule
    \end{tabular}
    \caption{Comparison of various task-specific metrics on X-RiSAWOZ. ($\uparrow$) indicates metric is higher the better and ($\downarrow$) the indicates lower the better. Best results for a particular model-language combination are bolded.}
    \label{tab:detailed_xrisawoz}
\end{table*}

%% file: appendix/apex_cont.tex
\clearpage
% \subsection{Contamination}
% \label{sec:appendix_contamination}
% Tables \ref{tab:cont_tydiqa} to \ref{tab:cont_xnli} show the contamination values for the various datasets.

\begin{table*}[h]
\centering
{\small
\begin{tabular}[\textwidth]{@{}lllllllllll@{}}
\toprule
\multicolumn{1}{c}{Model} &
\multicolumn{1}{c}{ar} &
\multicolumn{1}{c}{en} &
\multicolumn{1}{c}{fi} &
\multicolumn{1}{c}{id} &
\multicolumn{1}{c}{ja} &
\multicolumn{1}{c}{ko} &
\multicolumn{1}{c}{ru} &
\multicolumn{1}{c}{sw} &
\multicolumn{1}{c}{te} &
\multicolumn{1}{c}{th} \\ \midrule
GPT-4 &
-0.25 &
0.73 &
0.45 &
0.36 &
0.36 &
0.40 &
0.53 &
0.40 &
0.41 &
0.46 \\
PaLM-2 &
0.55 &
0.64 &
0.07 &
0.16 &
0.72 &
0.60 &
0.61 &
0.23 &
NA &
0.17 \\ \bottomrule
\end{tabular}
\caption{Contamination values for the TydiQA dataset.}
\label{tab:cont_tydiqa}}
\end{table*}

\begin{table*}[t!]
\centering
{\small
\begin{tabular}[\textwidth]{@{}llllllll@{}}
\toprule
\multicolumn{1}{c}{Model} &
  \multicolumn{1}{c}{de} &
  \multicolumn{1}{c}{en} &
  \multicolumn{1}{c}{es} &
  \multicolumn{1}{c}{fr} &
  \multicolumn{1}{c}{ja} &
  \multicolumn{1}{c}{ko} &
  \multicolumn{1}{c}{zh} \\ \midrule
GPT-4 &
  0.77 &
  0.72 &
  0.66 &
  0.71 &
  0.55 &
  0.44 &
  0.65 \\
PaLM-2 &
  0.23 &
  0.63 &
  0.16 &
  0.23 &
  0.53 &
  0.57 &
  0.32 \\ \bottomrule
\end{tabular}
\caption{Contamination values for the PAWS-X dataset.}
\label{tab:cont_pawsx}}
\end{table*}

\begin{table*}[t!]
\centering
\small
\begin{tabular}{@{}llllllllllllll@{}}
\toprule
\multicolumn{1}{c}{Model} &
  \multicolumn{1}{c}{ar} &
  \multicolumn{1}{c}{bg} &
  \multicolumn{1}{c}{el} &
  \multicolumn{1}{c}{en} &
  \multicolumn{1}{c}{et} &
  \multicolumn{1}{c}{eu} &
  \multicolumn{1}{c}{fi} &
  \multicolumn{1}{c}{fr} &
  \multicolumn{1}{c}{hi} &
  \multicolumn{1}{c}{hu} &
  \multicolumn{1}{c}{it} &
  \multicolumn{1}{c}{ja} &
  \multicolumn{1}{c}{lt} \\ \midrule
GPT-4 &
  0.41 &
  0.05 &
  0.17 &
  0.21 &
  0.16 &
  0.13 &
  0.04 &
  0.09 &
  0.32 &
  0.2 &
  0.03 &
  0.23 &
  0.08 \\
PaLM-2 &
  -0.05 &
  -0.2 &
  -0.12 &
  0.04 &
  -0.2 &
  NA &
  -0.13 &
  0.12 &
  -0.17 &
  -0.16 &
  -0.03 &
  -0.08 &
  -0.13 \\ \bottomrule
\end{tabular}
\caption{Contamination values for the UDPOS dataset (part-1)}
\label{tab:cont_udpos_1}
\end{table*}

\begin{table*}[t!]
\centering
\small
\begin{tabular}{@{}llllllllllllll@{}}
\toprule
\multicolumn{1}{c}{Model} &
  \multicolumn{1}{c}{mr} &
  \multicolumn{1}{c}{pl} &
  \multicolumn{1}{c}{pt} &
  \multicolumn{1}{c}{ro} &
  \multicolumn{1}{c}{ru} &
  \multicolumn{1}{c}{ta} &
  \multicolumn{1}{c}{te} &
  \multicolumn{1}{c}{tr} &
  \multicolumn{1}{c}{uk} &
  \multicolumn{1}{c}{vi} &
  \multicolumn{1}{c}{wo} &
  \multicolumn{1}{c}{zh}  \\ \midrule
GPT-4 &
  0.32 &
  -0.01 &
  0.03 &
  -0.25 &
  0.15 &
  0.11 &
  -0.09 &
  0.17 &
  0.05 &
  -0.01 &
  -0.12 &
  0.07  \\
PaLM-2 &
  NA &
  -0.21 &
  -0.09 &
  -0.13 &
  -0.07 &
  NA &
  NA &
  0.12 &
  -0.23 &
  -0.2 &
  NA &
  -0.13  \\ \bottomrule
\end{tabular}
\caption{Contamination values for the UDPOS dataset (part-2)}
\label{tab:cont_udpos_2}
\end{table*}

\begin{table*}[t!]
\centering
\small
\begin{tabular}{@{}lllllllllll@{}}
\toprule
\multicolumn{1}{c}{Model} &
  \multicolumn{1}{c}{et} &
  \multicolumn{1}{c}{ht} &
  \multicolumn{1}{c}{id} &
  \multicolumn{1}{c}{it} &
  \multicolumn{1}{c}{sw} &
  \multicolumn{1}{c}{ta} &
  \multicolumn{1}{c}{th} &
  \multicolumn{1}{c}{tr} &
  \multicolumn{1}{c}{vi} &
  \multicolumn{1}{c}{zh} \\ \midrule
GPT-4 &
  NA &
  0.79 &
  0.55 &
  0.72 &
  0.64 &
  0.73 &
  0.67 &
  0.67 &
  0.37 &
  0.41 \\
PaLM-2 &
  -0.05 &
  NA &
  -0.17 &
  -0.07 &
  -0.07 &
  NA &
  -0.2 &
  -0.24 &
  -0.09 &
  -0.13 \\ \bottomrule
\end{tabular}
\caption{Contamination values for the X-COPA dataset.}
\label{tab:cont_xcopa}
\end{table*}

\begin{table*}[t!]
\centering
\small
\begin{tabular}{@{}llllllllllllllll@{}}
\toprule
\multicolumn{1}{c}{Model} &
  \multicolumn{1}{c}{ar} &
  \multicolumn{1}{c}{bg} &
  \multicolumn{1}{c}{de} &
  \multicolumn{1}{c}{el} &
  \multicolumn{1}{c}{en} &
  \multicolumn{1}{c}{es} &
  \multicolumn{1}{c}{fr} &
  \multicolumn{1}{c}{hi} &
  \multicolumn{1}{c}{ru} &
  \multicolumn{1}{c}{sw} &
  \multicolumn{1}{c}{th} &
  \multicolumn{1}{c}{tr} &
  \multicolumn{1}{c}{ur} &
  \multicolumn{1}{c}{vi} &
  \multicolumn{1}{c}{zh} \\ \midrule
GPT-4 &
  0.47 &
  0.4 &
  0.23 &
  0.51 &
  0.45 &
  0.36 &
  0.37 &
  0.59 &
  0.32 &
  0.24 &
  0.2 &
  0.37 &
  0.45 &
  0.25 &
  0.08 \\
PaLM-2 &
  0.47 &
  0.31 &
  0.17 &
  0.47 &
  0.37 &
  0.13 &
  0.27 &
  0.33 &
  0.25 &
  0.15 &
  0.09 &
  0.31 &
   NA &
  0.61 &
  0.36 \\ \bottomrule
\end{tabular}
\caption{Contamination values for the XNLI dataset.}
\label{tab:cont_xnli}
\end{table*}

\begin{table*}[]
\centering
\begin{tabular}{lccc}
\hline
Dataset & \multicolumn{1}{l}{Gemma 7B Instruct} & \multicolumn{1}{l}{Llama 2 7B Instruct} & \multicolumn{1}{l}{Mistral 7B Instruct} \\ \hline
PAWS-X      & \textbf{0.0}    & \textbf{0.0} & \textbf{0.0} \\
XCOPA       & \textbf{0.0007} & \textbf{0.0} & \textbf{0.0} \\
XNLI        & 0.4162          & 0.0374       & 0.1148       \\
XQUAD       & 0.0164          & \textbf{0.0} & \textbf{0.0} \\
XRiSAWOZ    & \textbf{0.0}    & \textbf{0.0} & \textbf{0.0} \\
XstoryCloze & 0.2917          & 0.0274       & 0.2743       \\ \hline
\end{tabular}
\caption{The statistical test was performed on a total of 5000 test points equally divided amongst all the languages of a given dataset. Our significance value is 0.001 which is calculated using $1/(1+r)$, where $r$ is the number of permutations per shard (for us it is 700). If a value is less than 0.001, then that test set is contaminated for the given model. The it suffix for the above model stands for Instruction-Tuned variant of that said model.}
\label{tab:opensource_contamination}
\end{table*}

% Please add the following required packages to your document preamble:
% \usepackage{booktabs}
% \usepackage[normalem]{ulem}
% \useunder{\uline}{\ul}{}
\begin{table*}[]
\centering
\resizebox{\textwidth}{!}{
\begin{tabular}{@{}llll@{}}
\toprule
\textbf{Language} & \textbf{Language Family} & \textbf{Language Script} & \textbf{ISO code} \\ \midrule
Afrikaans   & IE: Germanic     & Latin              & af \\
Amharic           & Afro-Asiatic             & Ge'ez (Ethiopic)         & am                \\
Arabic      & Afro-Asiatic     & Arabic             & ar \\
Assamese    & IE: Iranian      & Brahmic            & as \\
Azerbaijani & Turkic           & Latin              & az \\
Basque      & Basque           & Latin              & eu \\
Bengali     & IE: Iranian      & Brahmic            & bn \\
Bulgarian   & IE: Balto-Slavic & Cyrillic           & bg \\
Burmese     & Sino-Tibetan     & Brahmic            & my \\
Mandarin          & Sino-Tibetan             & Chinese ideograms        & zh                \\
Dutch       & IE: Germanic     & Latin              & nl \\
English     & IE: Germanic     & Latin              & en \\
Czech       & IE: Balto-Slavic & Latin              & cs \\
Estonian    & Uralic           & Latin              & et \\
Finnish     & Uralic           & Latin              & fi \\
French      & IE: Romance      & Latin              & fr \\
Georgian    & Kartvelian       & Georgian           & ka \\
German      & IE: Germanic     & Latin              & de \\
Greek       & IE: Greek        & Greek              & el \\
Gujarati    & IE: Iranian      & Brahmic            & gu \\
Hausa       & Afro-Asiatic     & Brahmic            & ha \\
Hebrew      & Afro-Asiatic     & Hebrew             & he \\
Hindi       & IE: Iranian      & Devanagari         & hi \\
Urdu        & IE: Iranian      & Arabic             & ur \\
Hungarian   & Uralic           & Latin              & hu \\
Igbo        & Niger-Congo      & Latin              & ig \\
Indonesian  & Austronesian     & Latin              & id \\
Italian     & IE: Romance      & Latin              & it \\
Japanese    & Japonic          & Japanese ideograms & ja \\
Javanese    & Austronesian     & Brahmic            & jv \\
Kannada     & Dravidian        & Brahmic            & kn \\
Kazakh      & Turkic           & Cyrillic           & kk \\
Kirundi     & Niger-Congo      & Latin              & rn \\
Korean      & Koreanic         & Hangul             & ko \\
Kyrgyz      & Turkic           & Cyrillic           & ky \\
Malay       & Austronesian     & Latin              & ms \\
Malayalam   & Dravidian        & Brahmic            & ml \\
Marathi     & IE: Iranian      & Devanagari         & mr \\
Nepali      & IE: Iranian      & Devanagari         & ne \\
Odia        & IE: Iranian      & Brahmic            & or \\
Oromo       & Afro-Asiatic     & Latin              & om \\
Pashto      & IE: Iranian      & Arabic             & ps \\ \bottomrule
\end{tabular}

\quad

% Please add the following required packages to your document preamble:
% \usepackage{booktabs}
\begin{tabular}{@{}llll@{}}
\toprule
\textbf{Language} & \textbf{Language Family} & \textbf{Language Script} & \textbf{ISO code} \\ \midrule
Persian          & IE: Iranian      & Arabic           & fa    \\
Pidgin           & IE: Germanic     & Latin            & pid   \\
Portuguese       & IE: Romance      & Latin            & pt    \\
Punjabi          & IE: Iranian      & Gurmukhi         & pa    \\
Russian          & IE: Balto-Slavic & Cyrillic         & ru    \\
Scottish\_gaelic & IE: Celtic       & Latin            & gd    \\
Serbian\_Cyrillic & IE: Balto-Slavic         & Cyrillic                 & sr                \\
Serbian\_Latin   & IE: Balto-Slavic & Latin            & sr    \\
Sinhala          & IE: Iranian      & Brahmic          & si    \\
Somali           & Afro-Asiatic     & Latin            & so    \\
Spanish          & IE: Romance      & Latin            & es    \\
Swahili          & Bantu            & Latin            & sw    \\
Tagalog          & Austronesian     & Brahmic          & tl    \\
Tamil            & Dravidian        & Brahmic          & ta    \\
Telugu           & Dravidian        & Brahmic          & te    \\
Thai             & Kra-Dai          & Brahmic          & th    \\
Tigrinya         & Afro-Asiatic     & Ge'ez (Ethiopic) & ti    \\
Turkish          & Turkic           & Latin            & tr    \\
Ukrainian        & IE: Balto-Slavic & Cyrillic         & uk    \\
Uzbek            & Turkic           & Latin            & uz    \\
Vietnamese       & Austro-Asiatic   & Latin            & vi    \\
Welsh            & IE: Celtic       & Latin            & cy    \\
Yoruba           & Niger-Congo      & Latin            & yo    \\
Bemba            & Bantu            & Latin            & bem   \\
Fon              & Niger-Congo      & Latin            & fon   \\
Kinyarwanda      & Bantu            & Latin            & rw    \\
Twi              & Kwa              & Latin            & tw    \\
Wolof            & Niger-Congo      & Latin            & wo    \\
Zulu             & Bantu            & Latin            & zu    \\
Czech            & IE: Balto-Slavic & Latin            & cs    \\
Danish           & IE: Germanic     & Latin            & da    \\
Norwegian        & IE: Germanic     & Latin            & no    \\
Polish           & IE: Balto-Slavic & Latin            & pl    \\
Swedish          & IE: Germanic     & Latin            & sv    \\
English-Hindi    & IE: Germanic     & Latin            & en-hi \\
English-Spanish  & IE: Romance      & Latin            & en-es \\
Kashmiri         & Dardic           & Arabic           & ks    \\
Lithuanian       & IE: Balto-Slavic & Latin            & lt    \\
Quechua          & Quechuan         & Latin            & qu    \\
Romanian         & IE: Romance      & Latin            & ro    \\
Haitian Creole   & French Creole    & Latin            & ht    \\
\\
\bottomrule
\end{tabular}
}
\caption{List of Languages and their corresponding Language Families, Language Scripts and ISO Codes benchmarked in }
\label{tab:lf_list}
\end{table*}